\RequirePackage{fix-cm}
\documentclass[twocolumn]{svjour3}          
 \pdfoutput=1
\smartqed  
\usepackage{graphicx}
\usepackage{amsmath}
\usepackage{mathrsfs}
\usepackage{amssymb}
\usepackage{url}
\usepackage{caption}
\usepackage{enumitem}
\usepackage{mathrsfs}
\usepackage{algorithm}
\usepackage{algpseudocode}
\usepackage{dsfont}
\usepackage{multirow}
\usepackage{booktabs}
\usepackage{array}
\usepackage{epstopdf}
\usepackage{times}
\usepackage{colortbl}
\usepackage{wrapfig}
\usepackage[normalem]{ulem}
\usepackage[pagebackref=false,breaklinks=true,letterpaper=true,colorlinks,urlcolor=magenta,citecolor=blue,linkcolor=blue,bookmarks=false]{hyperref}
\def\etal{\textit{et al}.}
\def\ie{\textit{i.e.}}
\def\eg{\textit{e.g.}}

\newcommand{\ryn}[1]{\textcolor{black}{#1}} 
\newcommand{\ghk}[1]{\textcolor{black}{#1}}

\usepackage{algorithmicx}
\floatname{algorithm}{Procedure}

\begin{document}
\begin{sloppypar}
\title{Delving into Dark Regions for Robust Shadow Detection 
}


\author{Huankang~Guan \and
        Ke~Xu    \and
        Rynson W.H. Lau 
}


\institute{
           Huankang Guan \at
           Huankang.Guan@my.cityu.edu.hk \at
           City University of Hong Kong, Hong Kong SAR, China.
           \and
           Ke Xu \at
           kkangwing@gmail.com     \at
           City University of Hong Kong, Hong Kong SAR, China.
           \and
           Rynson W.H. Lau \at
           Rynson.Lau@cityu.edu.hk \at
           City University of Hong Kong, Hong Kong SAR, China.
}

\date{This work was done in 2021.}

\maketitle

\begin{abstract}
Shadow detection is a challenging task as it requires a comprehensive understanding of shadow characteristics and global/local illumination conditions. We observe from our experiment that state-of-the-art deep methods tend to have higher error rates in differentiating shadow pixels from non-shadow pixels in dark regions (\ie, regions with low-intensity values). Our key insight to this problem is that existing methods typically learn discriminative shadow features from the whole image globally, covering the full range of intensity values, and may not learn the subtle differences between shadow and non-shadow pixels in dark regions. Hence, if we can design a model to focus on a narrower range of low-intensity regions, it may be able to learn better discriminative features for shadow detection. Inspired by this insight, we propose a novel shadow detection approach that first learns global contextual cues over the entire image and then zooms into the dark regions to learn local shadow representations. To this end, we formulate an effective dark-region recommendation (DRR) module to recommend regions of low-intensity values, and a novel dark-aware shadow analysis (DASA) module to learn dark-aware shadow features from the recommended dark regions. Extensive experiments show that the proposed method outperforms the state-of-the-art methods on three popular shadow detection datasets. Code is available at \url{https://github.com/guanhuankang/ShadowDetection2021.git}.
\keywords{Shadow detection \and Global/local shadow understanding \and Region-based shadow analysis}
\end{abstract}

\begin{figure}[t]
\begin{center}
	\includegraphics[width=\linewidth]{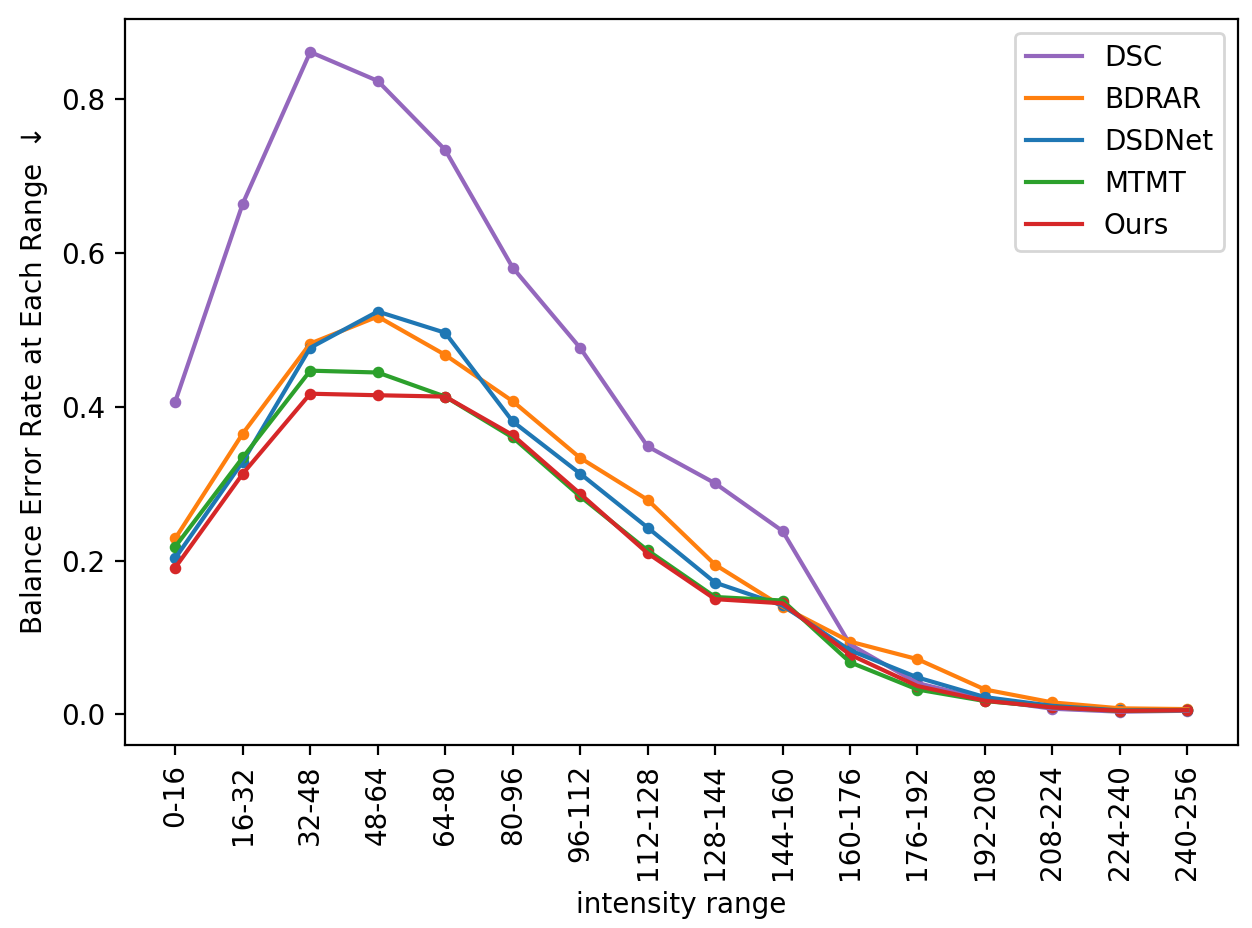}
\end{center}
\caption{\ghk{Distribution of BER scores (the smaller the better) of state-of-the-art shadow detection methods and Ours (red) on SBU~\cite{DBLP:conf/eccv/VicenteHYHS16}. It shows that dark ranges contain more errors than bright ranges. It also reveals that our method performs better at low-intensity ranges, \eg, intensity-32 to intensity-48, than state-of-the-art methods~\cite{DBLP:conf/cvpr/Chen0WW0H20,DBLP:conf/cvpr/ZhengQCL19,DBLP:conf/cvpr/Hu0F0H18,DBLP:conf/eccv/ZhuDHFXQH18}. 
}}
\label{fig:introduction}
\end{figure}

\section{Introduction}
Shadows are ubiquitous in natural scenes. While shadows can help provide information of various scene properties, \eg, light source conditions~\cite{ICCV.2009.5459163}, environmental illuminations~\cite{CVPR.2009.5206665,CVPR.2011.5995585,TPAMI.2012.110}, and object shape \cite{ICCV.2009.5459381,2070781.2024191}, the presence of shadows can degrade the performance of many existing computer vision methods (\eg, object detection \cite{TPAMI.2003.1233909} and visual tracking \cite{TPAMI.2004.51}) by attenuating object colors and textures. Hence, the shadow detection task is important.

Conventional methods detect shadows by using physical models with low-level features, \eg, color and illumination \cite{DBLP:journals/ijcv/FinlaysonDL09,DBLP:journals/pami/FinlaysonHLD06}, or by extracting hand-crafted features with data-driven approaches~\cite{DBLP:conf/iccv/HuangHTW11,CVPR.2010.5540209}. However, these methods typically fail to capture sufficient global contextual information, which limits their shadow detection performances on complex real-world scenes.

Deep learning based shadow detection methods~\cite{DBLP:conf/eccv/ZhuDHFXQH18,DBLP:conf/eccv/LeVNHS18,DBLP:conf/cvpr/Chen0WW0H20,DBLP:conf/cvpr/ZhengQCL19,DBLP:conf/cvpr/Hu0F0H18,DBLP:conf/cvpr/WangL018,DBLP:conf/eccv/VicenteHYHS16} are becoming very popular in recent years. 
They typically use CNNs to exploit global contextual information for learning shadow features, and achieve impressive performance improvements over the conventional methods.
However, we have observed that these deep methods suffer from a common problem: they tend to have higher error rates in dark (or low-intensity) regions. To verify this observation, we carry out an experiment to measure the balance error rates at different intensity ranges on SBU~\cite{DBLP:conf/eccv/VicenteHYHS16} dataset. As shown in Figure \ref{fig:introduction}, dark ranges contain more errors than bright ranges, suggesting that dark regions are more error-prone.
%
This happens to all state-of-the-art methods that we have tested. In particular, as DSC~\cite{DBLP:conf/cvpr/Hu0F0H18} relies heavily on the global contextual cues to detect shadows, it has the largest balance error rate at low- and medium-intensity ranges.
Our insight from this experiment is that existing shadow detection methods typically learn discriminative features from the whole image globally, covering the full range of intensity values. This weakens the models' ability in learning the subtle differences between shadow and non-shadow pixels in dark regions.
Based on this insight, in this paper, we propose a dark-region aware approach by delving into the error-prone dark regions so as to be able to extract more discriminative features for robust shadow detection. To this end, we have developed a novel shadow detection method with three main parts. First, we formulate a global context network aiming to learn global contextual cues from the entire image. Second, we propose a dark-region recommendation (DRR) module to recommend error-prone dark regions for the network to focus on. Third, we propose a novel dark-aware shadow analysis (DASA) module. The DASA module will focus on the recommended dark regions and tries to learn robust dark-aware shadow features. 
The key advantage of our design here is that as the DASA module would only focus on the dark regions, it becomes much sensitive to the dark pixels and is therefore able to capture the subtle differences between shadow and non-shadow pixels inside the error-prone dark regions for more accurate shadow detection.
As shown in Figure~\ref{fig:introduction}, the proposed method performs better in low-intensity ranges, {\it \eg} intensity-32 to intensity-48, comparing to state-of-the-art shadow detection methods.
Extensive experimental results on three popular shadow detection datasets verify its effectiveness over the state-of-the-art methods.

In summary, this paper makes three key contributions:
\begin{itemize}
\item We propose a novel shadow detection approach based on a dark-region focusing mechanism. The proposed method first acquires global contexts over the entire image, and then takes a closer look at the error-prone dark regions for learning robust dark-aware shadow features.


\item We propose a simple yet effective dark-region recommendation (DRR) module to recommend error-prone dark regions, and a novel dark-aware shadow analysis (DASA) module to learn the robust dark-aware shadow features by focusing on the recommended dark regions.



\item We extensively evaluate the proposed method on three popular shadow detection datasets and show its superior performances over the state-of-the-art methods.

\end{itemize}

\begin{figure*}[t]
\begin{center}
\begin{tabular}{c}
	\includegraphics[width=\linewidth]{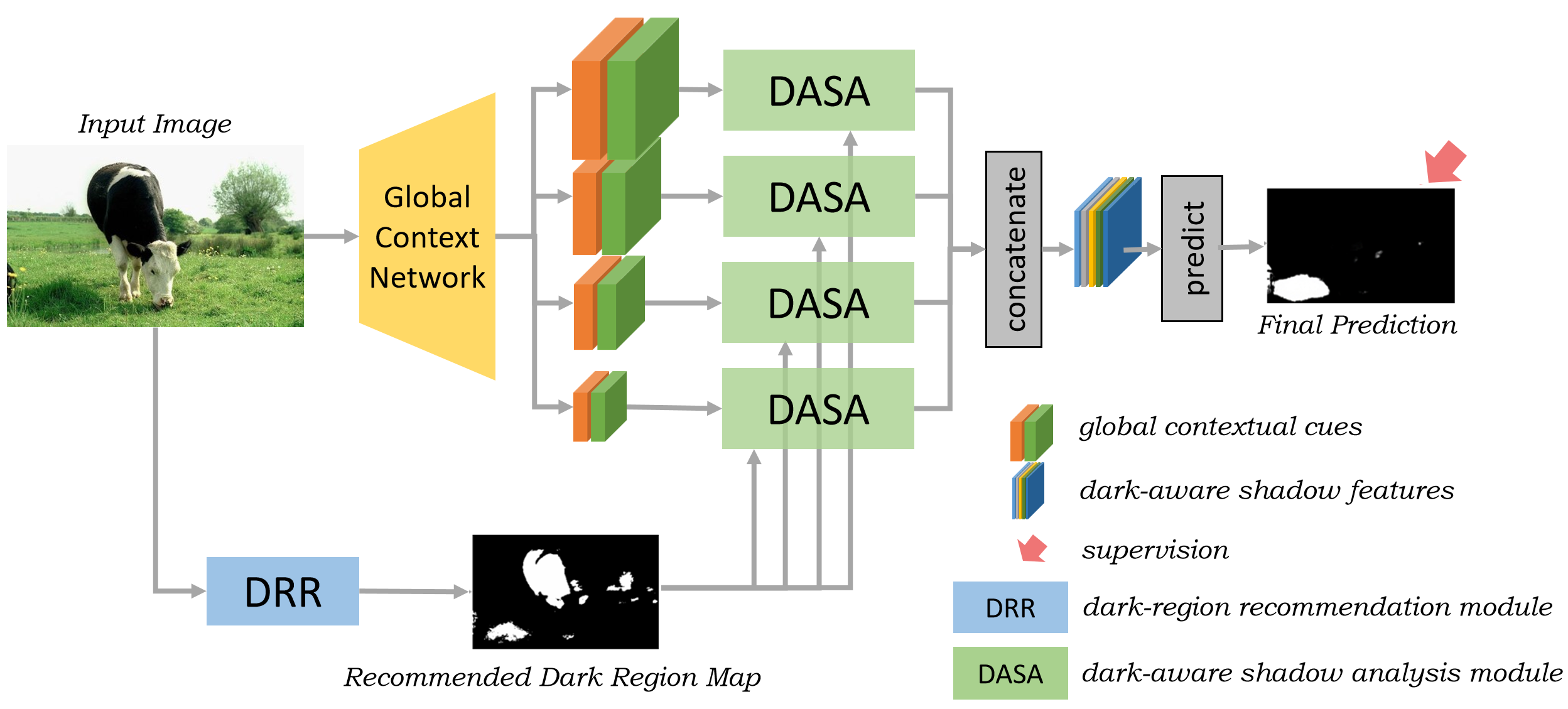}
\end{tabular}
\end{center}
\vspace{-3mm}
\caption{\ghk{Architecture of the proposed shadow detection framework. First, the global context network extracts multi-scale global contextual cues from the entire input image. Second, the dark-region recommendation (DRR) module recommends dark regions from the input image and forwards them to all dark-aware shadow analysis (DASA) modules. Third, the DASA modules at different layers take the corresponding global contextual cues and the recommended dark regions to produce dark-aware shadow features. Finally, the dark-aware shadow features across different layers are concatenated to predict the final shadow mask.}}
\label{fig:overview}
\end{figure*}

\section{Related Work}

\noindent
{\bf {Conventional methods.}} Early works~\cite{CVPR.2009.5206665,DBLP:journals/pr/TianQQT16,DBLP:journals/ijcv/FinlaysonDL09} primarily propose physical property (\eg, illumination and colors) based models for detecting shadows. However, these models may fail on real-world complex scenes, where the diverse shadow patterns may not fit these physical models.
Later, many data-driven approaches with hand-crafted shadow features~\cite{DBLP:conf/cvpr/GuoDH11,DBLP:conf/iccv/HuangHTW11,DBLP:conf/eccv/LalondeEN10,CVPR.2010.5540209,DBLP:conf/iccv/VicenteHS15} are proposed.
These methods relax the restrictions of the physical model based methods, but their hand-crafted features are still not sufficient for representing real-world shadow patterns.
In addition, these methods typically suffer from a common problem that there is insufficient global information incorporated into the features, which limits their performances in detecting shadows from complex real-world scenes.
\vspace{1mm}

\noindent
{\bf {Deep-learning based methods.}} Recently, deep convolutional neural networks (CNN) have shown their advantages on extracting discriminative features and learning global contextual information~\cite{DBLP:journals/corr/SimonyanZ14a,DBLP:journals/pami/ChenPKMY18,DBLP:journals/corr/HeZRS15}.
Benefited from the development of deep learning techniques, many deep shadow detectors are proposed.
Khan \etal \cite{DBLP:conf/cvpr/KhanBST14} propose the first deep method that utilizes a 7-layer CNN to learn the shadow features and uses the conditional random field (CRF) model to refine the shadow maps. Nguyen \etal \cite{DBLP:conf/iccv/NguyenVZHS17} propose a conditional Generative Adversarial Network (cGAN) to condition the shadow detection process on the input image as well as an additional parameter that controls the network sensitivity to shadows. Le \etal \cite{DBLP:conf/eccv/LeVNHS18} propose a GAN-based method for detecting shadows, in which a shadow attenuation network is incorporated during training to generate adversarial training examples.
Zhu \etal \cite{DBLP:conf/eccv/ZhuDHFXQH18} propose a recurrent attention residual (RAR) module to combine the contextual information in every two adjacent CNN layers, and further aggregate these information in both top-down and bottom-up manners.
%
%
Hu \etal \cite{DBLP:conf/cvpr/Hu0F0H18} propose to learn global contextual information for shadow detection in a direction-aware manner, by running a spatial RNN in four directions with two rounds.
\ghk{
Ding~\etal~\cite{DBLP:conf/iccv/DingLZX19} propose an Attentive Recurrent Generative Adversarial Network (ARGAN) for joint shadow detection and removal.
Cun~\etal~\cite{cun2019ghostfree} propose a Dual Hierarchical Aggregation Network (DHAN), aiming to produce realistic shadow-free images and shadow mattes jointly. DHAN~\cite{cun2019ghostfree} also employs a generative adversarial network to augment the shadow images for training.
}
Zheng \etal \cite{DBLP:conf/cvpr/ZhengQCL19} introduce the concept of distraction to the shadow detection task, and construct a distraction dataset of false predictions from previous methods for their proposed network to learn distraction-aware shadow features.
%
%
Chen \etal \cite{DBLP:conf/cvpr/Chen0WW0H20} propose a semi-supervised method to improve shadow detection performances by leveraging extra unlabeled shadow images.
They also explore shadow count and edge labels as multi-task learning to boost shadow detection performances.
Zhu~\etal~\cite{zhu2021mitigating} propose a shadow feature decomposition and reweighting method to re-balance intensity-variant and intensity-invariant features for shadow detection.

Although deep-learning based shadow detection methods are effective, as demonstrated in our initial experiment shown in Figure~\ref{fig:introduction}, these methods typically have higher error rates in the dark regions due to their limited ability in differentiate shadow and non-shadow pixels in these regions. To address this problem, in addition to global understanding, we propose in this paper that the network should also take a closer look at the error-prone dark regions to learn the subtle contrast between shadow and non-shadow pixels in these regions, which has not been done before. Experimental results show that our dark-region focusing approach plays favorably against existing methods.

\section{Proposed Method}
Our work in this paper begins by observing that state-of-the-art shadow detection methods typically have higher error rates in the darker regions of an image. Our preliminary experiment shown in Figure \ref{fig:introduction} verifies our observation and reveals that the distribution of the incorrect predictions is hugely uneven. The probability of making an incorrect prediction in a low-intensity region is much higher than that in a medium- or high-intensity region. Our insight into this problem is that previous methods typically learn discriminative features under global illumination conditions over the entire image, covering the full range of intensity values, which may bury the subtle contrasts between shadow and non-shadow pixels inside the dark regions. To address this limitation, in addition to the global understanding, we propose to take a closer look at the error-prone dark regions to enhance the model's sensitivity to subtle contrasts among the dark pixels for more robust shadow detection.

Based on the above idea, we design our proposed model with three main parts: the global context network, dark-region recommendation (DRR) module and dark-aware shadow analysis (DASA) module. Figure \ref{fig:overview} presents the overall architecture of the proposed shadow detection framework. First, we construct a global context network to extract multi-scale global contextual cues from the entire image. Second, we embed DASA modules at different layers to learn dark-aware shadow features by delving into the dark regions recommended by our DRR module.
Third, we concatenate the dark-aware shadow features across different layers to predict the final shadow mask.

We organize the rest of this section as follows. We describe the global context network in Section~\ref{sec:gcn}, the DRR module in Section~\ref{sec:lpm}, the DASA module in Section~\ref{sec:zim}, and the training strategy in Section~\ref{sec:los}.

\begin{figure}[t]
\begin{center}
\begin{tabular}{c}
	\includegraphics[width=\linewidth]{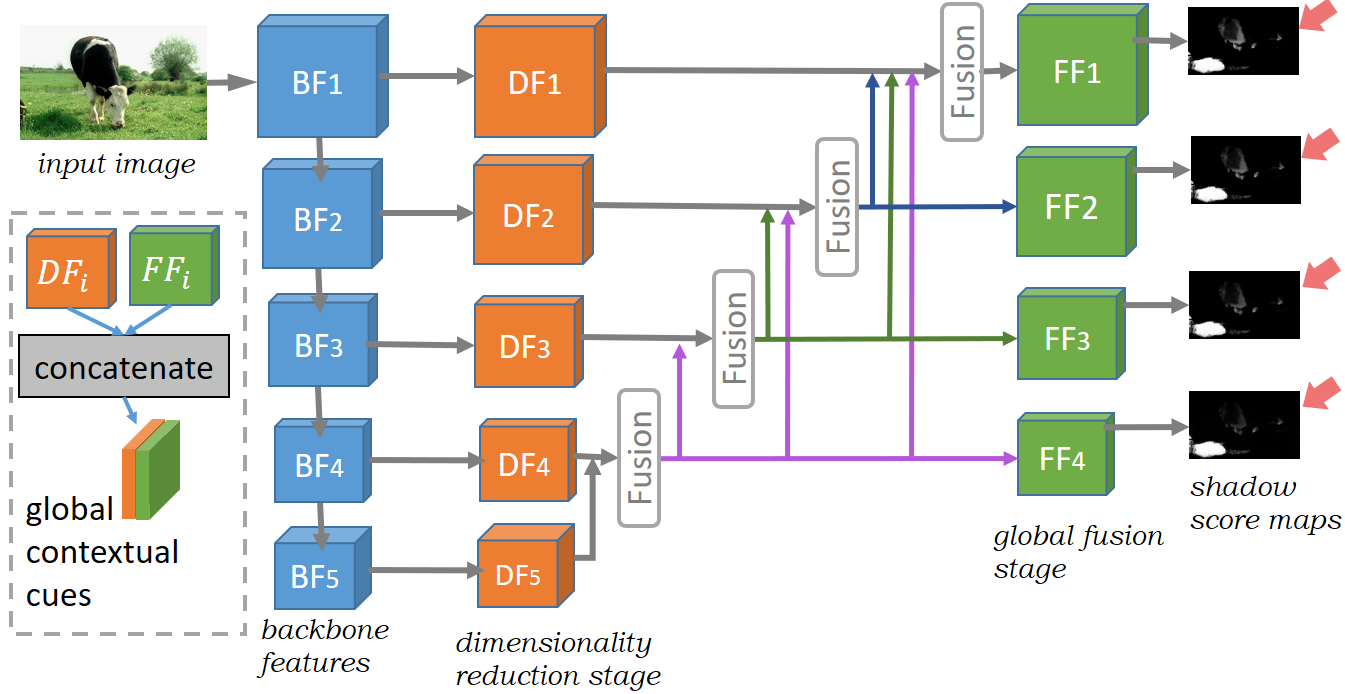}
\end{tabular}
\end{center}
\caption{Global Context Network. We choose ResNeXt-101~\cite{DBLP:conf/cvpr/XieGDTH17} as the backbone to extract features from the input image. To reduce memory footprint, we perform a dimensionality reduction on the backbone features to obtain $DF_i$, $i\in \{1,2,3,4,5\}$. We then employ a series of short connections~\cite{DBLP:journals/pami/HouCHBTT19} to connect the outputs of the deeper fusion blocks to the shallower fusion blocks in the fusion stage. \ghk{We further adopt a deep supervision strategy on the fusion features $FF_i$, $i\in \{1,2,3,4\}$, to accelerate the convergence. Finally, we concatenate $DF_i$ and $FF_i$ in each layer ($i\in \{1,2,3,4\}$) as our global contextual cues for the DASA modules.}}
\label{fig:gcn}
\end{figure}

\subsection{Global Context Network}
\label{sec:gcn}
The Global Context Network is designed to extract multi-scale global contextual cues under global illumination conditions over the entire image. Figure~\ref{fig:gcn} shows its architecture. 

Following previous methods, we adopt multi-scale feature extraction, fusion, and then a deep supervision strategy in our global context network. Specifically, we choose ResNeXt-101~\cite{DBLP:conf/cvpr/XieGDTH17} as the backbone to extract multi-scale backbone features, $BF_1, BF_2, BF_3, BF_4$ and $BF_5$. To reduce the memory footprint, we perform dimensionality reduction on these extracted backbone features, by feeding them into two successive convolutional layers (with $3\times3$ and $1\times1$ kernels) followed by batch normalization layers and rectified linear units. We denote the feature maps after dimensionality reduction as $DF_i$, where $i\,\in\,\{1,2,3,4,5\}$ (the orange blocks in Figure~\ref{fig:gcn}).

\begin{figure}[t]
\begin{center}
\begin{tabular}{cc}
	\includegraphics[width=0.45\linewidth]{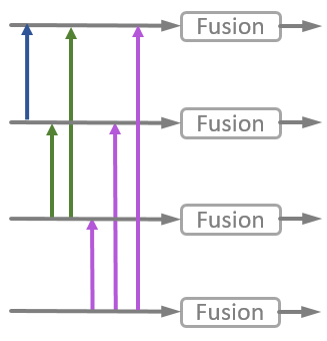} &
	\includegraphics[width=0.45\linewidth]{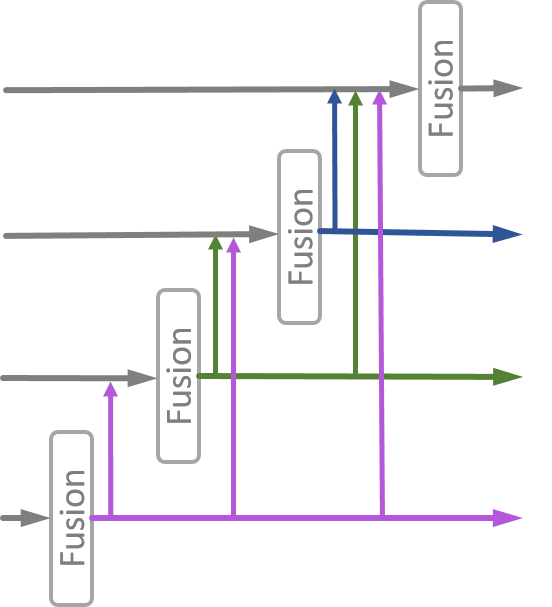} \\
	(a) Fusion in~\cite{DBLP:conf/cvpr/Chen0WW0H20,DBLP:conf/cvpr/ZhengQCL19} & (b) Ours
\end{tabular}
\end{center}
\caption{Fusion strategy comparison. Our fusion strategy allows more high level semantics of upper-stream features being exploited by down-stream features.}
\label{fig:fusestrategy}
\end{figure}

To help acquire a better global understanding and enrich the feature maps of the shallower layers with high-level semantics, we adopt a series of short connections~\cite{DBLP:journals/pami/HouCHBTT19} to connect the outputs of the deeper fusion blocks to shallower fusion blocks. Different from previous works~\cite{DBLP:conf/cvpr/Chen0WW0H20,DBLP:conf/cvpr/ZhengQCL19}, as shown in Figure~\ref{fig:fusestrategy}, we first perform the fusion operation then apply the short connections. This benefits downstream features with richer high-level semantics as more upper-stream features can be backtracked. Finally, we upscale the fusion features $FF_i$, where $i\,\in\,\{1,2,3,4\}$, to the size of the input image and adopt a deep supervision strategy to impose the supervision signal to each layer. In each layer, we concatenate $DF_i$ and $FF_i$, where $i \in \{1, 2, 3, 4\}$, as our global contextual cues and forward them to the DASA modules.



In this way, the Global Context Network is formulated to harvest global contextual information for shadow detection. However, relying solely on global understanding of the input image is not good enough for robust shadow detection. For example, DSC~\cite{DBLP:conf/cvpr/Hu0F0H18} relies heavily on global understanding, and yields a high error rate at low-intensity regions (see Figure~\ref{fig:introduction}). Hence, in addition to the global understanding, we specially design two novel modules, DRR and DASA, to learn robust shadow features from dark regions.


\subsection{Dark-Region Recommendation (DRR) Module}
\label{sec:lpm}

If a network is trained under global illumination conditions, it may not be sensitive enough to detect the subtle contrasts between shadow and non-shadow pixels inside a dark region, causing in a higher error rate in this region. Thus, we propose the DRR module to highlight the error-prone dark regions, such that our network can focus on them in order to learn to capture such subtle contrast information.

\begin{figure}[t]
\begin{center}
\begin{tabular}{c}
\includegraphics[width=\linewidth]{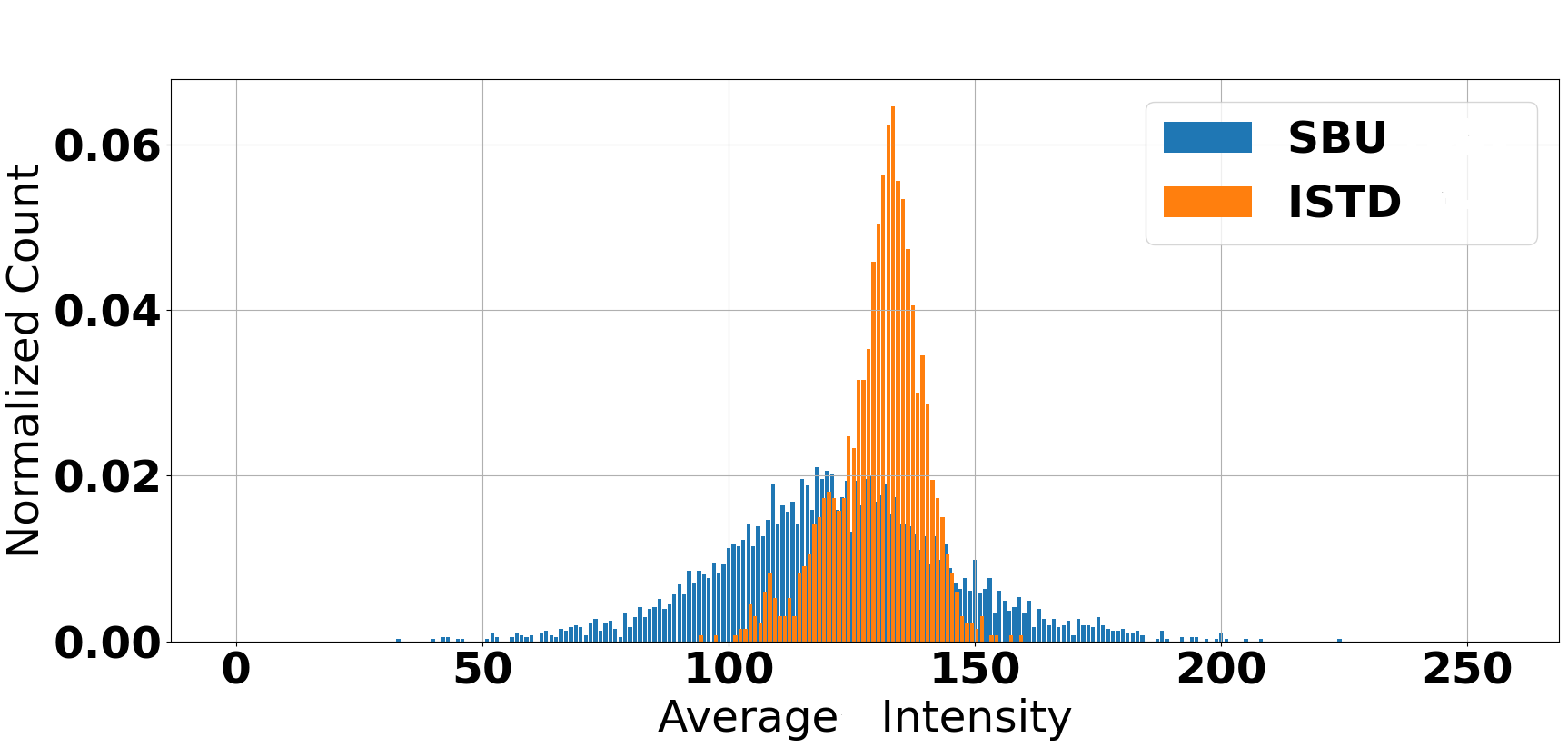} \\
\end{tabular}
\end{center}
\caption{Image-level average intensity histograms of two popular datasets, SBU and ISTD.}
\label{fig:pdf}
\end{figure}

To identify the error-prone dark regions, a straightforward solution is to apply a threshold policy on the intensity maps. Unfortunately, a universally ``appropriate'' threshold value does not exist, as the intensity distribution varies from image to image. For example, as shown in Figure~\ref{fig:pdf}, the images in ISTD~\cite{DBLP:conf/cvpr/WangL018} are often brighter than those in the SBU~\cite{DBLP:conf/eccv/VicenteHYHS16} dataset. Further, the average intensity distribution of SBU covers a wide range because of the complex illumination conditions. However, a threshold policy can only handle simple scenes captured under well-lit conditions.

\begin{figure}[t]
\begin{center}
\begin{tabular}{c}
	\includegraphics[width=\linewidth]{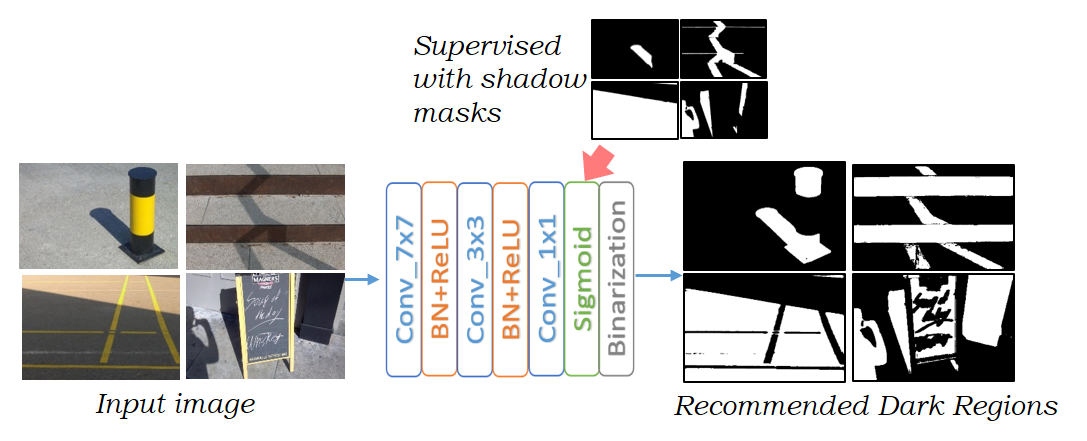}
\end{tabular}
\end{center}
\caption{The dark-region recommendation (DRR) module. We embed this lightweight module into our shadow detection framework and supervise it with paired data (shadow images and shadow masks).}
\label{fig:lrp}
\end{figure}

\ghk{Instead of applying a threshold policy here, we propose a simple yet effective approach to identify dark regions from paired data (shadow images and shadow masks). Since a shadow is formed when light fails to reach a surface, shadow regions are usually darker compared to well-lit regions. If we train a shallow network to detect shadows, due to the lack of global contextual information, a shallow shadow detector tends to detect all dark regions as shadow regions. We leverage this phenomenon to use the predictions from our DRR module as dark regions.
As shown in Figure~\ref{fig:lrp}, our lightweight DRR module consists of three convolutional layers. Our preliminary statistics show that the average intensity of the detected dark regions is 72/92 (out of 255) while the average intensity of the non-dark regions is 145/147 in SBU/ISTD. Besides, the recommended dark regions include about 27.6\% of the pixels over the entire SBU dataset. By applying DSDNet~\cite{DBLP:conf/cvpr/ZhengQCL19} and MTMT~\cite{DBLP:conf/cvpr/Chen0WW0H20} on SBU, we find that 65.97\% and 64.68\%, respectively, of detection errors by these two detectors occur within this 27.6\% of pixels. This means that the dark regions recommended by our DRR module are error-prone, which \ryn{justifies} our observation that existing methods tend to make detection mistake in dark regions.}
\ryn{Section~\ref{sec:abla} verifies the effectiveness of our dark-region recommendation strategy}.

\begin{figure}[t]
\begin{center}
\begin{tabular}{c}
	\includegraphics[width=\linewidth]{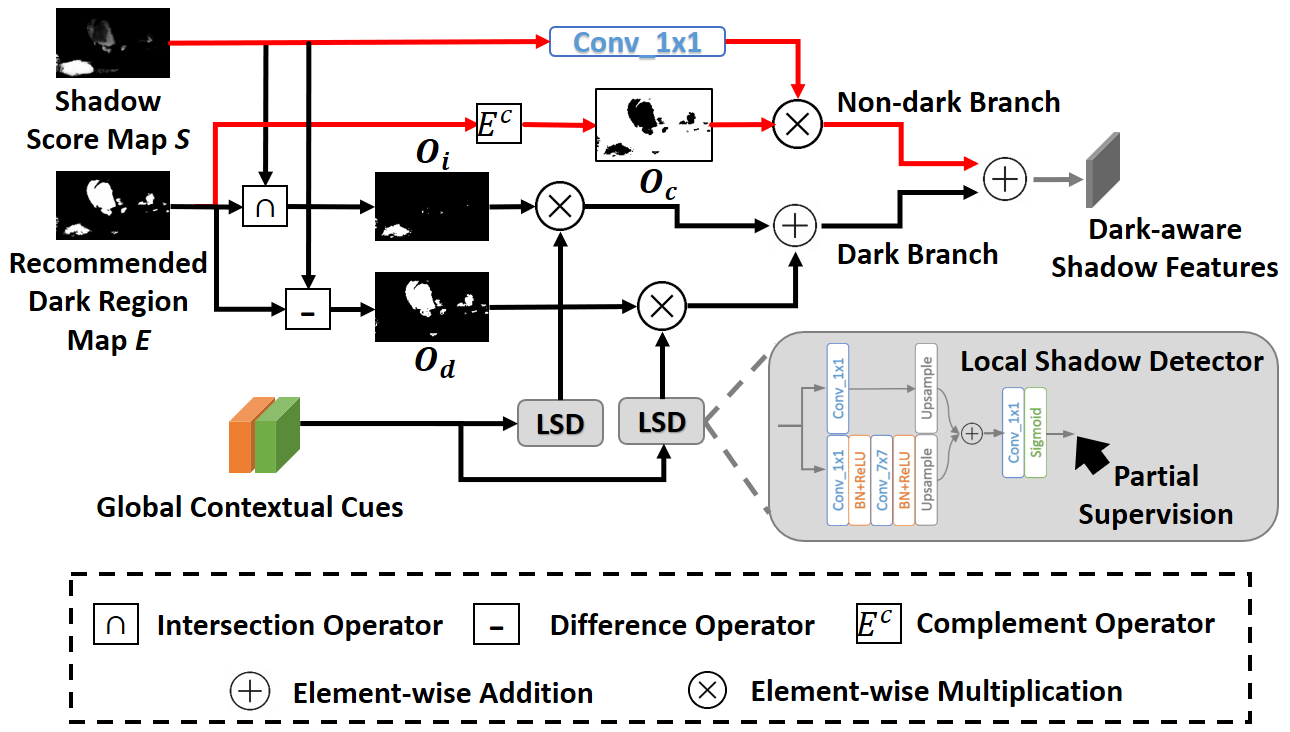}
\end{tabular}
\end{center}
\caption{The dark-aware shadow analysis (DASA) module with two branches. The dark branch learns the shadow features from the recommended dark regions, while the non-dark branch learns the shadow features from outside of the recommended dark regions. The two types of shadow features are then merged to form the dark-aware shadow features.}
\label{fig:darkaware}
\end{figure}

\subsection{Dark-Aware Shadow Analysis (DASA) Module}
\label{sec:zim}
Our aim of the DASA module is to learn dark-aware robust shadow features by focusing its attention to the error-prone dark regions as recommended by our DRR module. As shown in Figure~\ref{fig:darkaware}, the DASA module takes the global contextual cues and the shadow score map from the global context network, and the recommended dark region map from our DRR module as inputs, and outputs the dark-aware shadow features via two branches for final shadow prediction.

\noindent
{\bf The Dark Branch.} To learn the subtle contrast differences between shadow/non-shadow pixels in the recommended dark regions, we propose to leverage the shadow score maps from the global contextual network to further divide these error-prone dark regions based on the global shadow confidence scores. We apply the intersection and difference operators to exploit the pixel co-occurrence and divergence between the input shadow score map $S$ and the recommended dark region map $E$. Specifically, the intersection operator takes $S$ and $E$ as inputs, and outputs an intersection map $O_i~=~E\cap S$, which helps locate the pixels that are more likely to be shadow pixels inside the dark regions (referred to as \emph{dark shadow pixels}).
The difference operator also takes $S$ and $E$ as inputs, but outputs a difference map {$O_d~=E-S$}, 
which helps locate the pixels that are more likely to be non-shadow pixels inside the dark regions (referred to as \emph{dark non-shadow pixels}).
%
In this way, these two operators roughly separate the recommended dark regions into two types of pixels, one is more likely to be in shadow while the other is unlikely to be in shadow.
%
We use $O_i$ and $O_d$ as masks to provide partial supervisions to the local shadow detectors (LSDs). By penalizing the detection errors only inside $O_i$, one LSD focuses on learning the representation of dark shadow pixels. Similarly, by penalizing the detection errors inside $O_d$, the other LSD learns the representation of dark non-shadow pixels. The two representations are then merged to form the dark branch.

\noindent
{\bf Local Shadow Detectors (LSDs).} We construct two LSDs to perform fine-grained dark-aware shadow feature analysis in the dark branch. Each LSD takes the global contextual cues as input to detect shadows under partial supervisions masked by $O_i$ and $O_d$. As shown in the bottom-right region of Figure~\ref{fig:darkaware}, each LSD consists only of a few convolutional layers. The key design is that we split the detector into two paths. While the top (shallow) path focuses on the current pixel, the bottom path aggregates the surrounding contextual information, so that the detector may learn the subtle contrast information between the current pixel and its surroundings for detecting shadow pixels.

\noindent
{\bf The Non-dark Branch.} In addition to the dark branch,
we also construct a non-dark branch for restoring relatively bright shadow pixels from the non-dark regions. We first apply the complement operator to take the binarized recommended dark region map $E$ as input, to produce an inverse map $O_c~=1-E$. Since our previous study shows that the detection errors in the non-dark regions are far lower than those in the dark regions, to save computational costs, we only use one convolutional layer to transform the shadow score map $S$ into shadow features and then apply the inverse map $O_c$ to select the shadow features in the non-dark regions, to form the non-dark branch. Finally, the non-dark branch features and the dark branch features are merged to produce the dark-aware shadow features.

\subsection{Loss Function}
\label{sec:los}

Since shadow detection is typically a binary classification task at pixel-level, and the numbers of shadow and non-shadow pixels are uneven in real-world scenes, we adopt the weighted binary cross-entropy loss to detect shadow pixels, which has the following form:
\begin{equation}
Loss = \sum_{i \in R} -w_{1} \cdot y_i \cdot log(p_i) - w_{2} \cdot (1 - y_i) \cdot log(1-p_i),
\label{eq:loss}
\end{equation}
where $R$ is the region that we want to penalize the detection errors. $y_i$ is the value of the $i^{th}$ pixel in the binary ground-truth shadow mask, and $p_i \in [0,1]$ is the predicted probability of the $i^{th}$ pixel being a shadow pixel. $w_{1}$ and $w_{2}$ are two balancing hyper-parameters.

Since the Global Context Network, the DRR Module and the whole model require global image understanding, we set the penalized region $R$ in Eq.~\ref{eq:loss} to the whole input image. Accordingly, the balancing hyper-parameters $w_{1}$ and $w_{2}$ in Eq.~\ref{eq:loss} are set as follows:
\begin{equation}
w_1 = \frac{ N_n}{N_p+N_n}, \ \ \ w_2 = \frac{N_p}{N_p+N_n},
\label{eq:darkglobal}
\end{equation}
where $N_n$ and $N_p$ are the numbers of non-shadow and shadow pixels in the binary groundtruth shadow mask, respectively.

Training the proposed DASA module requires a different strategy. As mentioned in Section~\ref{sec:zim}, we have two local shadow detectors (LSDs) in the DASA module focusing on the dark shadow and dark non-shadow pixels. Hence, we set the penalized region $R$ in Eq.~\ref{eq:loss} to $O_i$ and $O_d$.
The balancing hyper-parameters $w_{1}$ and $w_{2}$ in Eq.~\ref{eq:loss} are then set to:

\begin{eqnarray}
w_1 = \frac{N_{ln}}{N_{ln}+N_{lp}} + \frac{N_n}{N_p+N_n}
\label{eq:darkglobal3}
\end{eqnarray}
\begin{eqnarray}
w_2 = \frac{N_{lp}}{N_{ln}+N_{lp}} + \frac{N_p}{N_p+N_n},
\label{eq:darkglobal4}
\end{eqnarray}
where $N_{ln}$, $N_{lp}$ are the numbers of non-shadow and shadow pixels inside the penalized region (\ie, $O_i$ or $O_d$), respectively.
Note that we still need to add the global terms $\frac{N_n}{N_p+N_n}$ and $\frac{N_p}{N_p+N_n}$ in Eq.~\ref{eq:darkglobal3}~and Eq.~\ref{eq:darkglobal4} to smooth the learning process. If we only consider the local terms $\frac{N_ln}{N_lp+N_ln}$ and $\frac{N_lp}{N_lp+N_ln}$, it would cause the training procedure to be very unstable due to the imbalanced numbers of shadow and non-shadow pixels in regions $O_i$ and $O_d$.


\begin{table*}[t]
\centering
\caption{\ghk{Quantitative comparison on shadow detection performances. Both balanced error rate (BER) and shadow/non-shadow error rates are shown. The best results are marked in \textbf{bold}.}}
\label{tab:ti}
\resizebox{\textwidth}{!}{
\begin{tabular}{cc|ccc|ccc|ccc}

\multicolumn{1}{l}{~} & \multicolumn{1}{l|}{~} & \multicolumn{3}{c|}{SBU}                                      & \multicolumn{3}{c|}{UCF}                                              & \multicolumn{3}{c}{ISTD}                                      \\
Method                & Year                   & BER$\downarrow$           & shadow$\downarrow$               & non-shadow$\downarrow$          & BER$\downarrow$                 & shadow$\downarrow$               & non-shadow$\downarrow$           & BER$\downarrow$          & shadow$\downarrow$               & non-shadow$\downarrow$     \\

\hline
\hline
SRM \cite{DBLP:conf/iccv/WangBZZL17}         & 2017  & 6.51  & 10.52 &{2.50}       & 12.51       & 21.41     & {3.60}     & 7.92   & 13.97      & {1.86}  \\
PSPNet \cite{DBLP:conf/cvpr/ZhaoSQWJ17}      & 2017  & 8.57  & -     & -           & 11.75       & -         & -          & 4.26   & 4.51       & 4.02   \\
DeshadowNet \cite{DBLP:conf/cvpr/QuTHTL17}   & 2017  & 6.96  & -     & -           & 8.92        & -         & -          & -      & -          & -    \\
ARGAN~\cite{DBLP:conf/iccv/DingLZX19} & 2019 & 3.09 & - & - & {\bf 3.76} & - & - & 2.01 & - & -\\
DHAN~\cite{cun2019ghostfree} & 2019 & 4.29 & {\bf 2.71} & 5.87 & - & - & - & - & - & - \\
EGNet \cite{DBLP:conf/iccv/ZhaoLFCYC19}      & 2019  & 4.49  & 5.23  & 3.75        & 9.20        & 11.28     & 7.12       & 1.85   & 1.75       & 1.95 \\
\hline
Unary-Pairwise \cite{DBLP:conf/cvpr/GuoDH11} & 2011  & 25.03 & 36.26 & 13.80       & -           & -         & -          & -      & -          & -   \\
scGAN \cite{DBLP:conf/iccv/NguyenVZHS17}     & 2017  & 9.04  & 8.39  & 9.69        & 11.52       & {7.74}    & 15.30      & 4.70   & 3.22       & 6.18   \\
ST-CGAN \cite{DBLP:conf/cvpr/WangL018}       & 2018  & 8.14  & 3.75  & 12.53       & 11.23       & {\bf 4.94}& 17.52      & 3.85   & 2.14       & 5.55  \\
ADNet \cite{DBLP:conf/eccv/LeVNHS18}         & 2018  & 5.37  & 4.45  & 6.30        & 9.25        & 8.37      & 10.14      & -      & -          & -   \\
BDRAR \cite{DBLP:conf/eccv/ZhuDHFXQH18}      & 2018  & 3.64  & 3.41  & 3.88        & 7.81        & 9.69      & 5.94       & 2.69   & {\bf0.50}  & 4.87   \\
DSC \cite{DBLP:conf/cvpr/Hu0F0H18}           & 2018  & 5.59  & 9.77  & {\bf 1.42}  & 10.54       & 18.08     & {\bf 3.00} & 3.42   & 3.85       & 3.00   \\
DSDNet \cite{DBLP:conf/cvpr/ZhengQCL19}      & 2019  & 3.45  & {3.33}& 3.58        & {7.59}      & 9.74      & 5.44       & 2.17   & 1.36       & 2.98    \\
MTMT \cite{DBLP:conf/cvpr/Chen0WW0H20}       & 2020  & 3.15  & 3.73  & 2.57        & 7.51  & 10.37     & 4.64       & 1.72   & 1.36       & 2.08   \\ 
\hline
Ours                                         & 2021     & {\bf3.04} & 2.85  & 3.23   & 7.75     & 9.17    & 6.34    & {\bf1.33} & 0.85  & {\bf1.82}     \\
\hline
\end{tabular}
}
\end{table*}

The overall training loss function for our model is then formulated as:
\begin{equation}
L_{overall} = \alpha \cdot L_{DRR} + \beta \cdot L_{GCN} + \gamma \cdot L_{DASA} + \theta \cdot L_{final},
\label{eq:overallloss}
\end{equation}
where $L_{GCN}$, $L_{DRR}$, $L_{DASA}$, and $L_{Final}$ are the weighted binary cross-entropy loss for the proposed global context network, DRR module, DASA module and the final prediction, respectively. $\alpha$, $\beta$, $\gamma$, and $\theta$ are four balancing hyper-parameters, which are empirically set to 1.



\section{Experiments}
\subsection{Experimental Settings}

{\bf Implementation details.} We have implemented the proposed network in the Pytorch framework \cite{paszke2017automatic}, and tested it on a PC with an i7 3.7GHz CPU and a RTX2080ti GPU. \ghk{We initialize the network parameters randomly, except for the backbone network (see Figure~\ref{fig:gcn}), which we follow previous shadow detection methods~\cite{DBLP:conf/cvpr/Chen0WW0H20,DBLP:conf/cvpr/ZhengQCL19,DBLP:conf/eccv/ZhuDHFXQH18} to adopt ResNeXt-101~\cite{DBLP:conf/cvpr/XieGDTH17} pretrained on the ImageNet to extract image features.} We augment \ryn{the} training set with random horizontal flipping. We use stochastic gradient descent (SGD) with \ryn{momentum} 0.9 and weight decay $5 \times 10^{-4}$ for optimization. The initial learning rate is set to $5 \times 10^{-3}$, decreased by the polynomial strategy with a power of 0.9. Both the input images and labeled shadow masks are resized to $320 \times 320$ for training and inference. We train our network for 5,000 iterations with a batch size of 6. For inference, like previous methods~\cite{DBLP:conf/cvpr/Chen0WW0H20,DBLP:conf/cvpr/ZhengQCL19,DBLP:conf/eccv/ZhuDHFXQH18}, we apply CRF~\cite{DBLP:conf/nips/KrahenbuhlK11} as a post-process for refinement.


\noindent
{\bf Benchmark datasets.}
We evaluate our model on three popular shadow detection datasets. The first one is SBU \cite{DBLP:conf/eccv/VicenteHYHS16}, which consists of 4,089 training pairs (shadow images and shadow masks) and other 638 pairs for testing.
%
%
The second one is UCF \cite{CVPR.2010.5540209}, which contains only 245 pairs (shadow images and shadow masks) and 110 of them are used for testing.
The third dataset is ISTD \cite{DBLP:conf/cvpr/WangL018}, which consists of 1,870 triples (shadow images, shadow masks and shadow-free images) and 540 triples of them are used for testing. In this work, we only use the shadow images and shadow masks for shadow detection.

For SBU and UCF, we train our model on the SBU~\cite{DBLP:conf/eccv/VicenteHYHS16} training set and test on both SBU~\cite{DBLP:conf/eccv/VicenteHYHS16} and UCF~\cite{CVPR.2010.5540209} test sets, following~\cite{DBLP:conf/cvpr/Chen0WW0H20,DBLP:conf/cvpr/ZhengQCL19,DBLP:conf/cvpr/Hu0F0H18,DBLP:conf/eccv/ZhuDHFXQH18}. For ISTD, we train our model on its training set and test it on its test set, similar to previous method~\cite{DBLP:conf/cvpr/Chen0WW0H20,DBLP:conf/cvpr/ZhengQCL19}. 
%
%

\noindent
{\bf Evaluation metric.}
We adopt the commonly-used metric, balanced error rate (BER), to quantitatively evaluate the shadow detection performance. BER takes accuracy of both shadow region and non-shadow region into consideration. Mathematically, it is formulated as:
\begin{equation}
\centering
BER = (1.0 - \frac{1}{2}(\frac{TP}{N_p}+\frac{TN}{N_n})) \times 100,
\end{equation}
where TP, TN represent the numbers of true positive and true negative, respectively, and $N_p$, $N_n$ are the numbers of shadow pixels and non-shadow pixels. A lower BER score indicates a better performance.

\begin{figure*}[t]
\renewcommand{\tabcolsep}{1pt}
\begin{center}
\begin{tabular}{cccccccc}
	\includegraphics[width=0.124\linewidth]{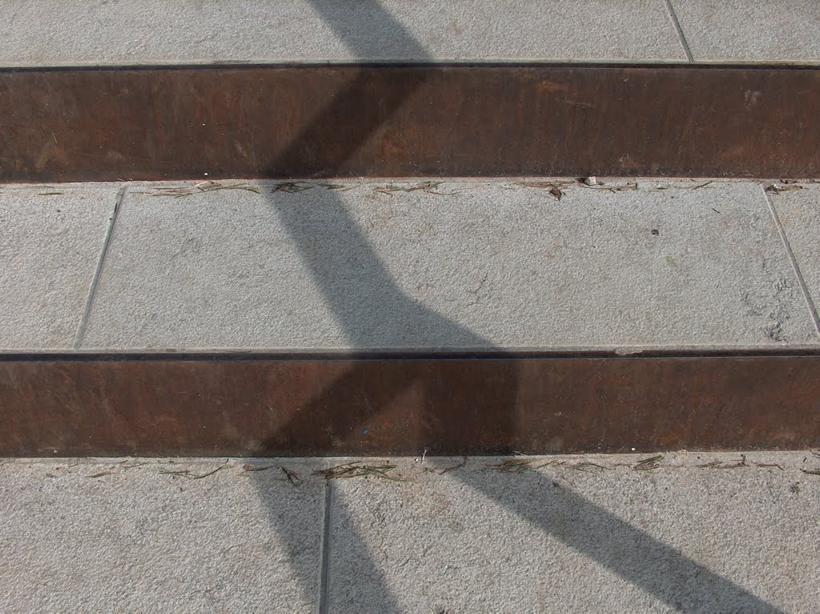} &
	\includegraphics[width=0.124\linewidth]{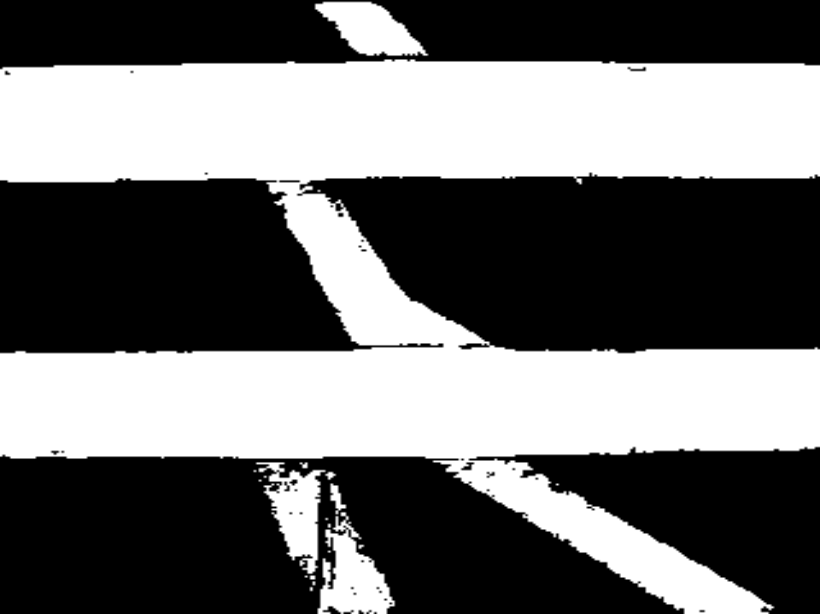} &
	\includegraphics[width=0.124\linewidth]{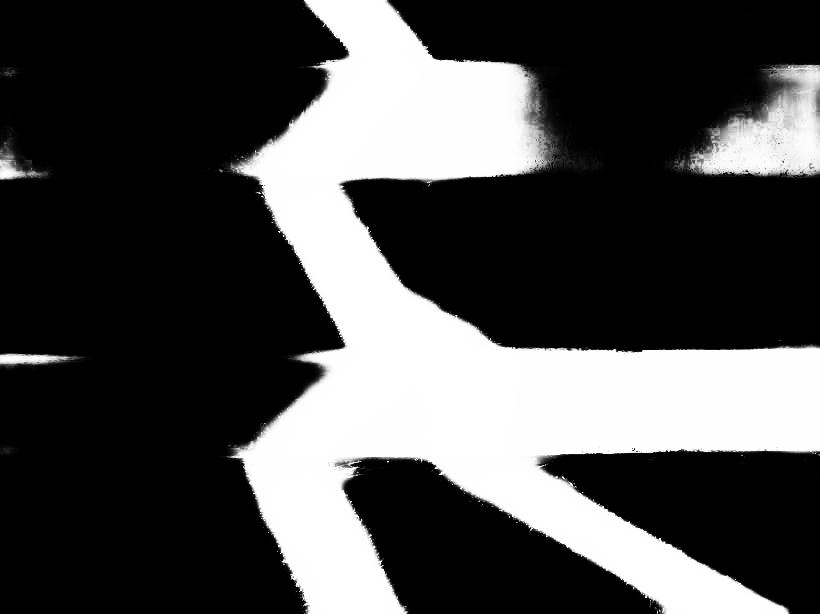} &
	\includegraphics[width=0.124\linewidth]{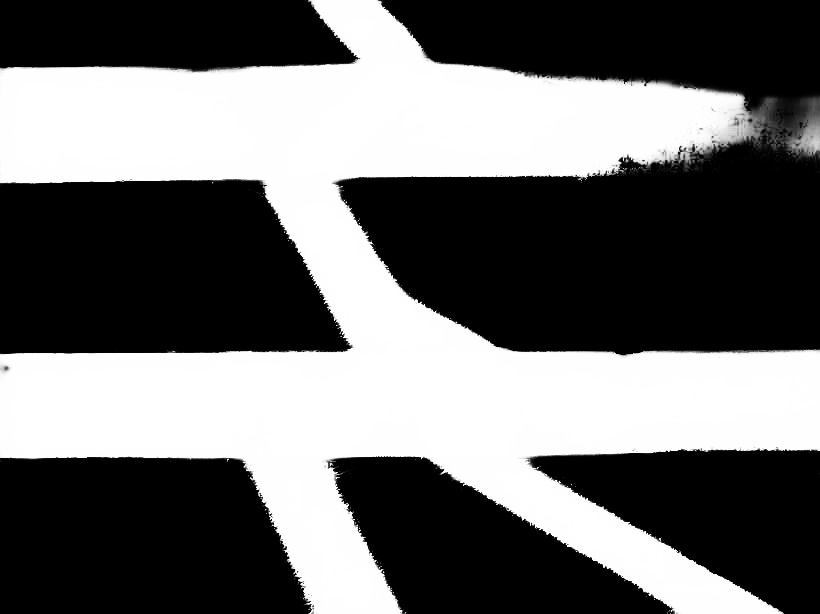} &
	\includegraphics[width=0.124\linewidth]{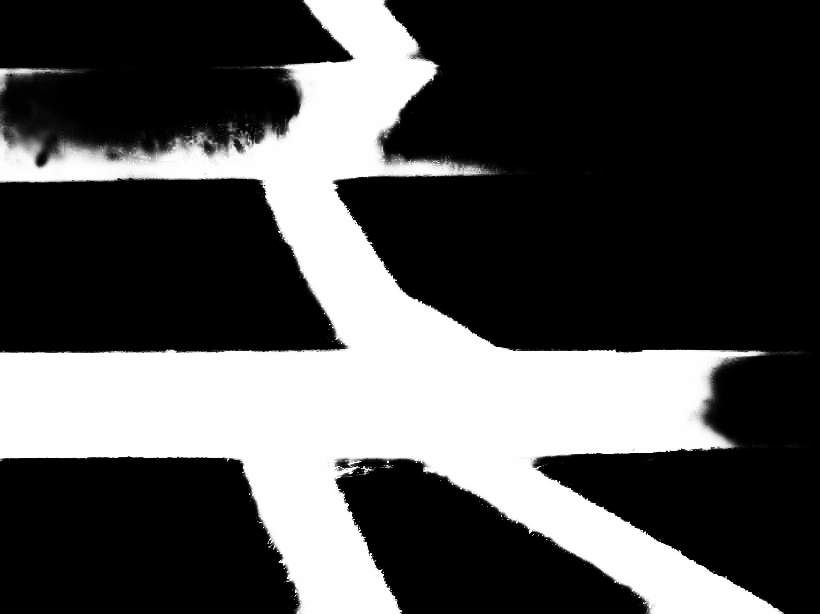} &
	\includegraphics[width=0.124\linewidth]{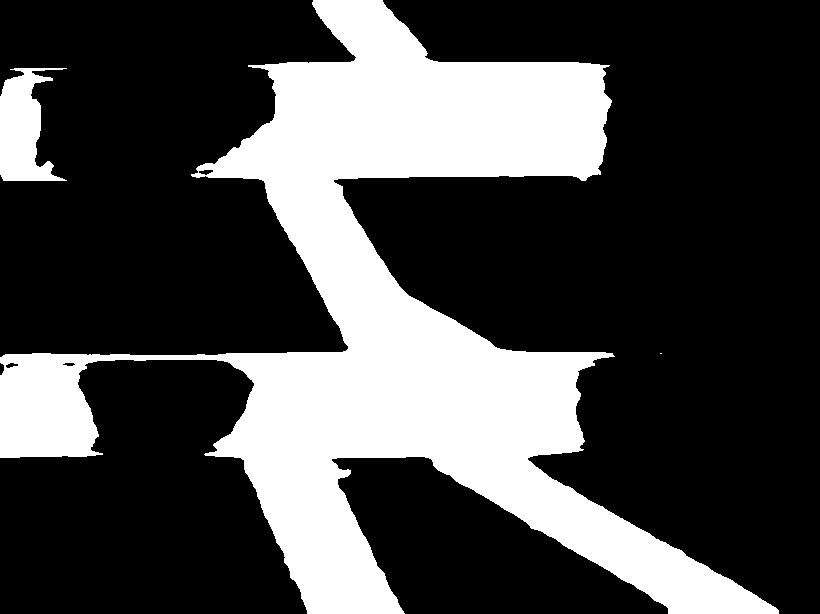} &
	\includegraphics[width=0.124\linewidth]{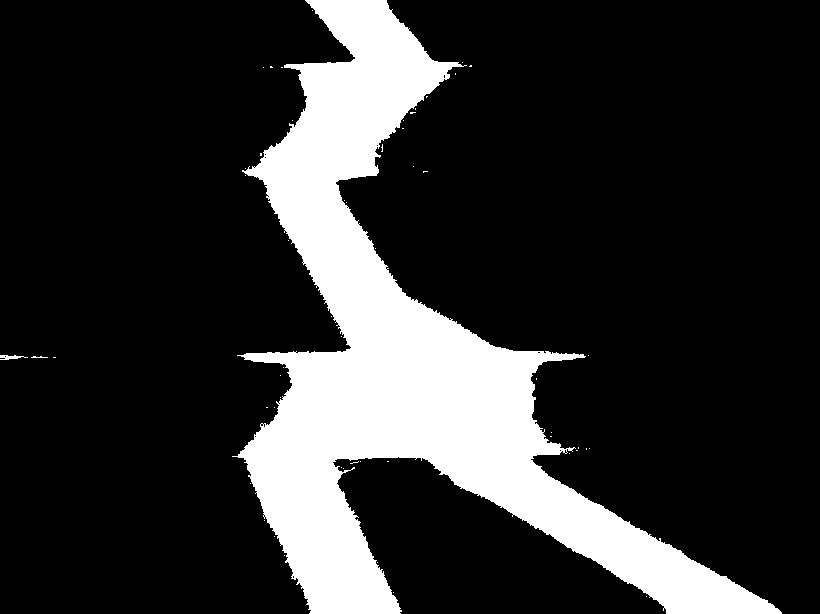} &
	\includegraphics[width=0.124\linewidth]{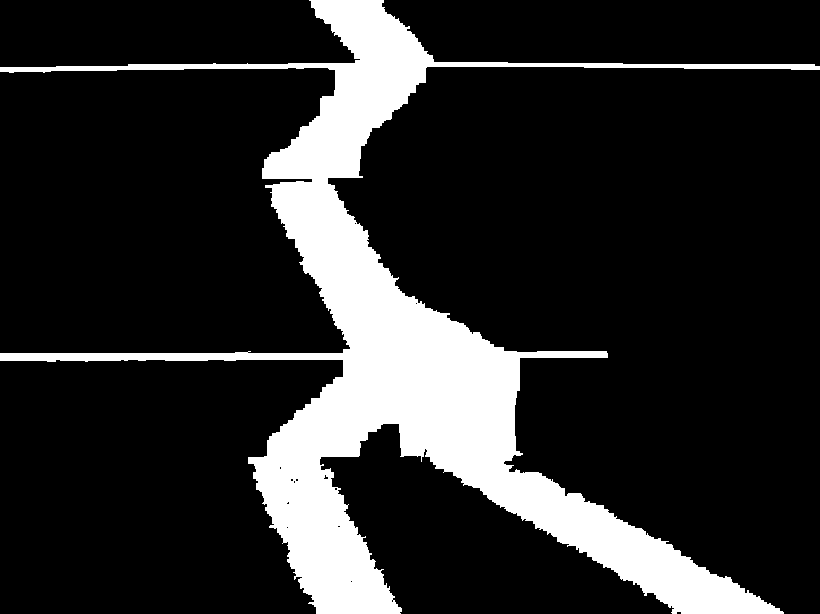} \\
	
	\includegraphics[width=0.124\linewidth]{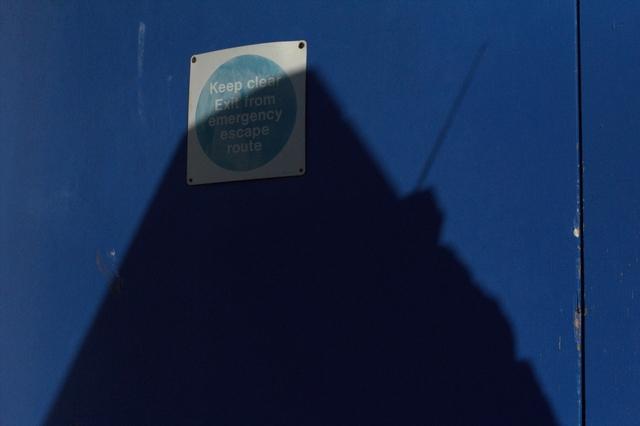} &
	\includegraphics[width=0.124\linewidth]{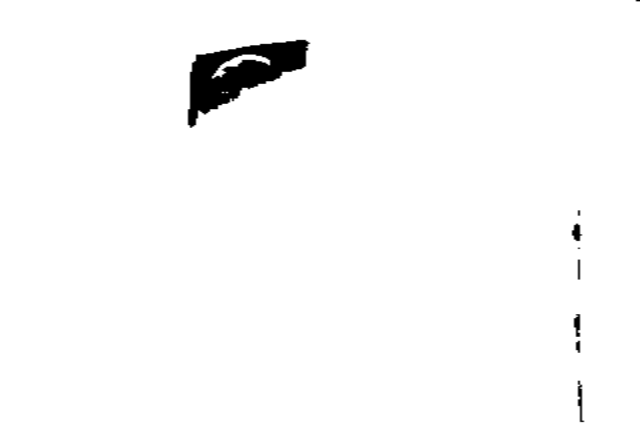} &
	\includegraphics[width=0.124\linewidth]{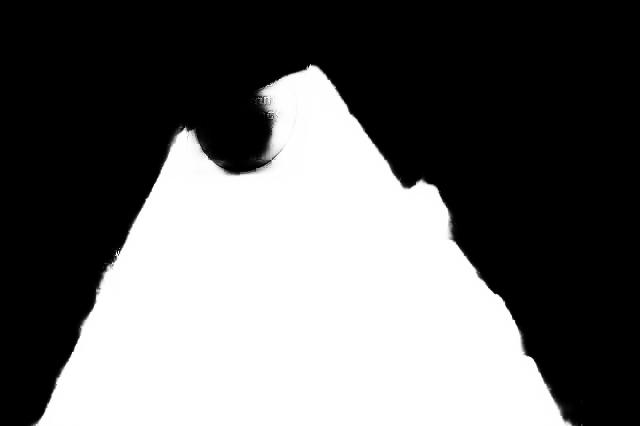} &
	\includegraphics[width=0.124\linewidth]{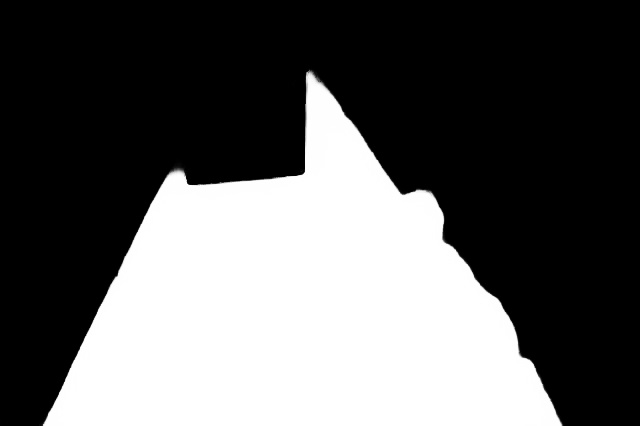} &
	\includegraphics[width=0.124\linewidth]{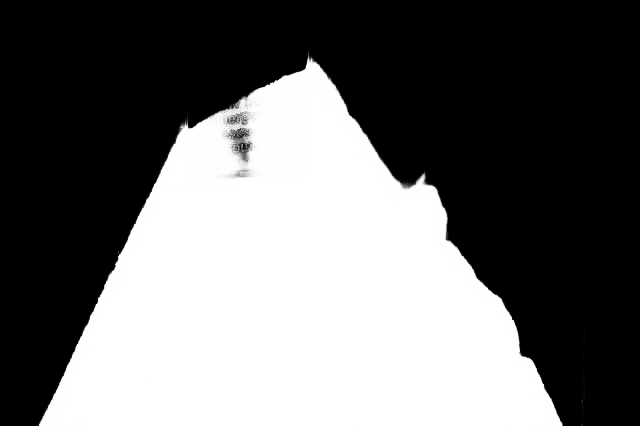} &
	\includegraphics[width=0.124\linewidth]{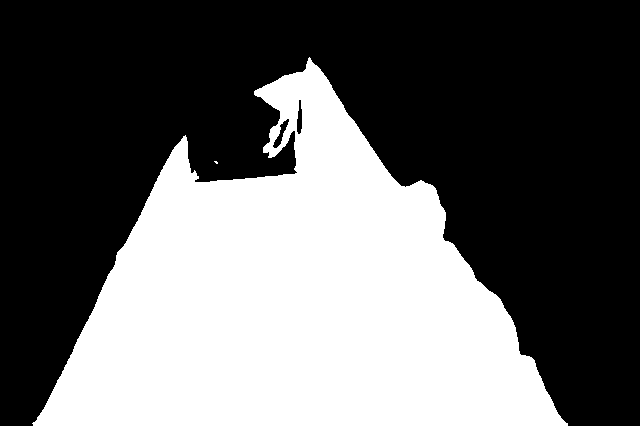} &
	\includegraphics[width=0.124\linewidth]{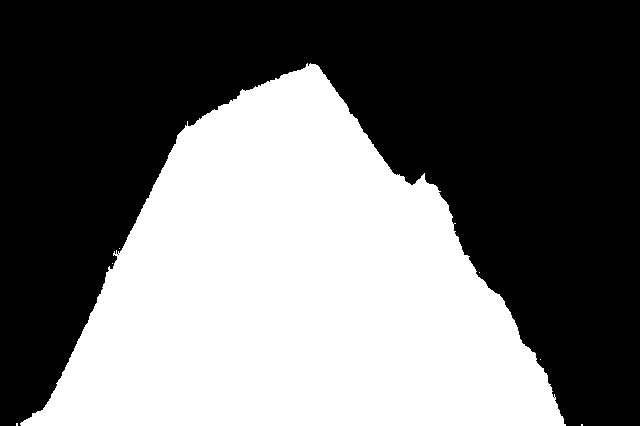} &
	\includegraphics[width=0.124\linewidth]{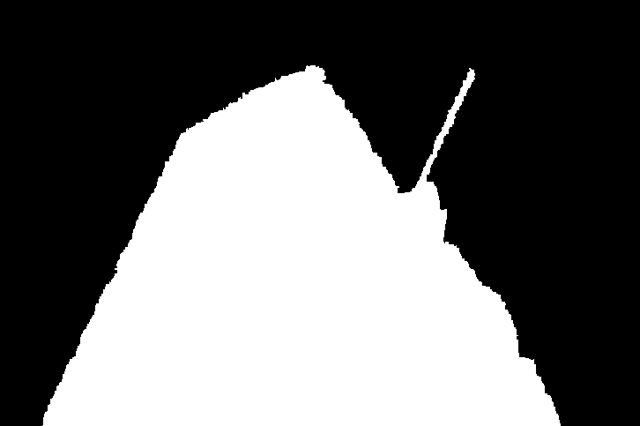} \\
	
	\includegraphics[width=0.124\linewidth]{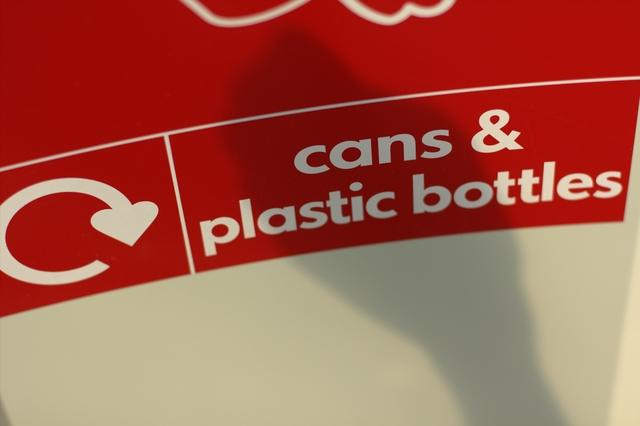} &
	\includegraphics[width=0.124\linewidth]{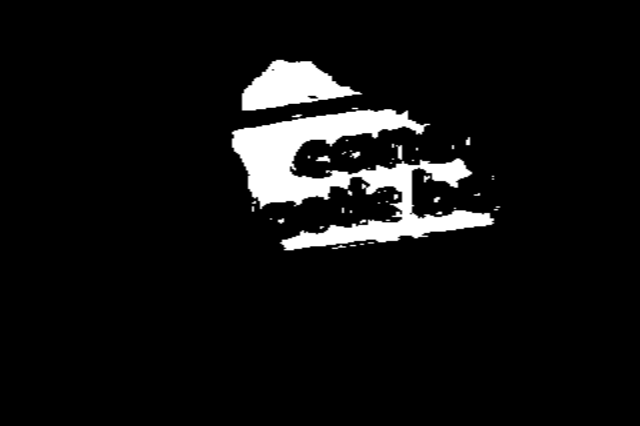} &
	\includegraphics[width=0.124\linewidth]{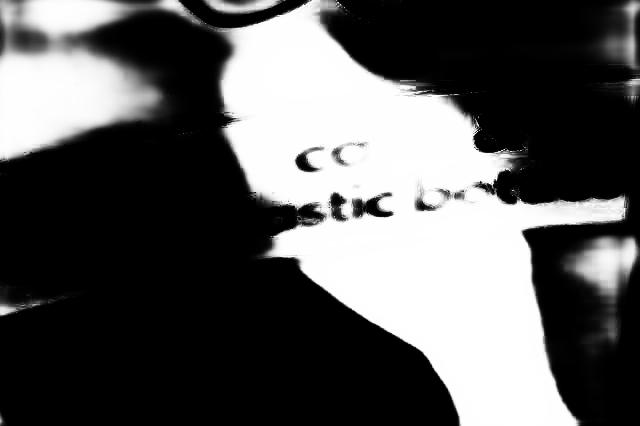} &
	\includegraphics[width=0.124\linewidth]{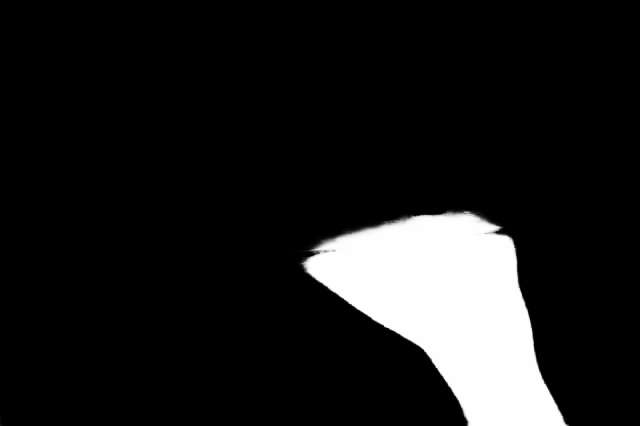} &
	\includegraphics[width=0.124\linewidth]{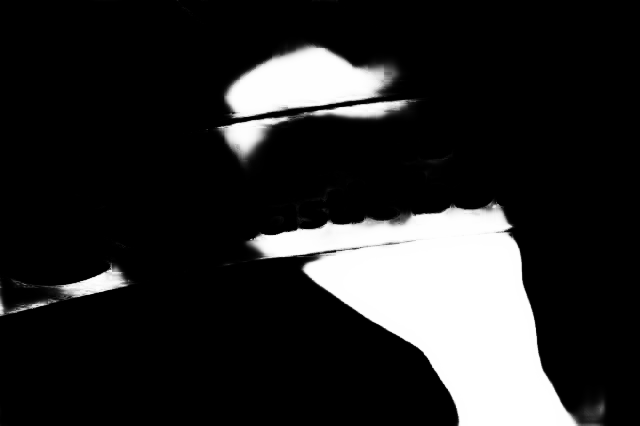} &
	\includegraphics[width=0.124\linewidth]{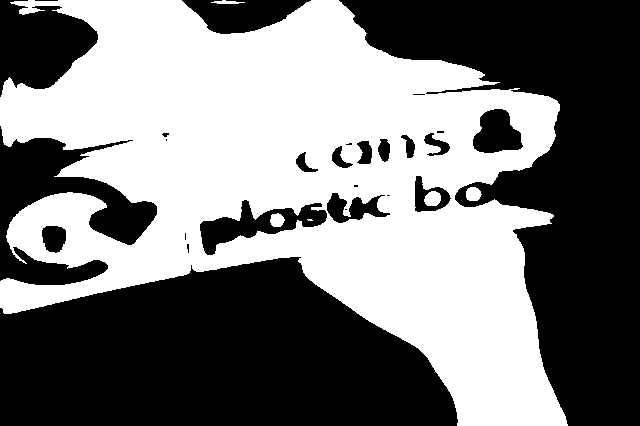} &
	\includegraphics[width=0.124\linewidth]{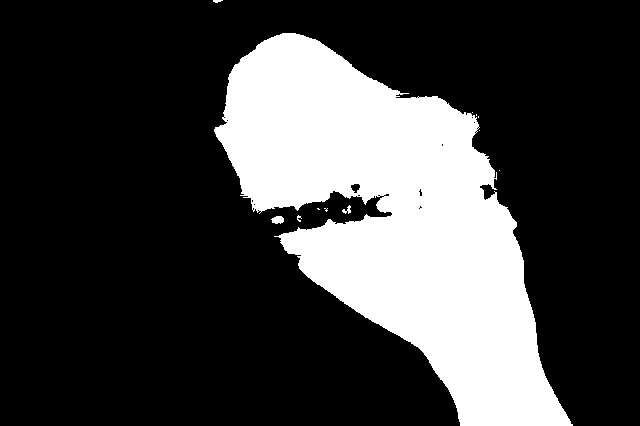} &
	\includegraphics[width=0.124\linewidth]{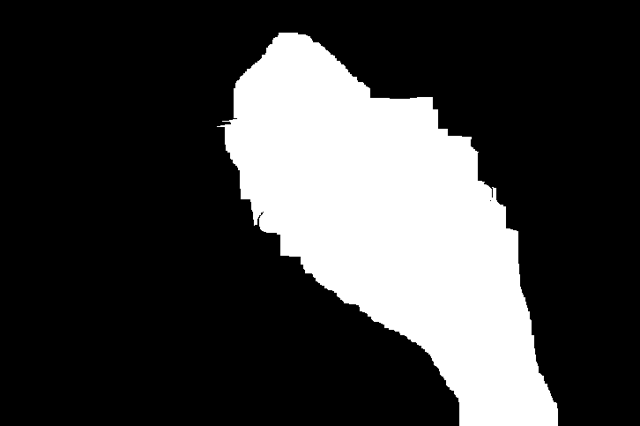} \\
	
	\includegraphics[width=0.124\linewidth]{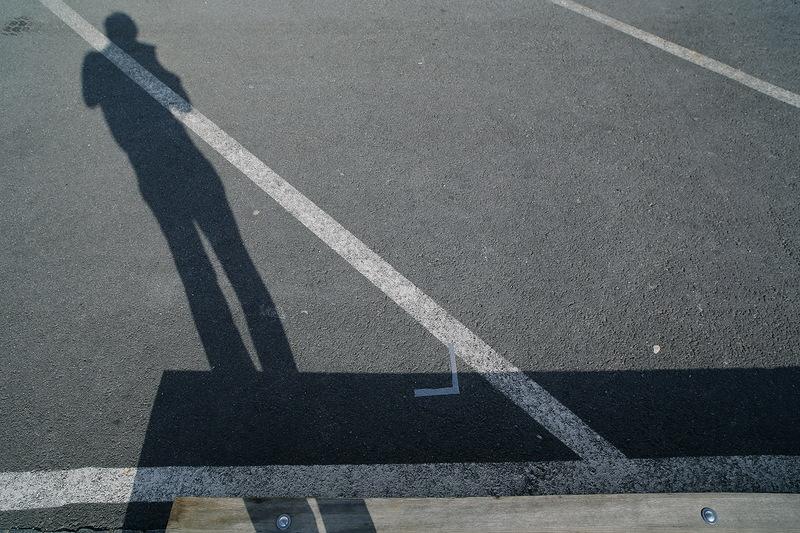} &
	\includegraphics[width=0.124\linewidth]{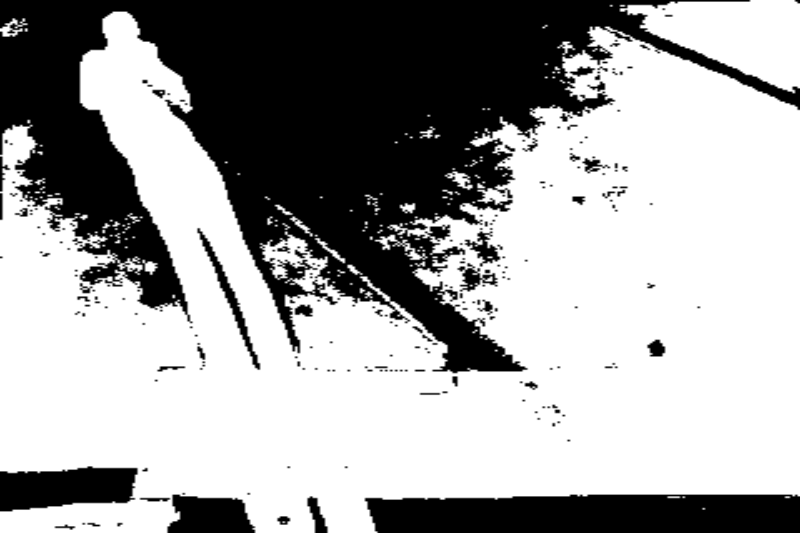} &
	\includegraphics[width=0.124\linewidth]{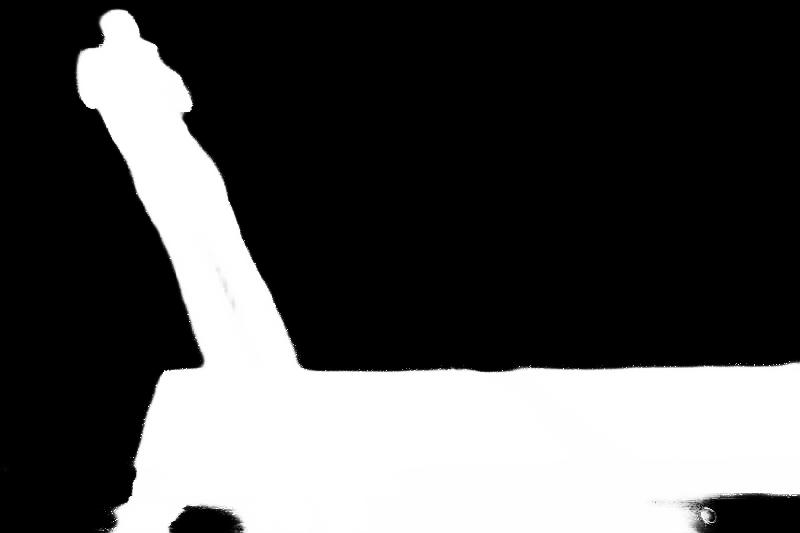} &
	\includegraphics[width=0.124\linewidth]{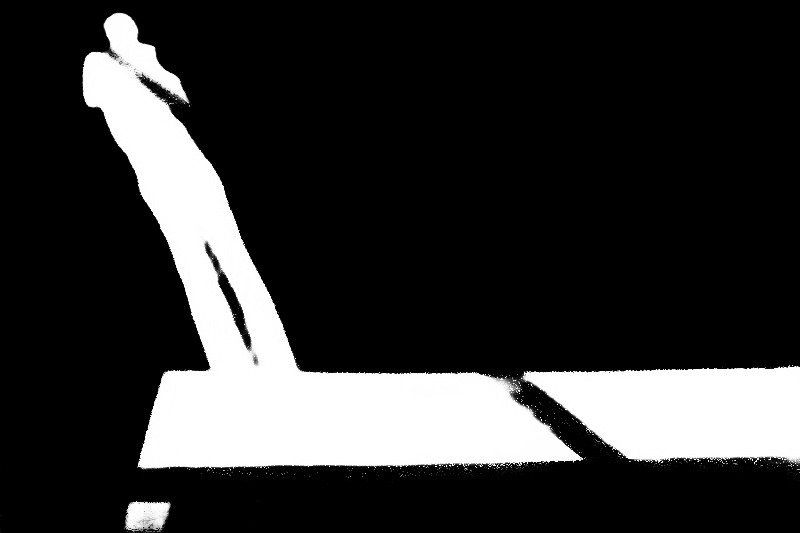} &
	\includegraphics[width=0.124\linewidth]{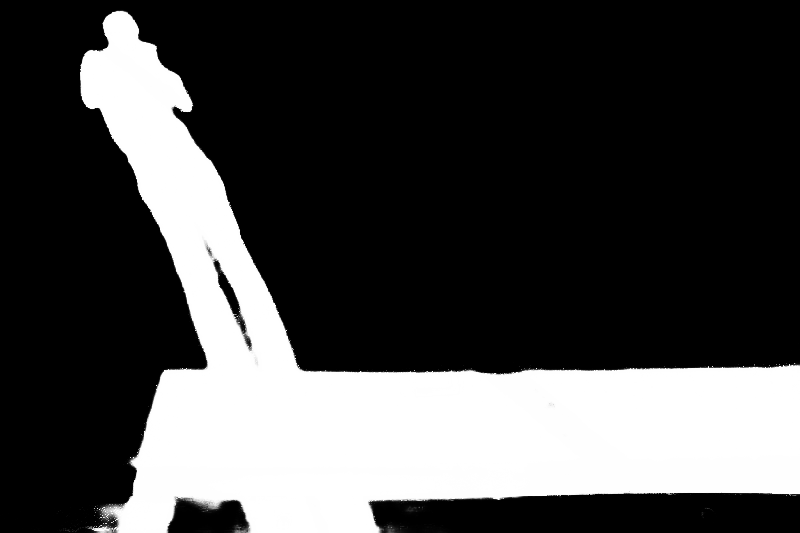} &
	\includegraphics[width=0.124\linewidth]{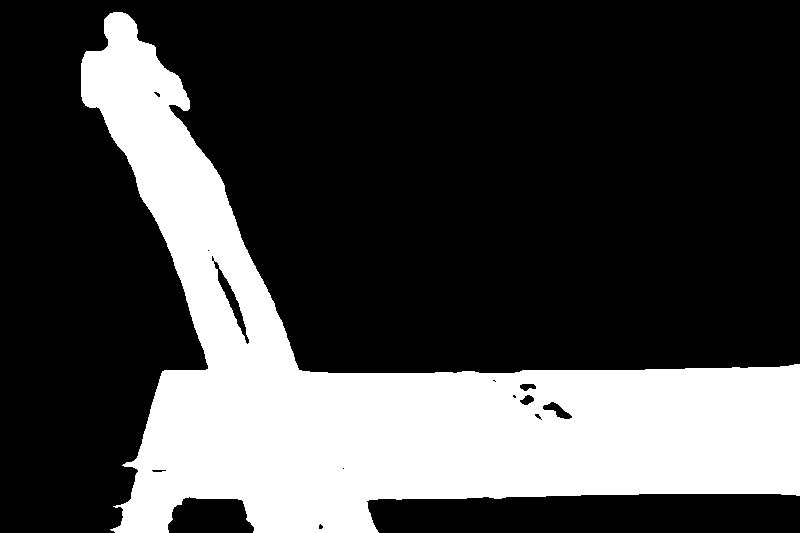} &
	\includegraphics[width=0.124\linewidth]{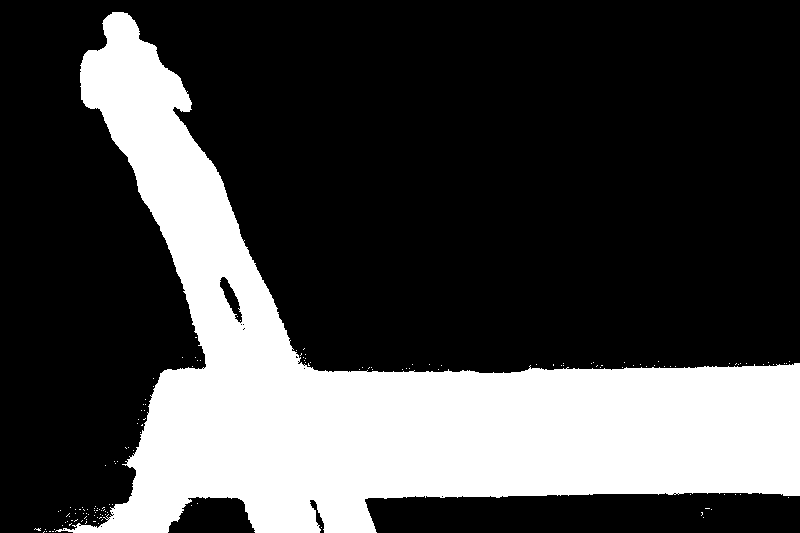} &
	\includegraphics[width=0.124\linewidth]{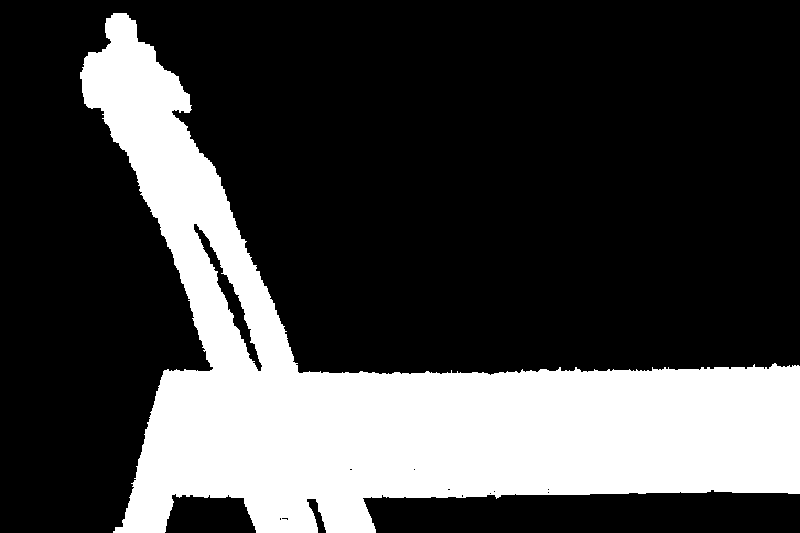} \\
	
	\includegraphics[width=0.124\linewidth]{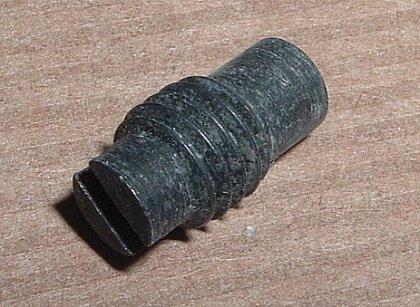} &
	\includegraphics[width=0.124\linewidth]{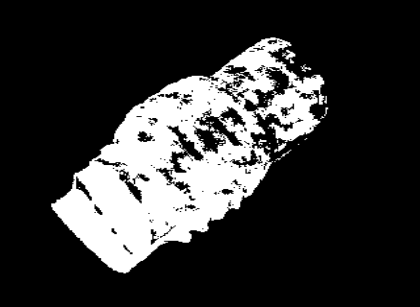} &
	\includegraphics[width=0.124\linewidth]{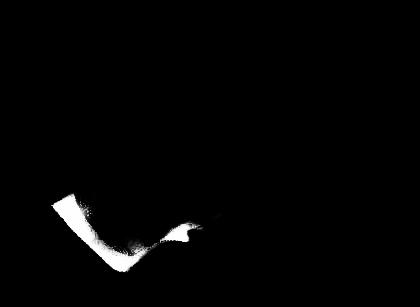} &
	\includegraphics[width=0.124\linewidth]{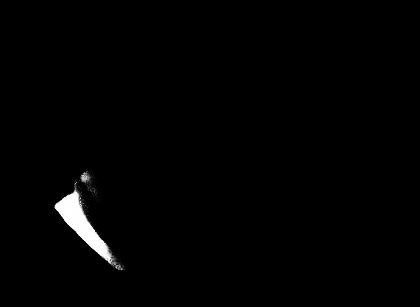} &
	\includegraphics[width=0.124\linewidth]{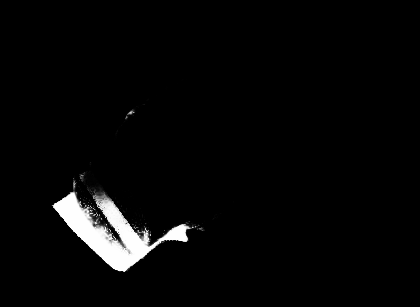} &
	\includegraphics[width=0.124\linewidth]{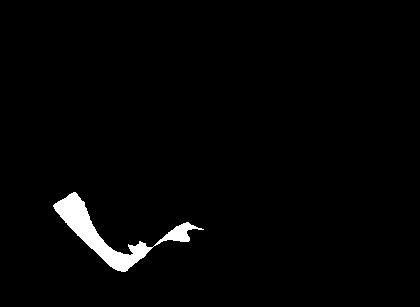} &
	\includegraphics[width=0.124\linewidth]{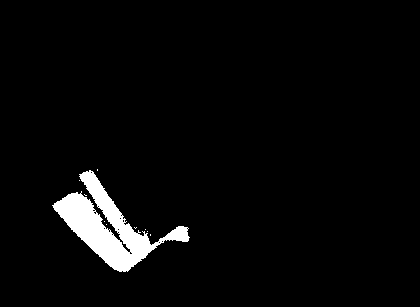} &
	\includegraphics[width=0.124\linewidth]{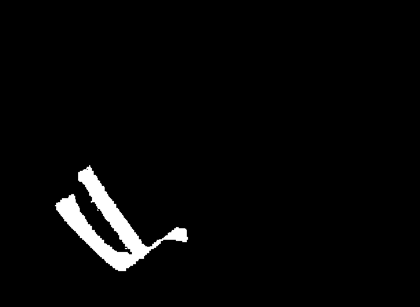} \\
	
	\includegraphics[width=0.124\linewidth]{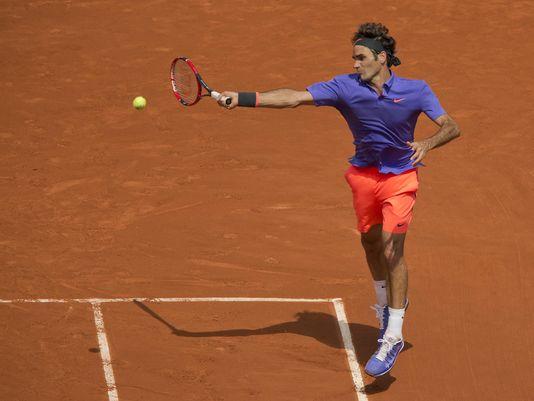} &
	\includegraphics[width=0.124\linewidth]{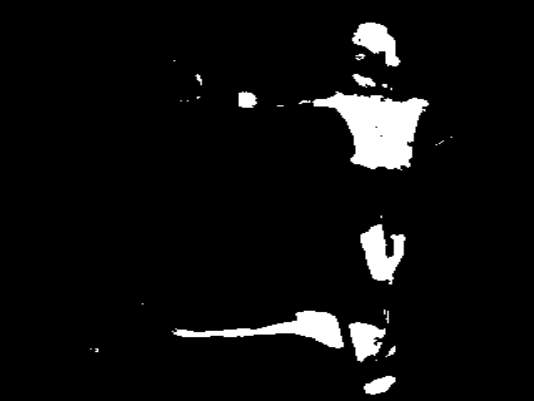} &
	\includegraphics[width=0.124\linewidth]{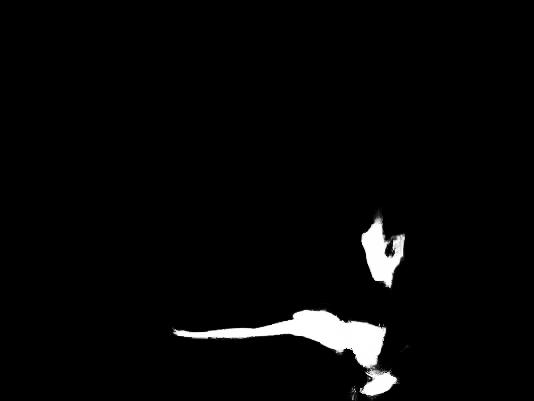} &
	\includegraphics[width=0.124\linewidth]{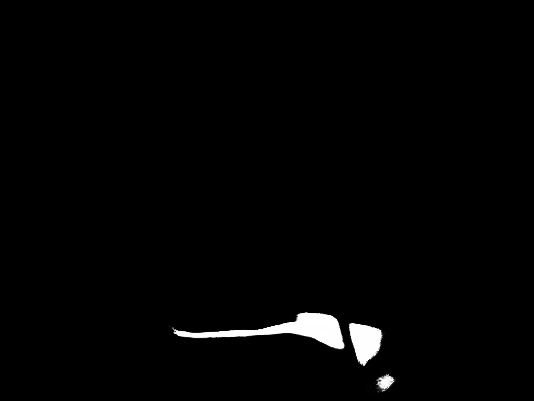} &
	\includegraphics[width=0.124\linewidth]{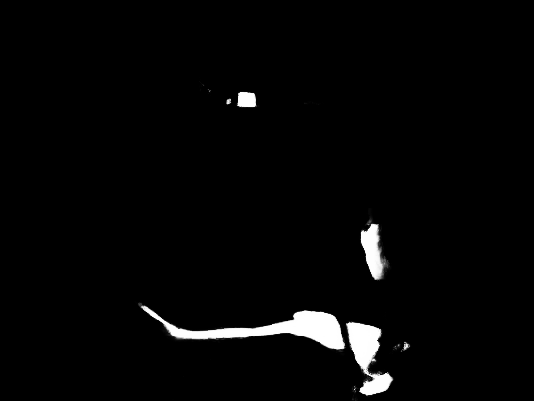} &
	\includegraphics[width=0.124\linewidth]{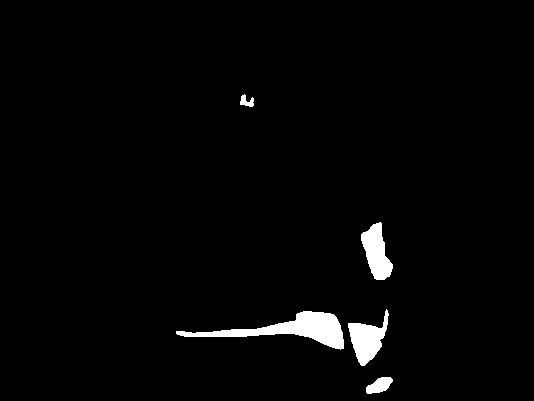} &
	\includegraphics[width=0.124\linewidth]{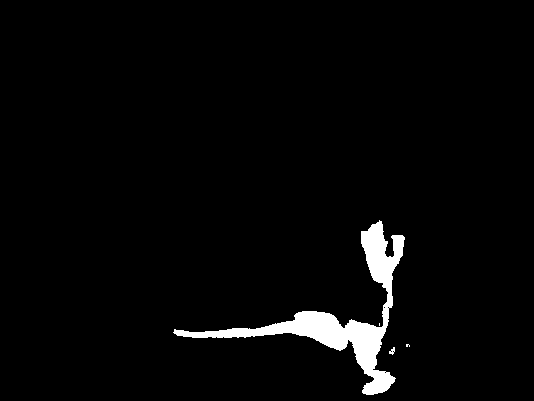} &
	\includegraphics[width=0.124\linewidth]{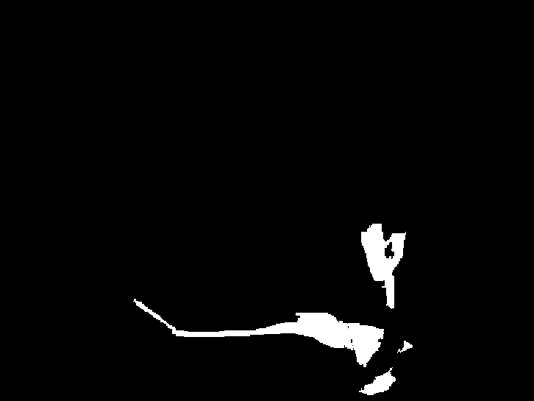} \\
	
	\includegraphics[width=0.124\linewidth]{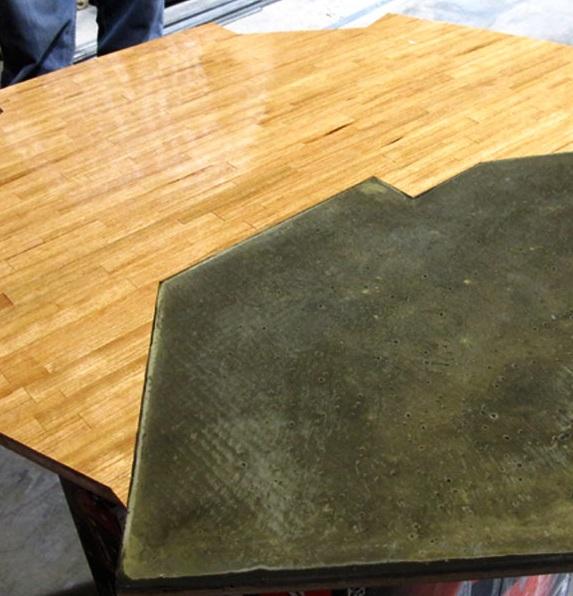} &
	\includegraphics[width=0.124\linewidth]{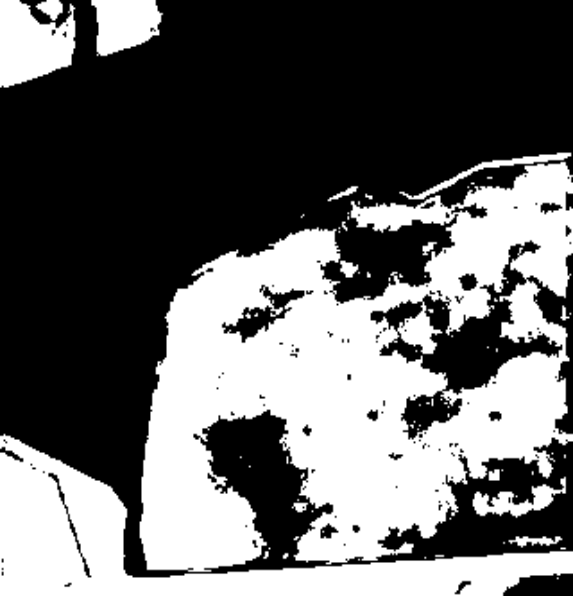} &
	\includegraphics[width=0.124\linewidth]{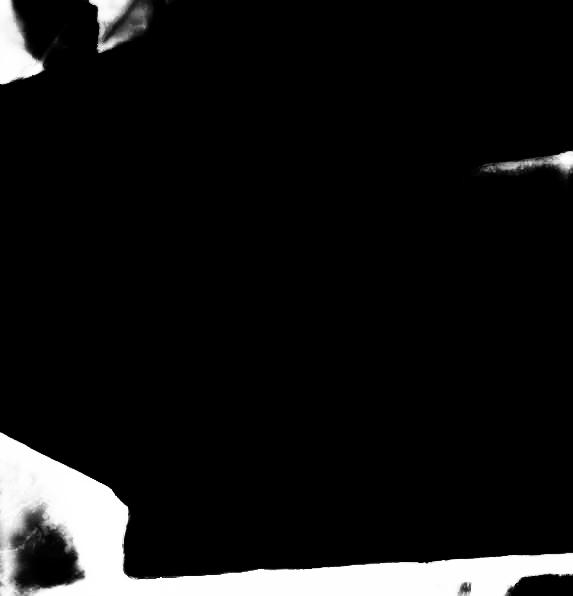} &
	\includegraphics[width=0.124\linewidth]{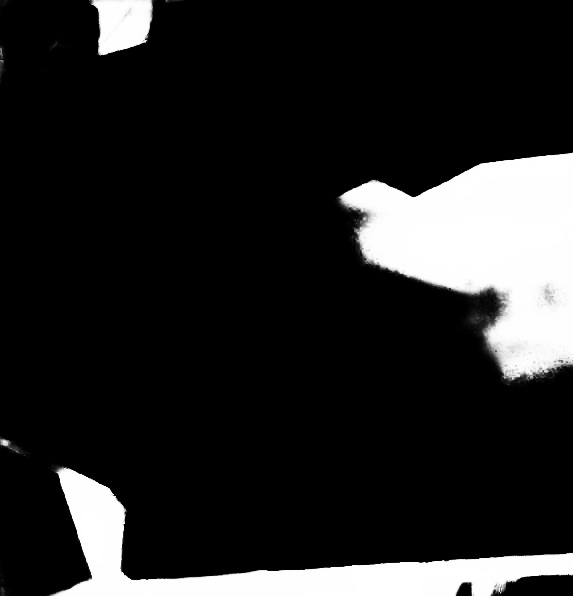} &
	\includegraphics[width=0.124\linewidth]{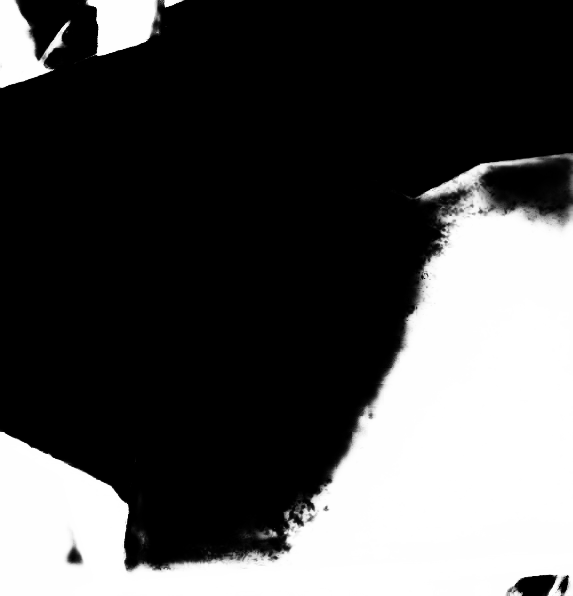} &
	\includegraphics[width=0.124\linewidth]{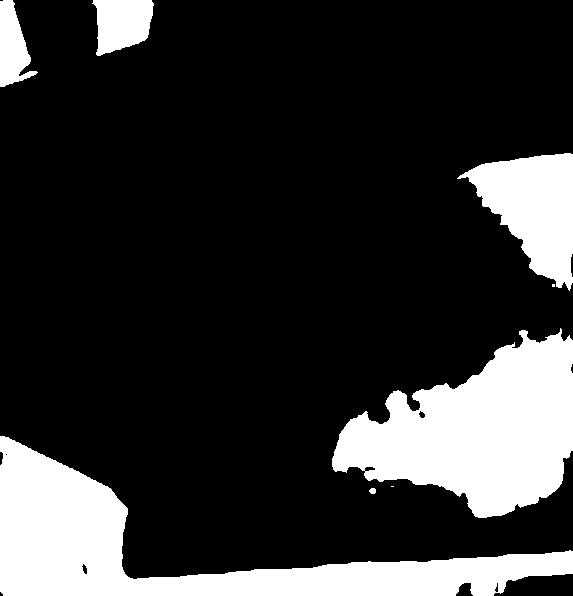} &
	\includegraphics[width=0.124\linewidth]{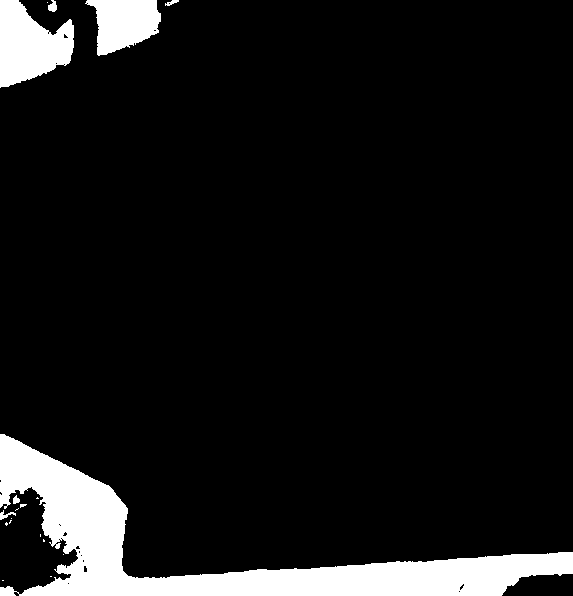} &
	\includegraphics[width=0.124\linewidth]{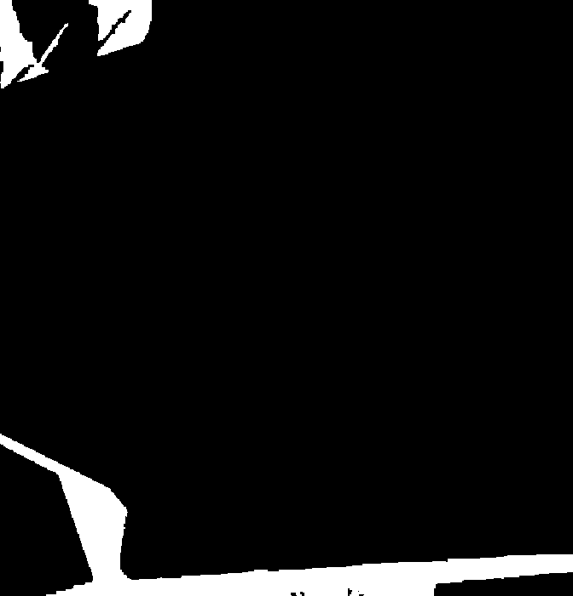} \\
	
	\includegraphics[width=0.124\linewidth]{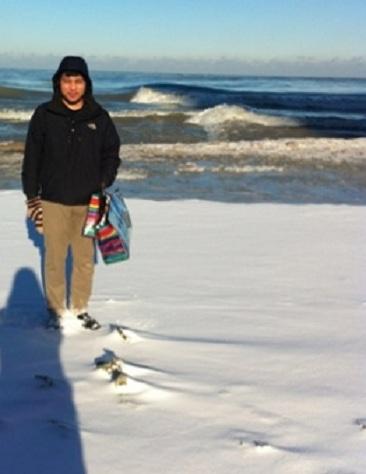} &
	\includegraphics[width=0.124\linewidth]{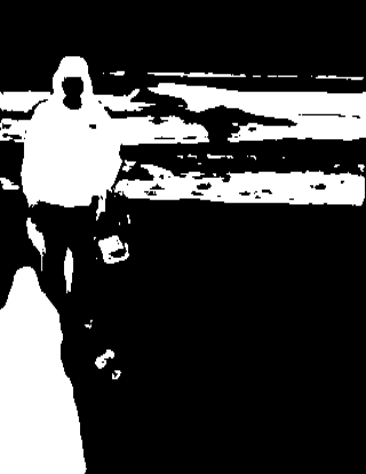} &
	\includegraphics[width=0.124\linewidth]{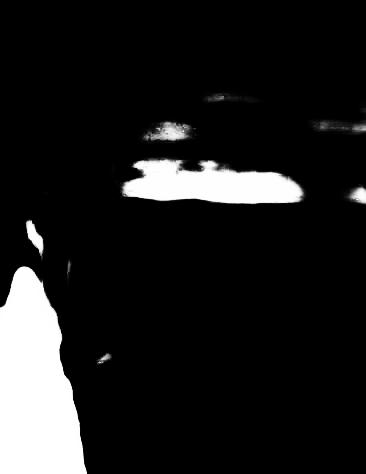} &
	\includegraphics[width=0.124\linewidth]{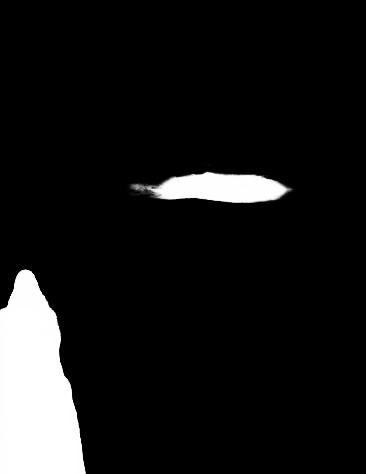} &
	\includegraphics[width=0.124\linewidth]{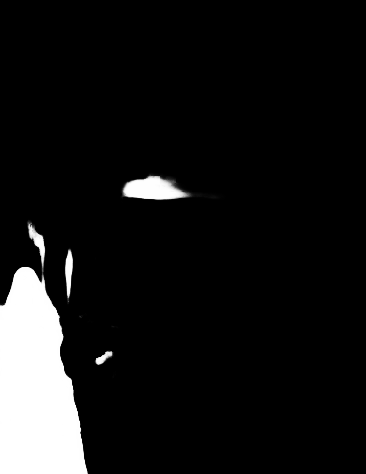} &
	\includegraphics[width=0.124\linewidth]{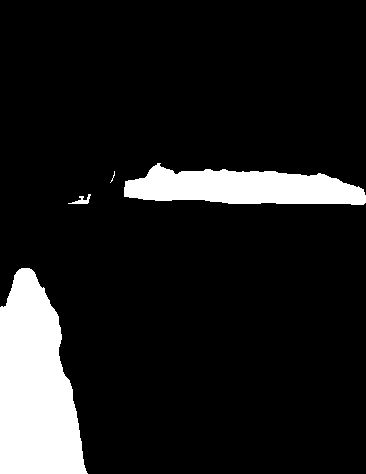} &
	\includegraphics[width=0.124\linewidth]{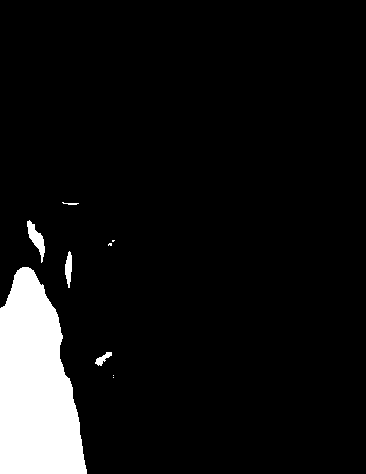} &
	\includegraphics[width=0.124\linewidth]{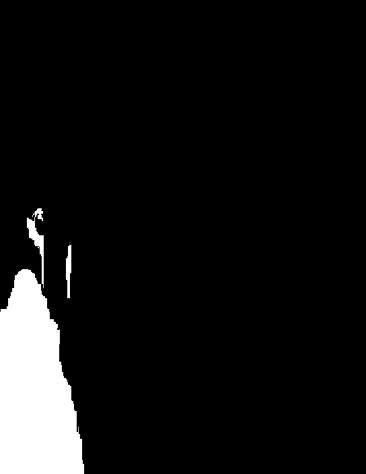} \\

{ \footnotesize{(a) Input}} &
{ \footnotesize{(b) dark regions}} &
{ \footnotesize{(c) BDRAR~\cite{DBLP:conf/eccv/ZhuDHFXQH18}}} &
{ \footnotesize{(d) DSC~\cite{DBLP:conf/cvpr/Hu0F0H18}}} &
{ \footnotesize{(e) DSDNet~\cite{DBLP:conf/cvpr/ZhengQCL19}}} &
{ \footnotesize{(f) MTMT~\cite{DBLP:conf/cvpr/Chen0WW0H20}}} &
{ \footnotesize{(g) Ours}}  &
{ \footnotesize{ (h) GT}} \\

\end{tabular}
\end{center}
\caption{Visual comparison of the shadow detection performances between our method ($7^{th}$ column) and the state-of-the-art methods ($3^{rd}-6^{th}$ columns) on  SBU~\cite{DBLP:conf/eccv/VicenteHYHS16}. The second column shows the recommended dark regions produced by our DRR module. Our method produces the closest results to the ground truth, particularly inside the recommended dark regions.}
\label{fig:visual}
\end{figure*}

\begin{figure*}[t]
\renewcommand{\tabcolsep}{1pt}
\begin{center}
\begin{tabular}{cccccccc}
	\includegraphics[width=0.124\linewidth]{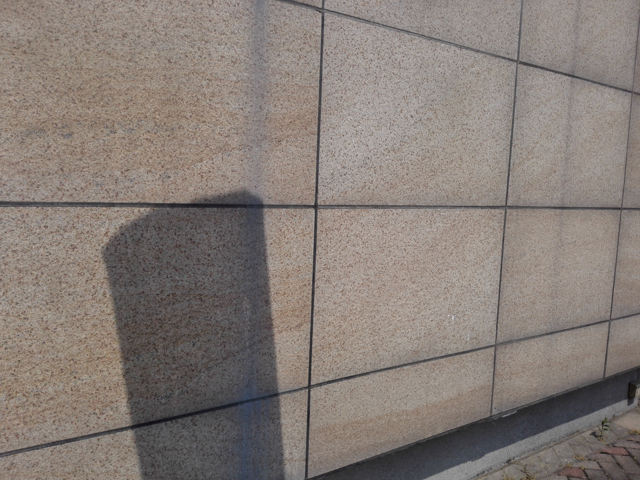} &
	\includegraphics[width=0.124\linewidth]{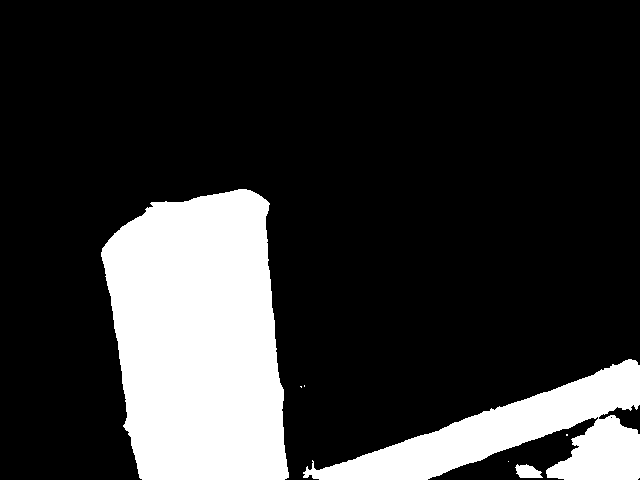} &
	\includegraphics[width=0.124\linewidth]{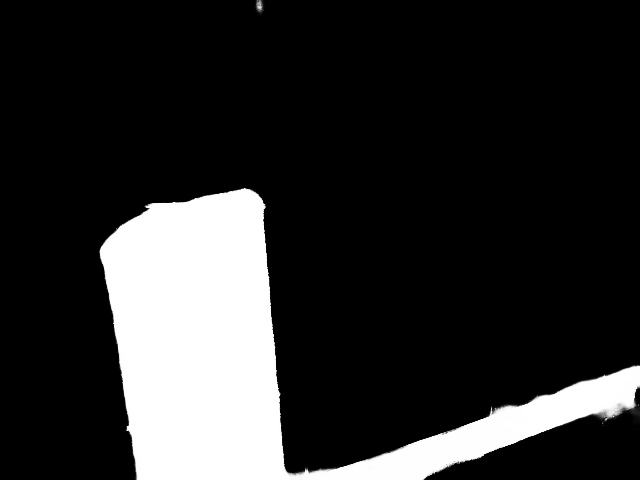} &
	\includegraphics[width=0.124\linewidth]{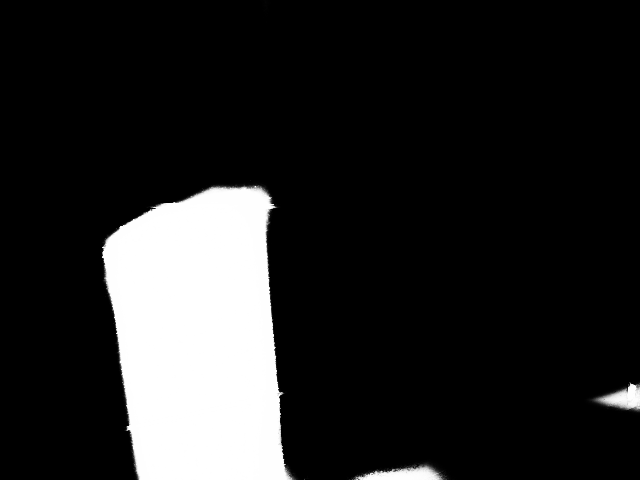} &
	\includegraphics[width=0.124\linewidth]{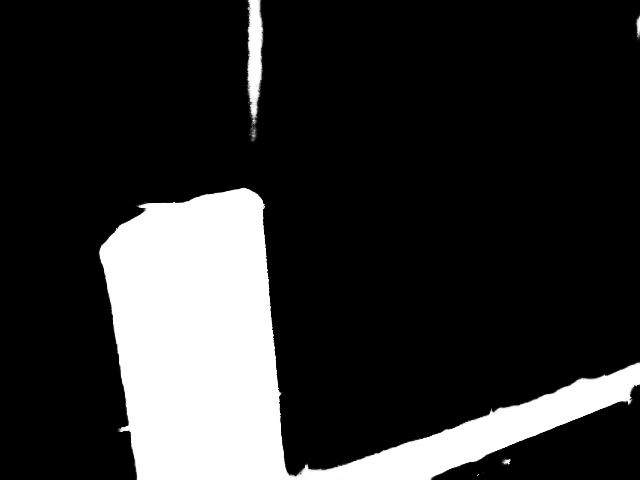} &
	\includegraphics[width=0.124\linewidth]{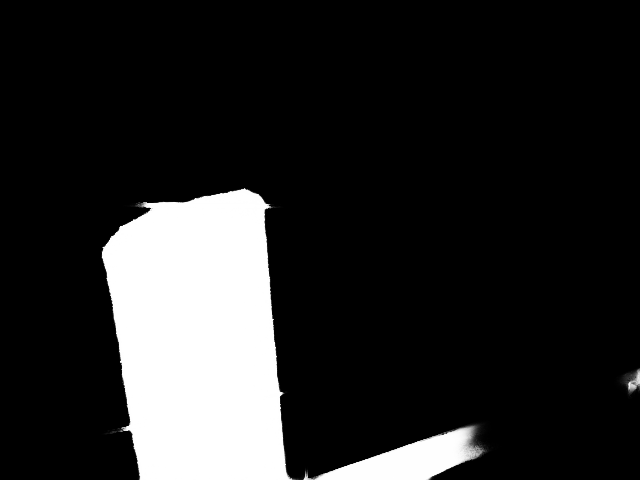} &
	\includegraphics[width=0.124\linewidth]{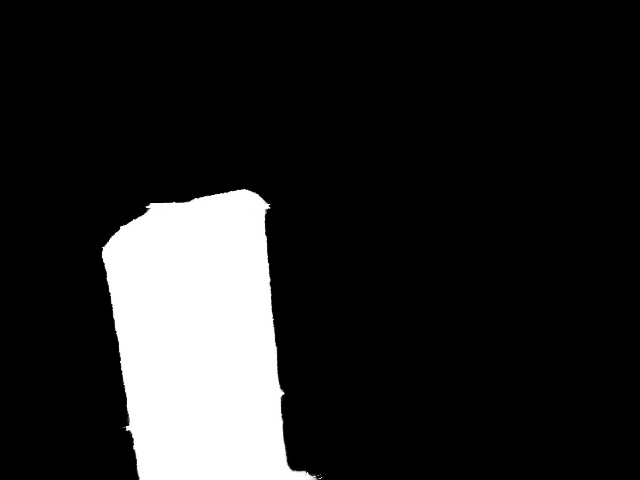} &
	\includegraphics[width=0.124\linewidth]{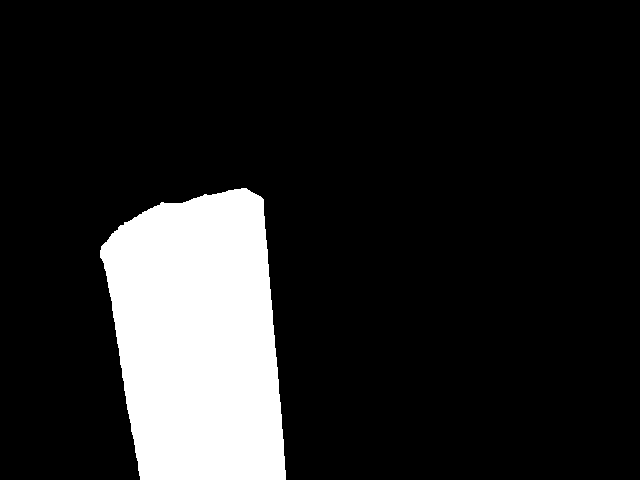} \\
	
	\includegraphics[width=0.124\linewidth]{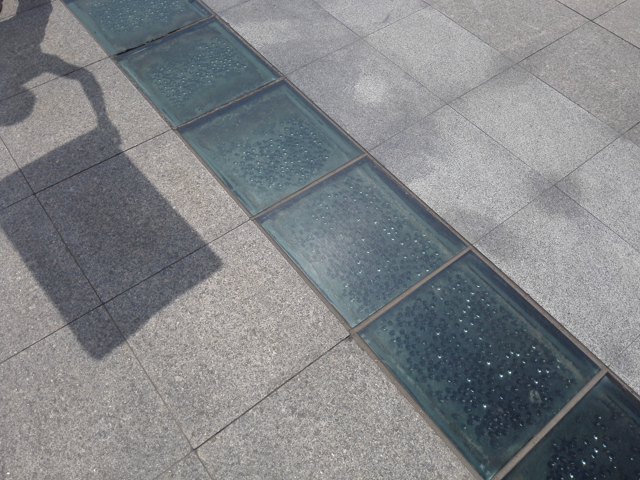} &
	\includegraphics[width=0.124\linewidth]{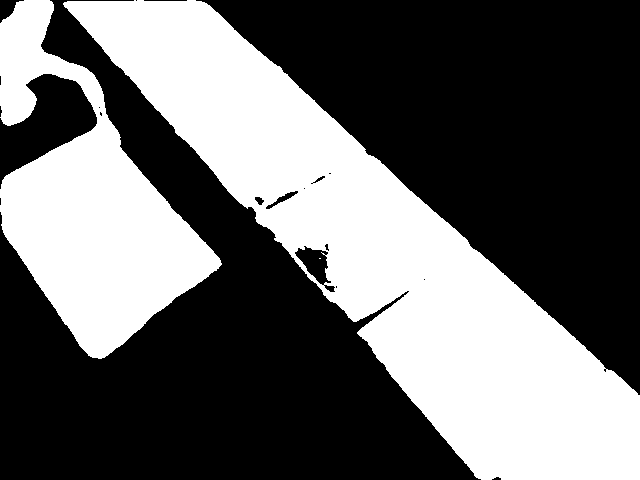} &
	\includegraphics[width=0.124\linewidth]{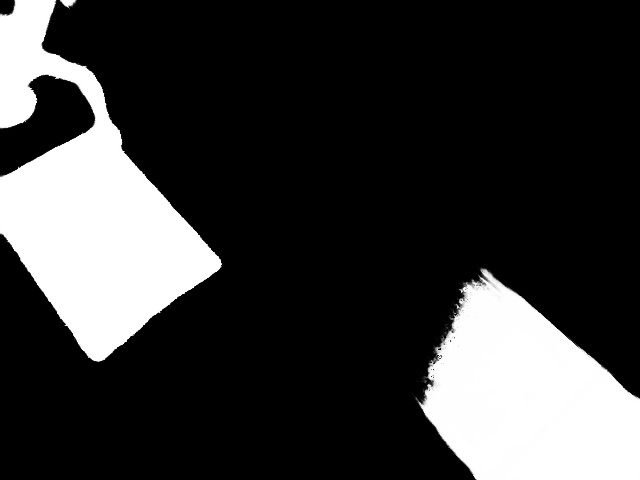} &
	\includegraphics[width=0.124\linewidth]{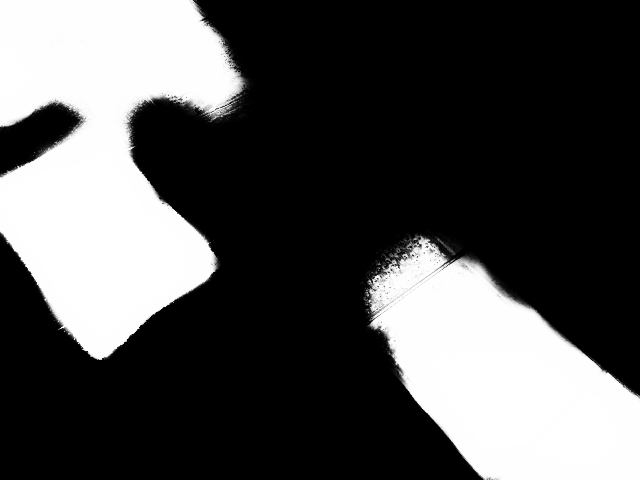} &
	\includegraphics[width=0.124\linewidth]{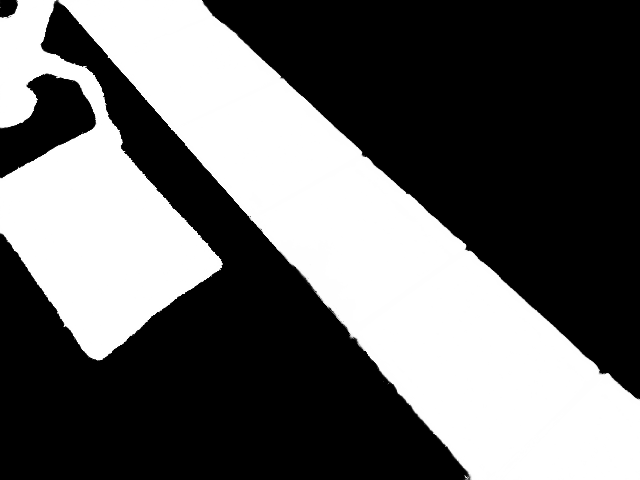} &
	\includegraphics[width=0.124\linewidth]{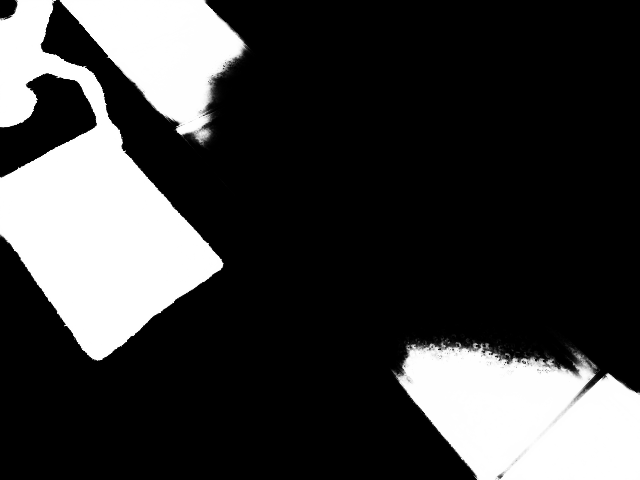} &
	\includegraphics[width=0.124\linewidth]{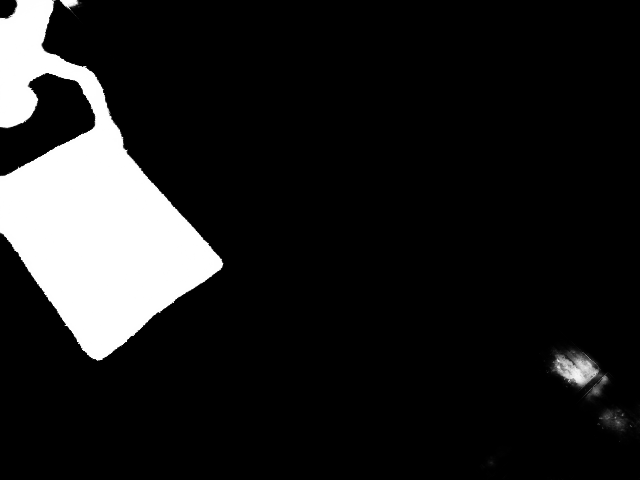} &
	\includegraphics[width=0.124\linewidth]{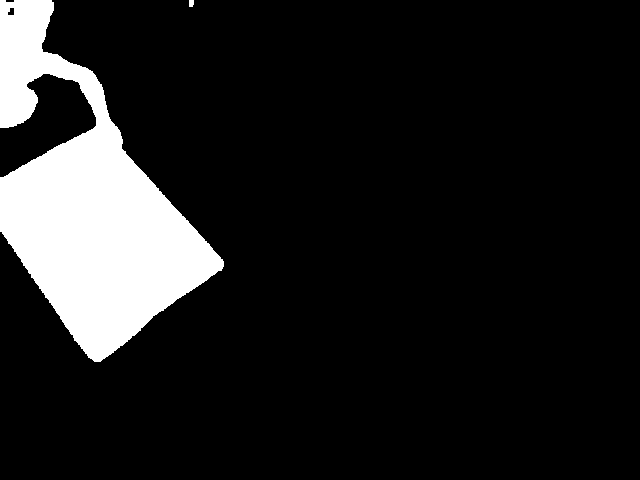} \\
	
	\includegraphics[width=0.124\linewidth]{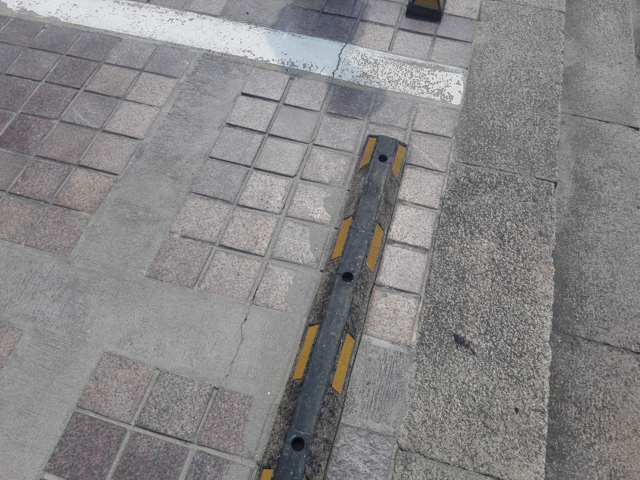} &
	\includegraphics[width=0.124\linewidth]{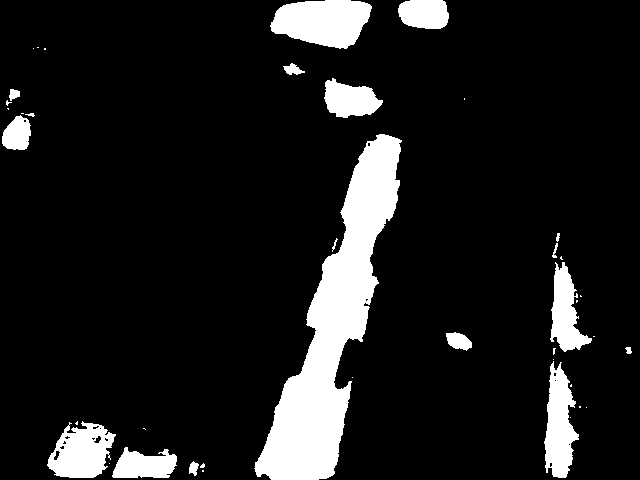} &
	\includegraphics[width=0.124\linewidth]{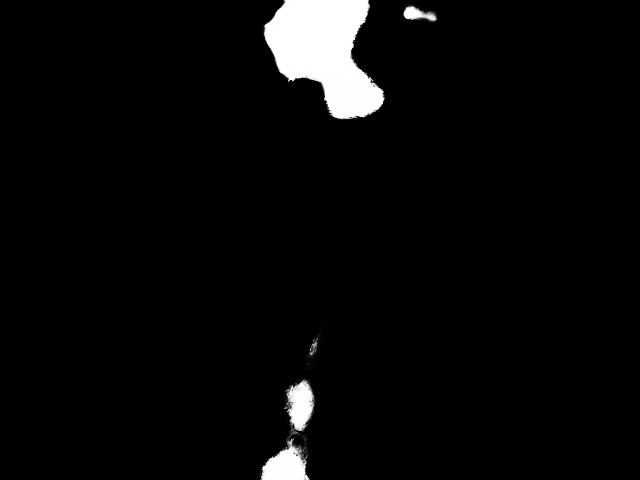} &
	\includegraphics[width=0.124\linewidth]{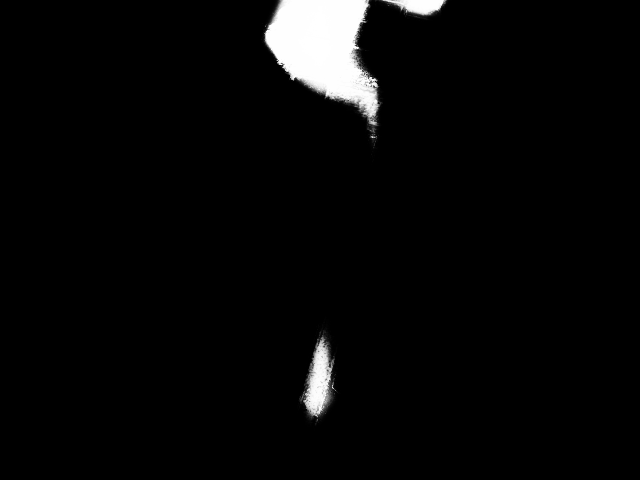} &
	\includegraphics[width=0.124\linewidth]{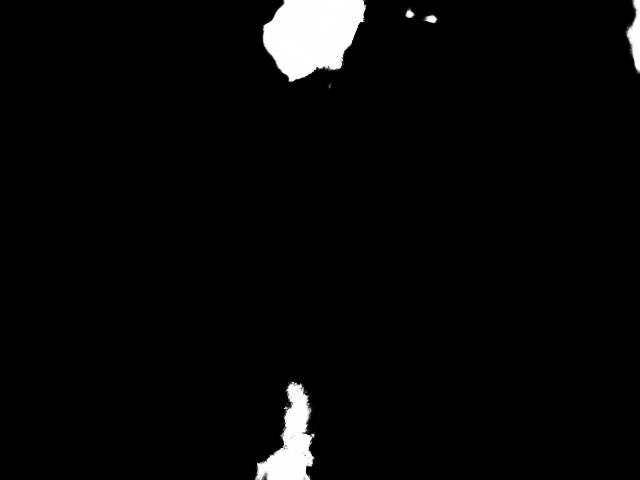} &
	\includegraphics[width=0.124\linewidth]{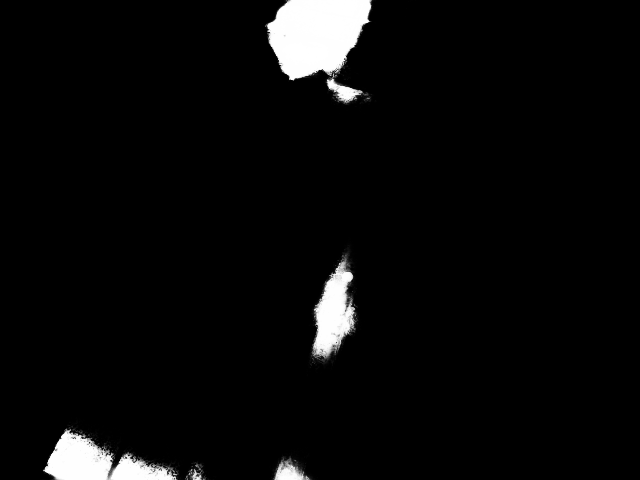} &
	\includegraphics[width=0.124\linewidth]{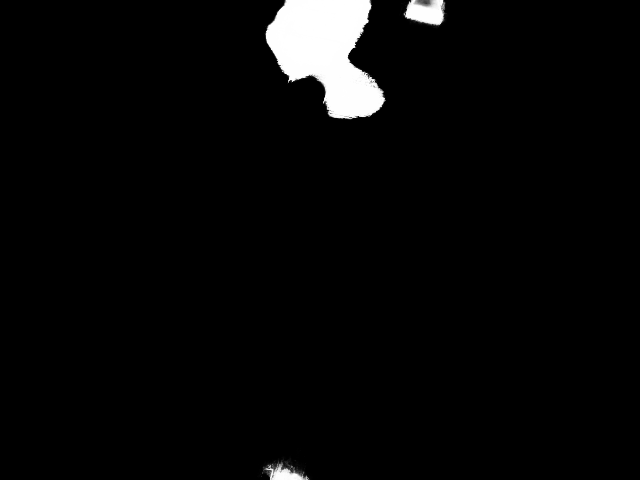} &
	\includegraphics[width=0.124\linewidth]{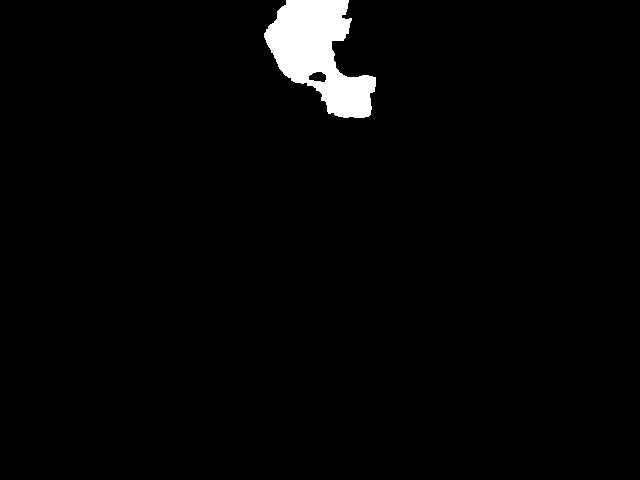} \\
	
	\includegraphics[width=0.124\linewidth]{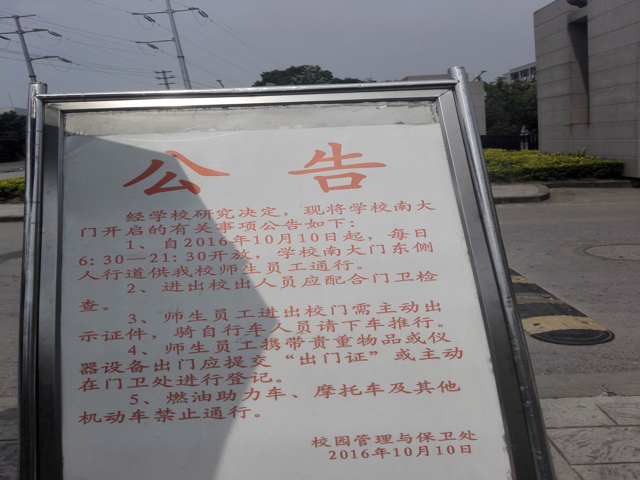} &
	\includegraphics[width=0.124\linewidth]{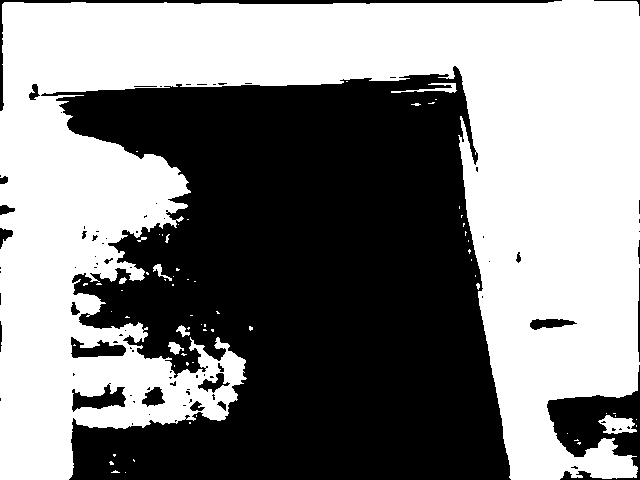} &
	\includegraphics[width=0.124\linewidth]{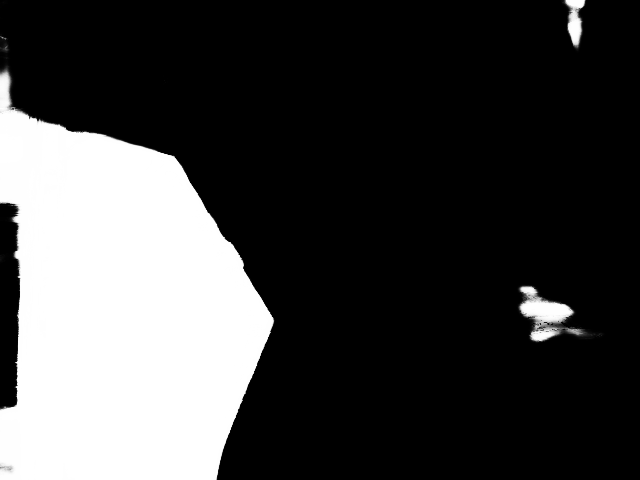} &
	\includegraphics[width=0.124\linewidth]{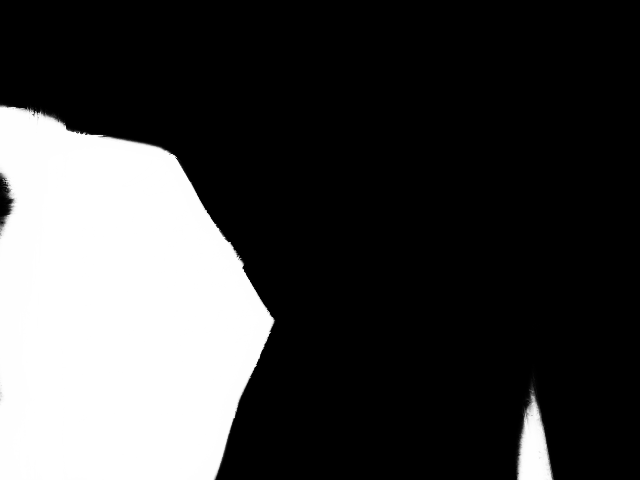} &
	\includegraphics[width=0.124\linewidth]{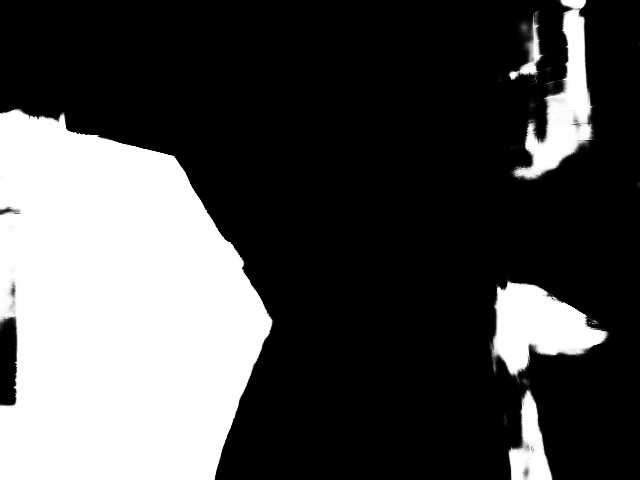} &
	\includegraphics[width=0.124\linewidth]{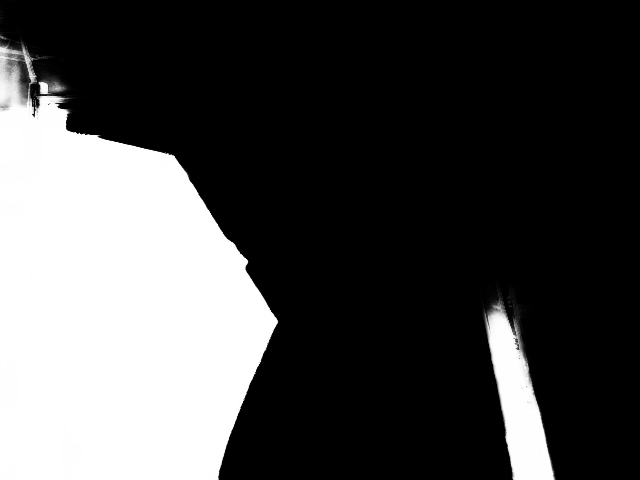} &
	\includegraphics[width=0.124\linewidth]{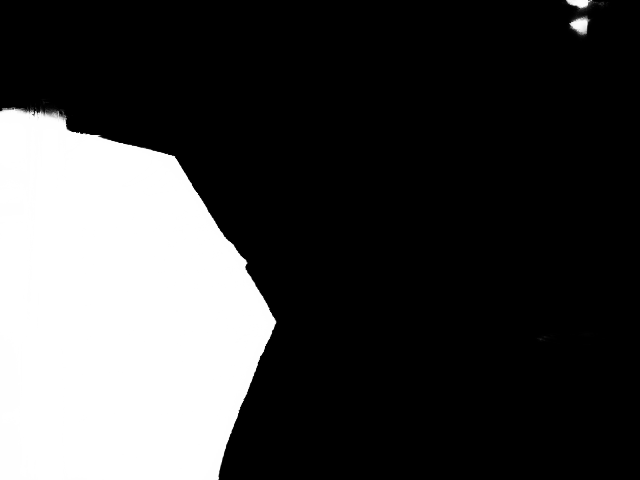} &
	\includegraphics[width=0.124\linewidth]{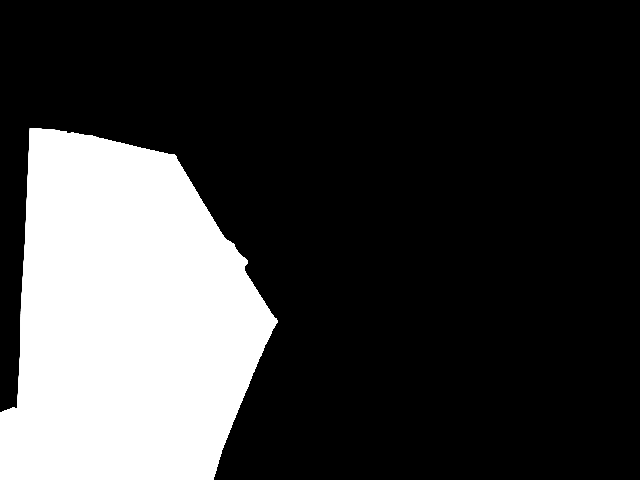} \\
	
	\includegraphics[width=0.124\linewidth]{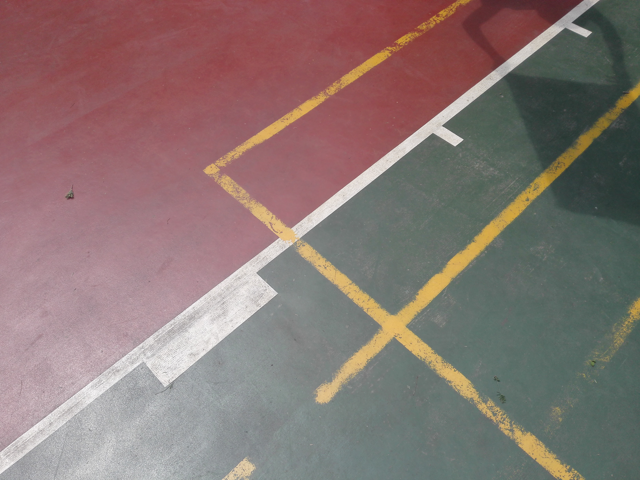} &
	\includegraphics[width=0.124\linewidth]{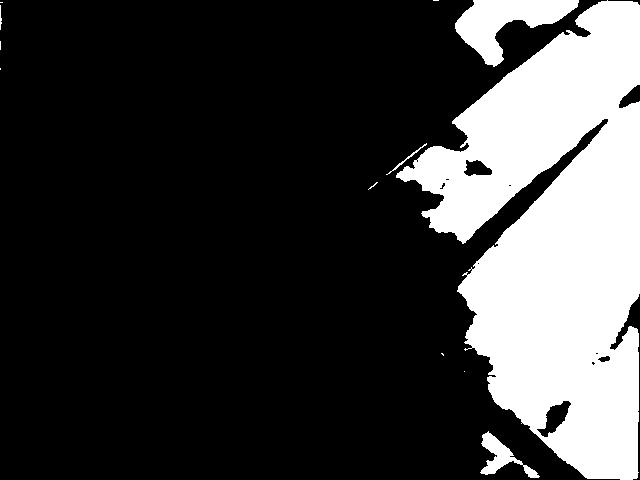} &
	\includegraphics[width=0.124\linewidth]{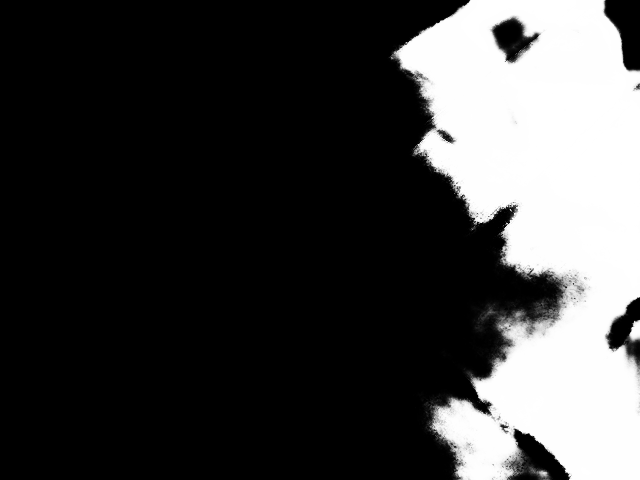} &
	\includegraphics[width=0.124\linewidth]{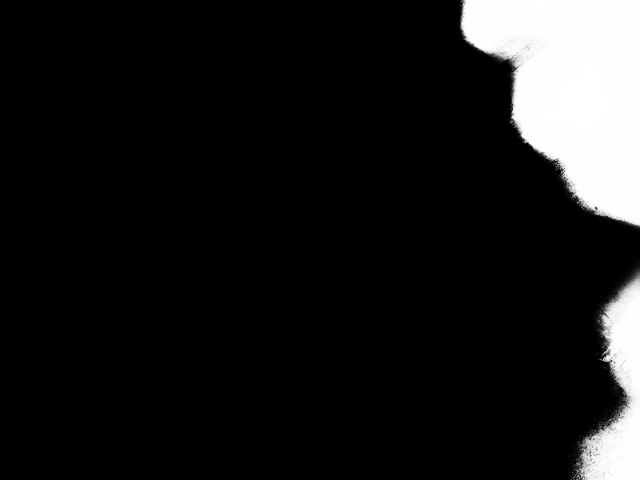} &
	\includegraphics[width=0.124\linewidth]{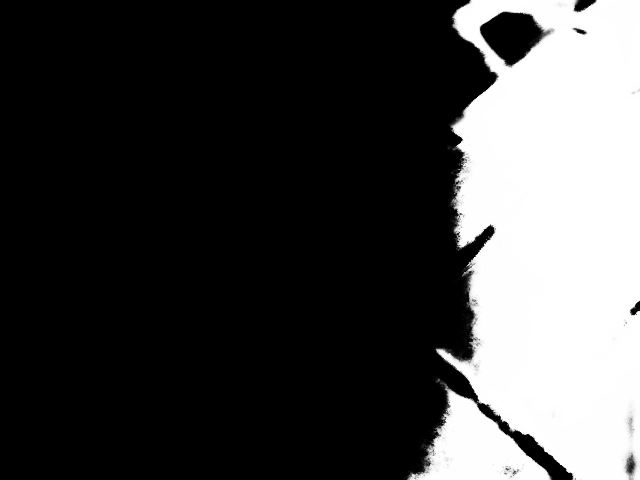} &
	\includegraphics[width=0.124\linewidth]{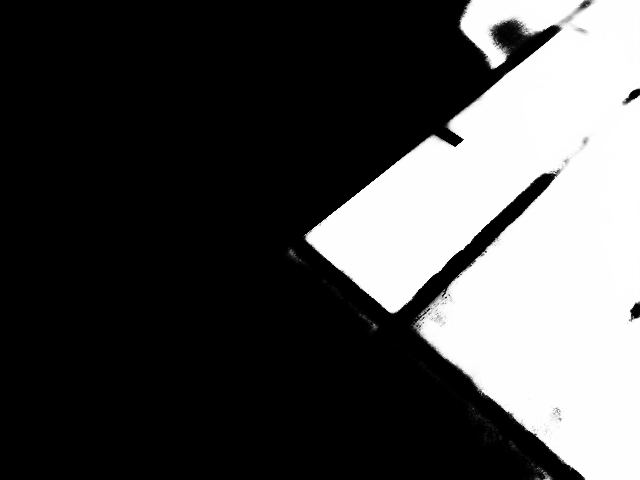} &
	\includegraphics[width=0.124\linewidth]{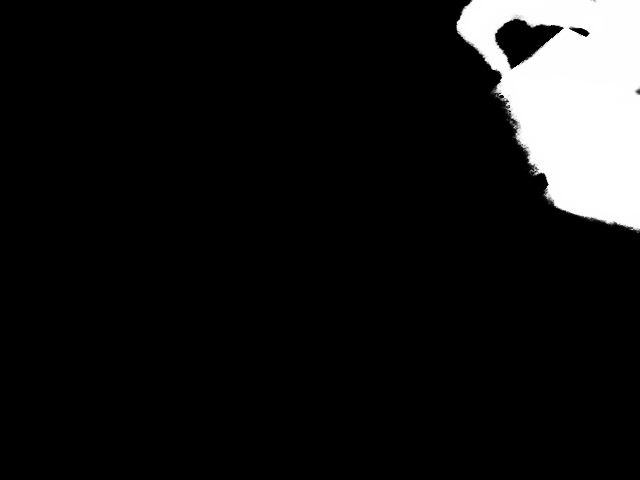} &
	\includegraphics[width=0.124\linewidth]{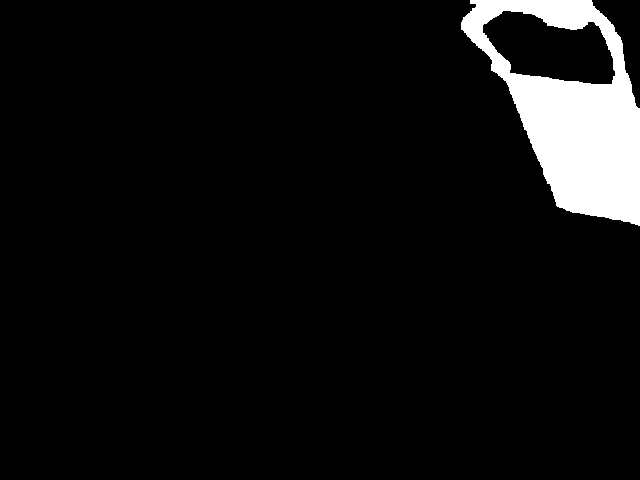} \\
	
	\includegraphics[width=0.124\linewidth]{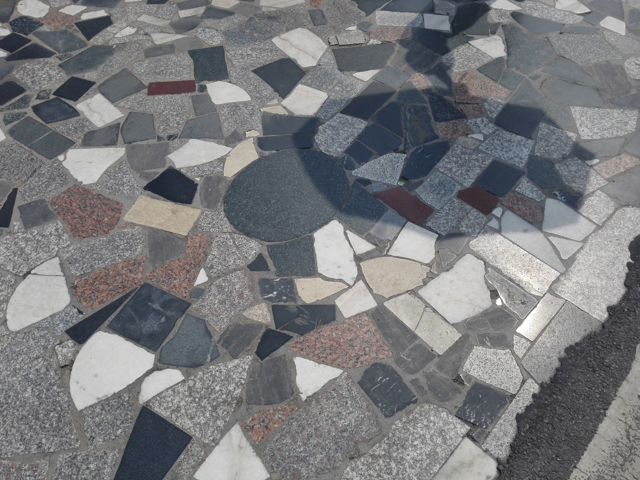} &
	\includegraphics[width=0.124\linewidth]{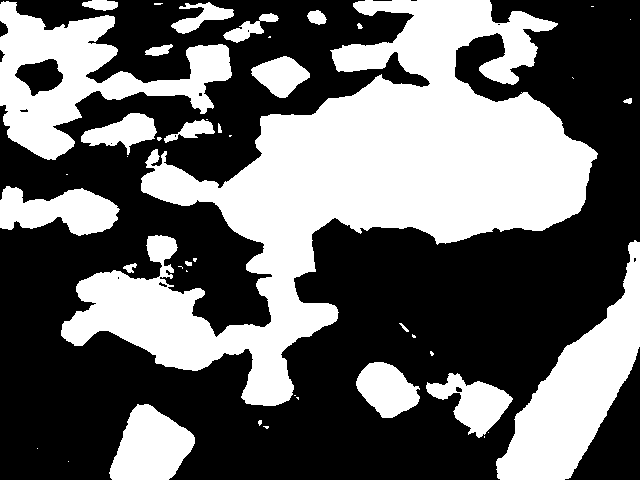} &
	\includegraphics[width=0.124\linewidth]{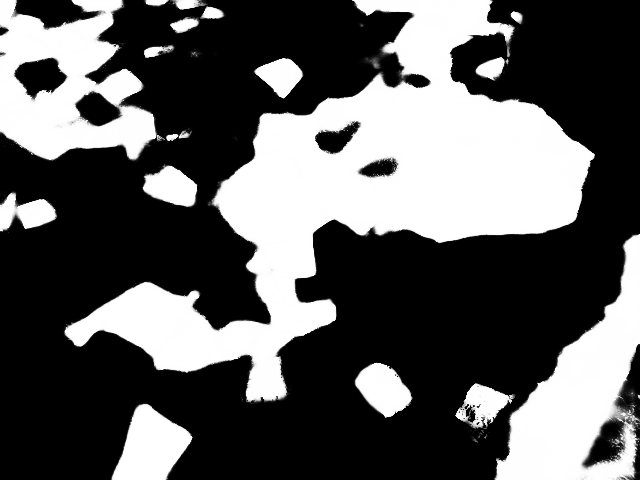} &
	\includegraphics[width=0.124\linewidth]{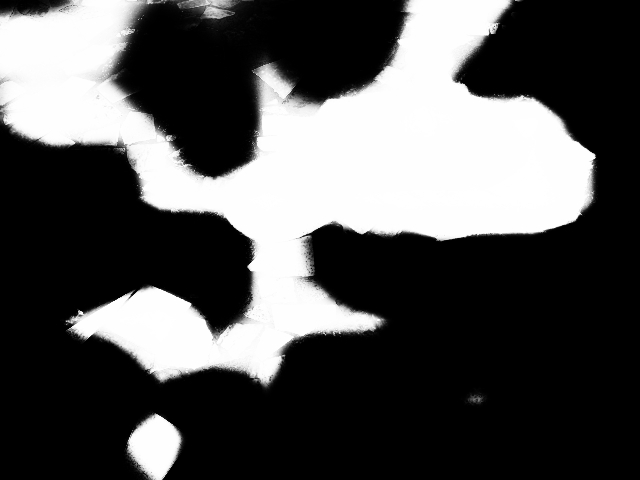} &
	\includegraphics[width=0.124\linewidth]{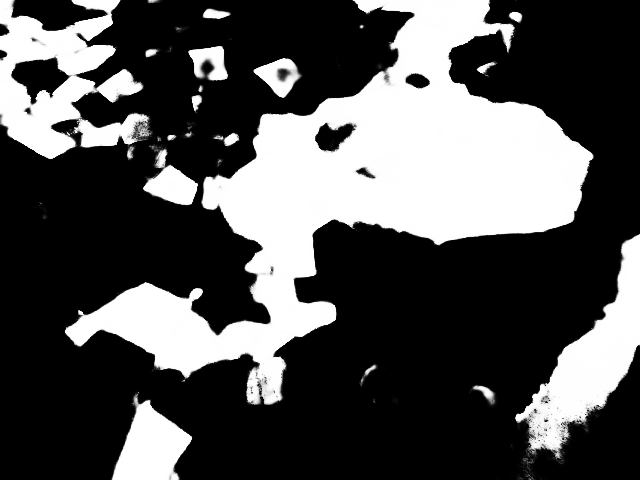} &
	\includegraphics[width=0.124\linewidth]{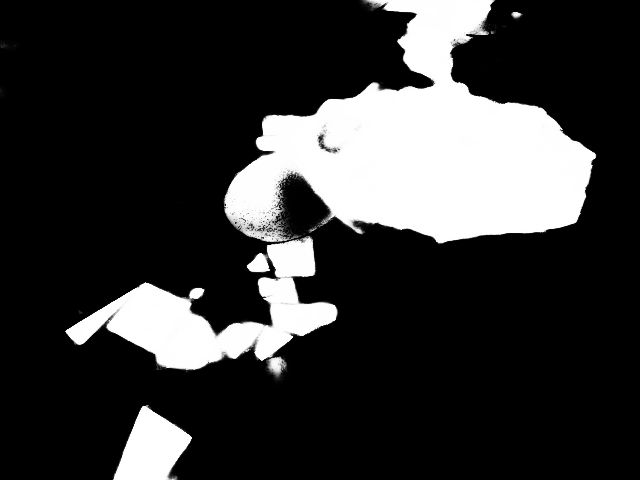} &
	\includegraphics[width=0.124\linewidth]{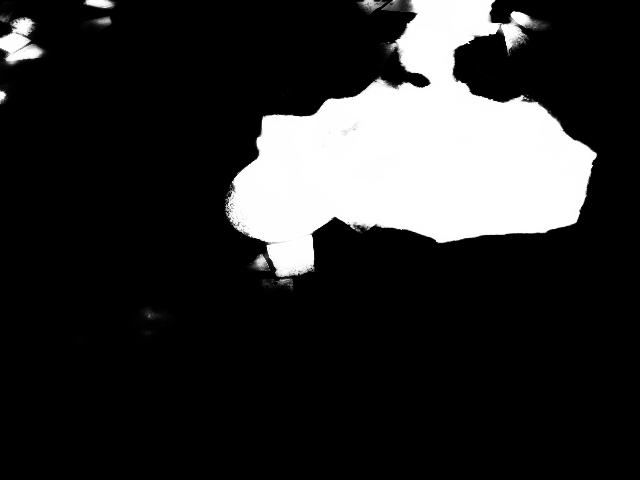} &
	\includegraphics[width=0.124\linewidth]{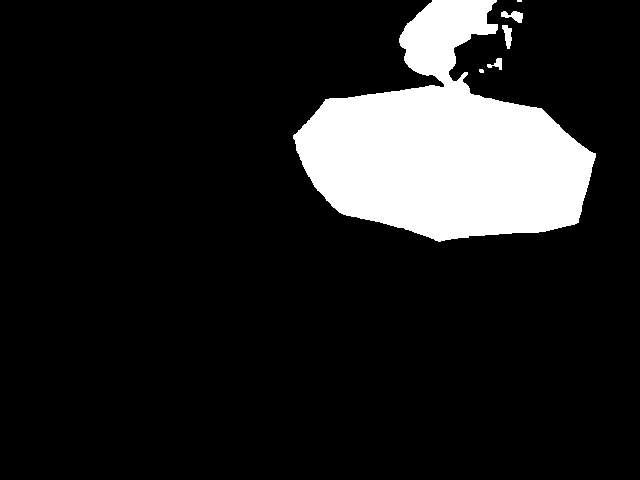} \\
	
{ \footnotesize{(a) Input}} &
{ \footnotesize{(b) dark regions}} &
{ \footnotesize{(c) BDRAR~\cite{DBLP:conf/eccv/ZhuDHFXQH18}}} &
{ \footnotesize{(d) DSC~\cite{DBLP:conf/cvpr/Hu0F0H18}}} &
{ \footnotesize{(e) DSDNet~\cite{DBLP:conf/cvpr/ZhengQCL19}}} &
{ \footnotesize{(f) MTMT~\cite{DBLP:conf/cvpr/Chen0WW0H20}}} &
{ \footnotesize{(g) Ours}}  &
{ \footnotesize{ (h) GT}} \\

\end{tabular}
\end{center}
\caption{Visual comparison of the shadow detection performances between our method ($7^{th}$ column) and the state-of-the-art methods ($3^{rd}-6^{th}$ columns) on ISTD~\cite{DBLP:conf/cvpr/WangL018}. The second column shows the recommended dark regions produced by our DRR module. Our method produces the closest results to the ground truth, particularly inside the recommended dark regions.}
\label{fig:visualISTD}
\end{figure*}


\subsection{Comparing to Shadow Detection Methods}

We compare our method with 8 state-of-the-art shadow detection methods: MTMT~\cite{DBLP:conf/cvpr/Chen0WW0H20}, DSDNet~\cite{DBLP:conf/cvpr/ZhengQCL19}, DSC~\cite{DBLP:conf/cvpr/Hu0F0H18}, BDRAR~\cite{DBLP:conf/eccv/ZhuDHFXQH18}, ADNet~\cite{DBLP:conf/eccv/LeVNHS18}, ST-CGAN~\cite{DBLP:conf/cvpr/WangL018}, scGAN~\cite{DBLP:conf/iccv/NguyenVZHS17}, and Unary-Pairwise~\cite{DBLP:conf/cvpr/GuoDH11}. All these methods are deep learning based, except for Unary-Pairwise~\cite{DBLP:conf/cvpr/GuoDH11}, which is based on hand-crafted features.

\noindent
{\bf Quantitative comparisons.}
The results of the quantitative comparison are shown in Table~\ref{tab:ti}. For a fair comparison, we report the performances of existing methods by directly referring to their papers.
For some unavailable results (\eg, DSC~\cite{DBLP:conf/cvpr/Hu0F0H18} lacks results on the ISTD dataset), we use the results either reported by Zheng~\etal~\cite{DBLP:conf/cvpr/ZhengQCL19} or by Chen~\etal~\cite{DBLP:conf/cvpr/Chen0WW0H20}.
Particularly, for the shadow detection performance of MTMT~\cite{DBLP:conf/cvpr/Chen0WW0H20} on the UCF dataset, we recompute its BER score based on their predicted shadow maps, due to the different ways of splitting the test data between Zheng~\etal~\cite{DBLP:conf/cvpr/ZhengQCL19} and Chen~\etal~\cite{DBLP:conf/cvpr/Chen0WW0H20}. 
%
%
%

As shown in Table~\ref{tab:ti}, our method achieves the best BER scores on both large-scale SBU~\cite{DBLP:conf/eccv/VicenteHYHS16} and ISTD~\cite{DBLP:conf/cvpr/WangL018} datasets. In particular, with ISTD, our method surpasses the second best method, MTMT~\cite{DBLP:conf/cvpr/Chen0WW0H20}, by 22.7\% on BER. Besides, comparing to the two latest sota methods, MTMT~\cite{DBLP:conf/cvpr/Chen0WW0H20} and DSDNet~\cite{DBLP:conf/cvpr/ZhengQCL19}, we observe that our method has a stronger discriminative ability at shadow regions with a much lower error rate, even though both MTMT~\cite{DBLP:conf/cvpr/Chen0WW0H20} and DSDNet~\cite{DBLP:conf/cvpr/ZhengQCL19} use additional data for training. We believe that this is because our strategy of delving into dark regions helps enhance the discriminative ability of the model in dark regions, leading to a performance boost at shadow regions since the shadow regions are usually much darker than non-shadow regions. Further, some existing methods ({\it \eg}, BDRAR~\cite{DBLP:conf/eccv/ZhuDHFXQH18}, DHAN~\cite{cun2019ghostfree} and ST-CGAN~\cite{DBLP:conf/cvpr/WangL018}) try to reduce the error rates at shadow regions by sacrificing the detection performance at non-shadow regions. In contrast, our method aims to achieve a better performance at the error-prone dark regions through learning the dark-aware features, without the need to sacrifice the performance at non-dark regions. In fact, our method achieves the lowest error rate at non-shadow regions on ISTD.

\ghk{Comparing to the performance gain on ISTD, our method does not have a significant performance gain over MTMT on the SBU test set (638 images), and does not outperforms MTMT on the small-scale UCF test set (110 images).
In our own opinion, the reason for the problem is mainly due to different shadow annotation procedures used for creating these two datasets. \ryn{In SBU, a lazy labeling strategy was adopted to quickly label the main shadow areas, and the annotations were then refined} with a Kernel Least Squares SVM. In contract, \ryn{in ISTD, the initial shadow masks were computed with a threshold policy and morphological filtering, and the label masks for each erroneous pixel were then manually corrected}.
It turns out that the shadow annotations of ISTD are much cleaner than those in SBU, while the shadow annotations of SBU contain many ``erroneous negatives" - shadow regions that were incorrectly labeled as negative~\cite{DBLP:conf/eccv/VicenteHYHS16}. These erroneous negative annotations can lead to a false high error rate in non-shadow areas. In fact, we do observe that our method has a higher error rate in the non-shadow areas in both SBU and UCF test sets (Since UCF is a very small dataset, we follow MTMT to train our model on SBU and directly test on UCF), but a low error rate in the shadow areas.
Since our method is fully-supervised, it is inevitably affected by these ``erroneous negatives" in SBU. In contrast, MTMT~\cite{DBLP:conf/cvpr/Chen0WW0H20} is a semi-supervised method, it can remedy the problem by leveraging additional data.}

%

\noindent
{\bf Visual comparisons.\label{sec:visual}} We further compare our method with these state-of-the-art methods visually. Figure~\ref{fig:visual} and Figure~\ref{fig:visualISTD} compare our method with four state-of-the-art deep learning based shadow detection methods (\ie, DSC~\cite{DBLP:conf/cvpr/Hu0F0H18}, BDRAR~\cite{DBLP:conf/eccv/ZhuDHFXQH18}, DSDNet~\cite{DBLP:conf/cvpr/ZhengQCL19}, and MTMT~\cite{DBLP:conf/cvpr/Chen0WW0H20}) on images from the SBU~\cite{DBLP:conf/eccv/VicenteHYHS16} and ISTD~\cite{DBLP:conf/cvpr/WangL018} datasets, respectively. 
We first show the visual results of our DRR module, which aims to identify and recommend dark regions for the DASA module for analysis. As shown in Figure \ref{fig:visual}(b) and Figure~\ref{fig:visualISTD}(b), the recommended dark regions generally cover the low-intensity pixels that may be challenging for existing shadow detectors. 
We then show the visual results of existing methods and our method in Figure \ref{fig:visual}(c-g) and Figure~\ref{fig:visualISTD}(c-g). We can see that our method performs better than these existing methods, especially in the recommended dark regions. For example, in the first row of Figure~\ref{fig:visual}, existing methods wrongly classify the two dark horizontal stripes as shadow regions. Although our DRR module highlights these two regions as recommended dark regions, our DASA module can correctly differentiate them as non-shadow regions.
In the second row of Figure~\ref{fig:visual}, some existing methods may mis-recognize part of shadow region inside the poster as non-shadow, due to the intensity ambiguity around the region. In contrast, our model produces a shadow map that is closest to the ground truth shadow map. It is interesting to note that our result is even better than the result from DSDNet~\cite{DBLP:conf/cvpr/ZhengQCL19}, which explicitly learns distraction features to address this kind of problem.

\begin{figure}[t]
\begin{center}
\begin{tabular}{c}
	\includegraphics[width=\linewidth]{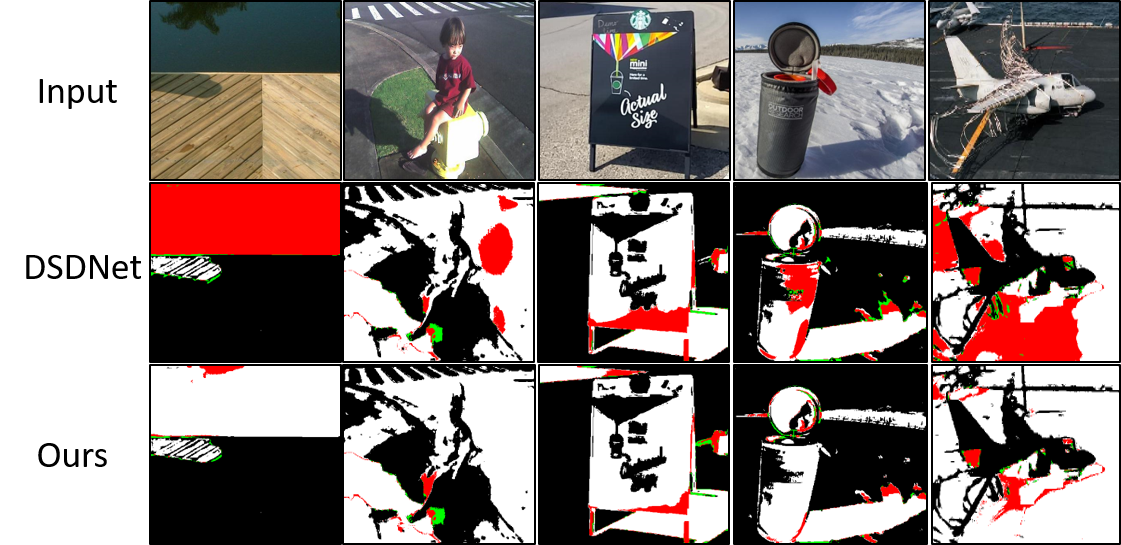}
\end{tabular}
\end{center}
\caption{Visualization of prediction errors inside and outside of the recommended dark regions.
The recommended dark regions are marked in white or red, while the rest is marked in black or green.
The prediction errors inside the recommended dark regions are marked in red, and those outside the recommended dark regions are marked in green.
Compared to DSDNet~\cite{DBLP:conf/cvpr/ZhengQCL19} (second row), our method performs significantly better by focusing on the error-prone dark regions as recommended by the DRR module, with much smaller red areas (third row).}
\label{fig:darkerr}
\end{figure}

\noindent
{\bf Visual comparisons based on the recommended dark regions.\label{sec:drrvisual}}
Finally, we visually compare the detection results of our method with DSDNet~\cite{DBLP:conf/cvpr/ZhengQCL19}, which constructs \emph{distraction} labels to train their model to distinguish between ambiguous shadows and non-shadow pixels, both inside and outside of the recommended dark regions. Figure~\ref{fig:darkerr} shows the comparison, in which we superimpose the detection errors of DSDNet~\cite{DBLP:conf/cvpr/ZhengQCL19} and of our proposed method. We visualize the prediction errors inside the recommended dark regions in red and those outside in green. We have three observations from this result:
(1) Errors inside the recommended dark regions are larger than errors outside; 
(2) Even with additional ``distraction'' labels, DSDNet~\cite{DBLP:conf/cvpr/ZhengQCL19} is still incapable of distinguishing between shadow and non-shadow pixels inside the recommended dark regions. For example, in the first column, DSDNet~\cite{DBLP:conf/cvpr/ZhengQCL19} would recognize the water at the top as shadow; and
(3) Our method makes far smaller mistakes in the recommended dark regions, verifying our motivation that one should delve into the dark regions to learn robust shadow features.

\subsection{Comparing to Related State-of-the-art Methods}
Since shadow detection is also related to salient object detection, semantic segmentation and shadow removal methods, for a more thorough evaluation of our method, \ghk{we also \ryn{compare with} two salient object detection methods (SRM \cite{DBLP:conf/iccv/WangBZZL17} and EGNet \cite{DBLP:conf/iccv/ZhaoLFCYC19}), one \ryn{semantic} segmentation method (PSPNet \cite{DBLP:conf/cvpr/ZhaoSQWJ17}), and three shadow removal \ryn{methods} (DeshadowNet \cite{DBLP:conf/cvpr/QuTHTL17},  DHAN~\cite{cun2019ghostfree} and ARGAN~\cite{DBLP:conf/iccv/DingLZX19}~\footnote{\ghk{The authors of ARGAN~\cite{DBLP:conf/iccv/DingLZX19} did not release their pre-trained model, detection results and training data. Their train/test set split is also unknown\ryn{. So,} we directly copy their reported numbers (fully-supervised) from their paper as a reference.}}), as shown in Table \ref{tab:ti}. Among these methods, shadow removal methods generally perform better, as shadow removal methods require an understanding of where \ryn{the shadows are in order to remove them} accurately. Besides, DHAN~\cite{cun2019ghostfree} achieves the best performance in shadow \ryn{areas} on SBU. This may \ryn{be due to} its generative adversarial network used for data augmentation. However, its error rates in non-shadow \ryn{areas on SBU are} much higher. Our proposed method, which explicitly learns to differentiate shadow and non-shadow pixels within the dark regions, plays favorably against these methods.}

\subsection{Internal Analysis}
\label{sec:abla}

{\bf Analysis of the DRR Module.}
In Figure~\ref{fig:introduction}, we have shown that low-intensity ranges contain more errors than medium-/high-intensity ranges. An intuitive and straightforward method to generate the recommended dark regions is to apply threshold policy. Thus, we create a baseline method for comparison by replacing our DRR module with an intensity threshold strategy.
Mathematically, given an intensity map $A \in R^{H \times W}$, threshold $\tau$, the corresponding dark regions mask is B, where $B_{ij} = A_{ij}<\tau $.
We experiment with three thresholds, 50, 100, and 150, according to Figure~\ref{fig:introduction} and Figure~\ref{fig:pdf}. We denote these three variants as \emph{intensity-50}, \emph{intensity-100}, and \emph{intensity-150}, respectively.

We further study how the size of the receptive field affects the performance, by replacing our shallow DRR module with CAN24~\cite{Chen_2017_ICCV}, which has a larger receptive field due to the increasing dilation rates layer by layer. We denote this variant as \emph{CAN24}.
We also compare with a deeper version of the DRR module that uses 16 convolutional layers. We denote this variant as \emph{deeper-DRR}.

As shown in Table \ref{tab:ablation1}, our DRR strategy outperforms all the above variants, including intensity-50, intensity-100, intensity-150, CAN24 and deeper-DRR, with a significant margin.
The main reason for the failure of the intensity thresholding method is that a single threshold does not work for all images due to the diverse illumination conditions presented in different images, as shown in Figure~\ref{fig:pdf}. By learning the dark regions adaptively for each input image, both CAN24~\cite{Chen_2017_ICCV} and deeper-DRR perform better than the intensity thresholding method.
However, incorporating larger receptive fields via dilated convolutions (CAN24~\cite{Chen_2017_ICCV}) or deeper architecture (deeper-DRR) still perform worse than our DRR module. One possible reason is that the dark region maps recommended by the above two variants shrink the divergence from the shadow score map, as a result of their larger receptive fields. Since this divergence information is an important source of learning the subtle contrast information, both variants suffer from the reduced divergence, resulting in a performance drop comparing to our shallow DRR module.

\begin{table}[t]
\centering
\caption{Evaluation of the DRR module. The BER scores are reported on SBU~\cite{DBLP:conf/eccv/VicenteHYHS16}.}
\label{tab:ablation1}
\begin{tabular}{l|l}
Method                & BER$\downarrow$            \\
\hline
intensity-50      & 3.57       \\
intensity-100         & 3.43      \\
intensity-150 & 3.64  \\
CAN24~\cite{Chen_2017_ICCV} & 3.35 \\
deeper-DRR & 3.38 \\
DRR module (Ours) & \textbf{3.04}           \\
\hline
\end{tabular}
\end{table}

\begin{table}[t]
\centering
\caption{Evaluation of the DASA module. The BER scores are reported on SBU~\cite{DBLP:conf/eccv/VicenteHYHS16}.}
\label{tab:ablation2}
\begin{tabular}{l|l}
Method                & BER$\downarrow$            \\
\hline
w/o DASA \& DRR                  & 3.57           \\
simple delving                     & 3.44           \\
divergence only                             & 3.28           \\
full model (Ours) & \textbf{3.04}           \\
\hline
\end{tabular}
\end{table}

\noindent
{\bf Analysis of the DASA Module.}
We also investigate the effectiveness of the proposed DASA module. Table \ref{tab:ablation2} reports the experiment results on SBU~\cite{DBLP:conf/eccv/VicenteHYHS16}. It shows that removing the DASA module and DRR module (\emph{w/o DASA, w/o DRR}) causes a significant performance drop from our proposed model. 
%
%
%
We then remove both the intersection and the difference operators (\ie, do not leverage the shadow score maps to further divide the recommended dark regions in the dark branch). We refer to this variant as \emph{simple delving}. We can see that this variant performs better than\emph{" w/o DASA, w/o DRR"}, but still performs inferior to the full DASA module. If we now add the intersection and difference operators but only focus on divergence with only one LSD (referred to as \emph{divergence only}),
%
%
we can see that the performance further improves. This once again demonstrates that our method can benefit from explicitly learning the divergence between shadow score map and the recommended dark region map. 
Finally, our proposed DASA model (Ours), which considers both divergence (via the difference operator) and co-occurrence (via the intersection operator), produces the best performance.

\noindent
{\bf Analysis of the Global Context Network.}
%
We further evaluate our fusion strategy adopted by the Global Context network.
To verify its effectiveness, we compare the performance of our global context network with two baselines (\ie, \emph{w/o DS module} and \emph{w/o DS loss}) and two ablated versions (\ie, \emph{w/o FN sub-module} and \emph{w/o FP sub-module}) of DSDNet~\cite{DBLP:conf/cvpr/ZhengQCL19}.
For a fair comparison, we use the results reported by DSDNet~\cite{DBLP:conf/cvpr/ZhengQCL19} in here. 
Besides, we also adapt the fusion blocks in our method in parallel, as DSDNet~\cite{DBLP:conf/cvpr/ZhengQCL19} does (see Figure~\ref{fig:fusestrategy}(a)). We refer to this as \emph{parallel fusion}. We compare it with our sequential strategy as shown in Figure~\ref{fig:fusestrategy}(b), and denote it as \emph{sequential fusion}. 

As shown in Table~\ref{tab:ablgcn}, our global context network outperforms all baselines and ablated versions of DSDNet~\cite{DBLP:conf/cvpr/ZhengQCL19} as well as the parallel fusion strategy. This indicates that our global context network with the sequential fusion strategy can learn better global context for shadow detection.

\begin{table}[t]
\centering
\caption{Evaluation of the Global Context Network. The BER scores are reported on SBU~\cite{DBLP:conf/eccv/VicenteHYHS16}.}
\label{tab:ablgcn}
\begin{tabular}{l|l}
Method                & BER$\downarrow$            \\
\hline
w/o DS Module and DS loss    & 4.42           \\
w/ only DS Module   & 3.62           \\
w/ only DS loss  & 3.89           \\
w/o FN Sub-module & 3.71 \\
w/o FP Sub-module & 3.68 \\
\hline
parallel fusion & 3.68 \\
sequential fusion (Ours) & \textbf{3.57} \\
\hline
\end{tabular}
\end{table}

\begin{table}[t]
\centering
\caption{Evaluation of the shadow detection architecture. Two streams with DRR: the dark regions are recommended by our DRR module. Two streams with intensity-50/100/150: the dark regions are determined by through intensity threshold at 50/100/150. The BER scores are reported on SBU~\cite{DBLP:conf/eccv/VicenteHYHS16}.}
\label{tab:ablation3}
\begin{tabular}{l|l}
Method                & BER$\downarrow$           \\
\hline
two streams with DRR & 5.82  \\
two streams with intensity-50 & 5.67  \\
two streams with intensity-100 & 6.24  \\
two streams with intensity-150 & 5.92  \\
\hline
single stream without DASA & \textbf{3.57} \\
\hline
\end{tabular}
\end{table}

\noindent
{\bf Analysis of the Shadow Detection Architecture.}
Finally, we evaluate how our shadow detection architecture affects the performance. To this end, we formulate another two-stream architecture that first divides the image into dark regions and non-dark regions, using our DRR (referred to as \emph{two streams with DRR}) or intensity thresholding (referred to as \emph{two streams with intensity-50}, \emph{two streams with intensity-100}, and \emph{two streams with intensity-150}), and then apply two global context networks separably on the two types of regions for shadow detection.

As shown in Table~\ref{tab:ablation3}, all two-stream variants yield worse results. Although a two-stream architecture may learn better local discriminative features, all these two-stream variants lack a good global understanding as the two steams focus only on their own regions. In contrast, our model focuses on a global understanding in the first stage (via the global context network) and takes a closer look at the dark regions in the second stage (via the DRR module and the DASA module), yielding the best results. 
%
This shows that global image understanding and dark-region focusing are essential and complementary to robust shadow detection.

\begin{figure}[t]
\renewcommand{\tabcolsep}{1pt}
\begin{center}
\begin{tabular}{ccc}
	\includegraphics[width=0.32\linewidth]{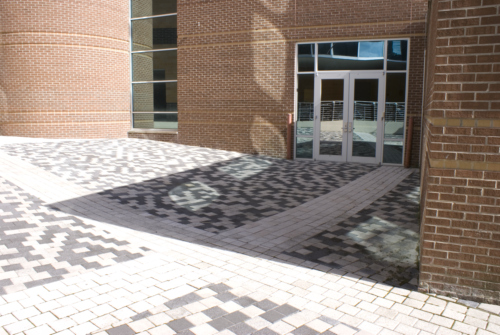} &
	\includegraphics[width=0.32\linewidth]{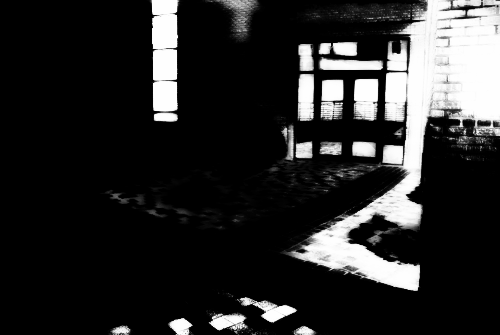} &
	\includegraphics[width=0.32\linewidth]{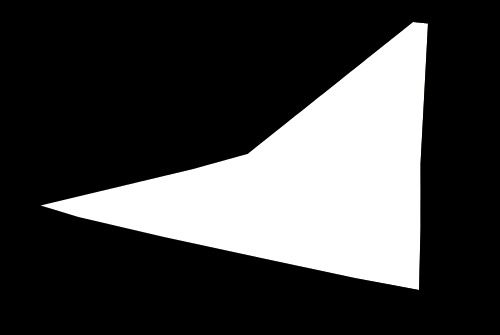} \\

	\includegraphics[width=0.32\linewidth]{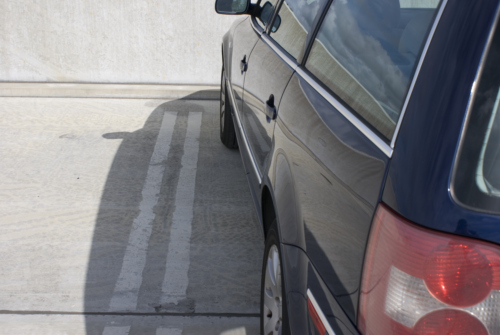} &
	\includegraphics[width=0.32\linewidth]{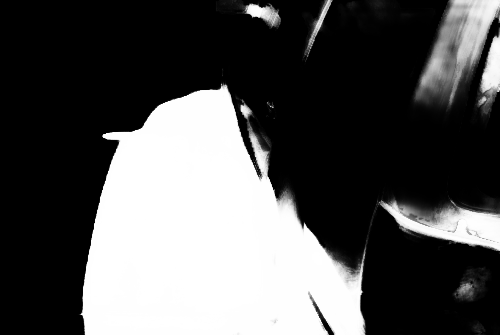} &
	\includegraphics[width=0.32\linewidth]{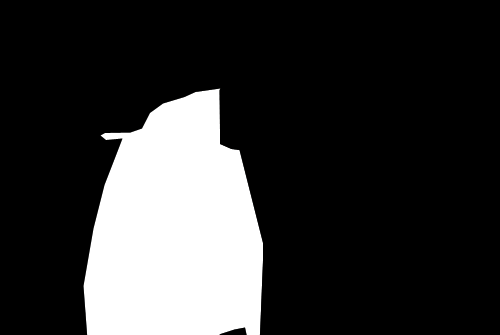} \\
{ \footnotesize{(a) Input}} &
{ \footnotesize{(b) Ours}}  &
{ \footnotesize{(c) Ground truth}} \\
\end{tabular}
\end{center}
\caption{Failure cases. Our method may fail to detect shadows accurately in some challenging scenes where the intensity differences in the dark regions are caused by other factors, \eg, highlights (first row) and weak reflections (second row).} 
\label{fig:failure}
\end{figure}

\begin{figure*}[t]
\renewcommand{\tabcolsep}{1pt}
\begin{center}
\begin{tabular}{cccccccccccc}
    \input{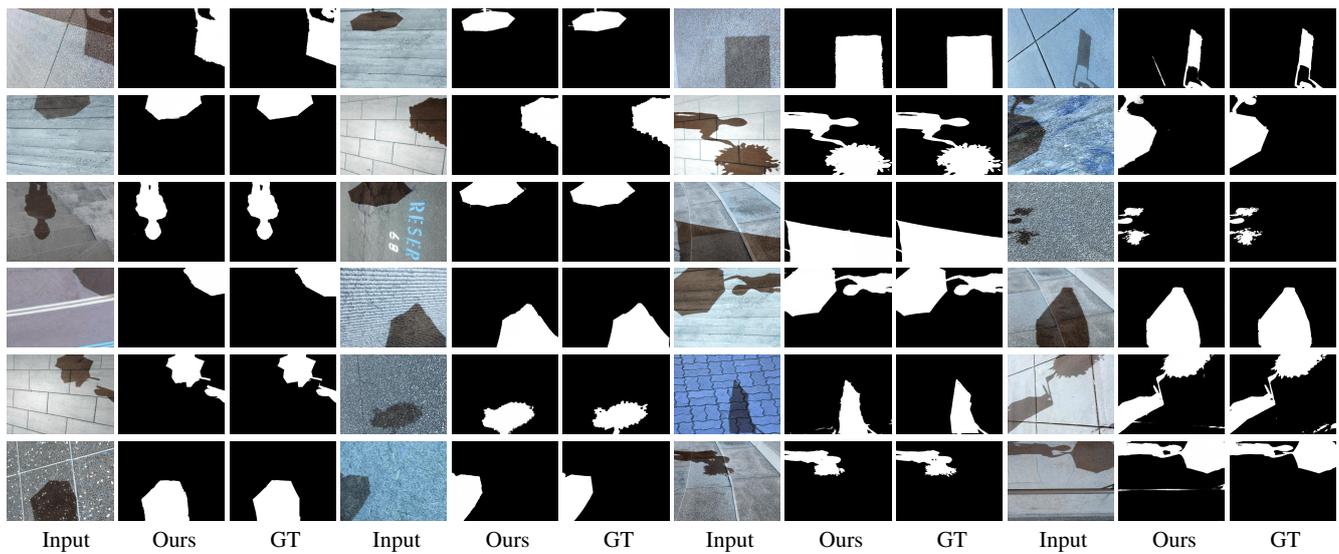}
{ \footnotesize{Input}} &
{ \footnotesize{Ours}} &
{ \footnotesize{GT}} &

{ \footnotesize{Input}} &
{ \footnotesize{Ours}} &
{ \footnotesize{GT}} &

{ \footnotesize{Input}} &
{ \footnotesize{Ours}} &
{ \footnotesize{GT}} &

{ \footnotesize{Input}} &
{ \footnotesize{Ours}} &
{ \footnotesize{GT}} \\

\end{tabular}
\end{center}
\caption{Visual results on synthetic data. We synthesize complex shadow images using SMGAN~\cite{cun2019ghostfree}. The shadow masks come from the ISTD~\cite{DBLP:conf/cvpr/WangL018} dataset, and the shadow-free images are collected from the USR~\cite{DBLP:conf/iccv/HuJFH19} dataset.}
\label{fig:synthesize_part}
\end{figure*}

\begin{figure*}[t]
\renewcommand{\tabcolsep}{1pt}
\begin{center}
\begin{tabular}{cccccccc}
\includegraphics[width=0.125\linewidth]{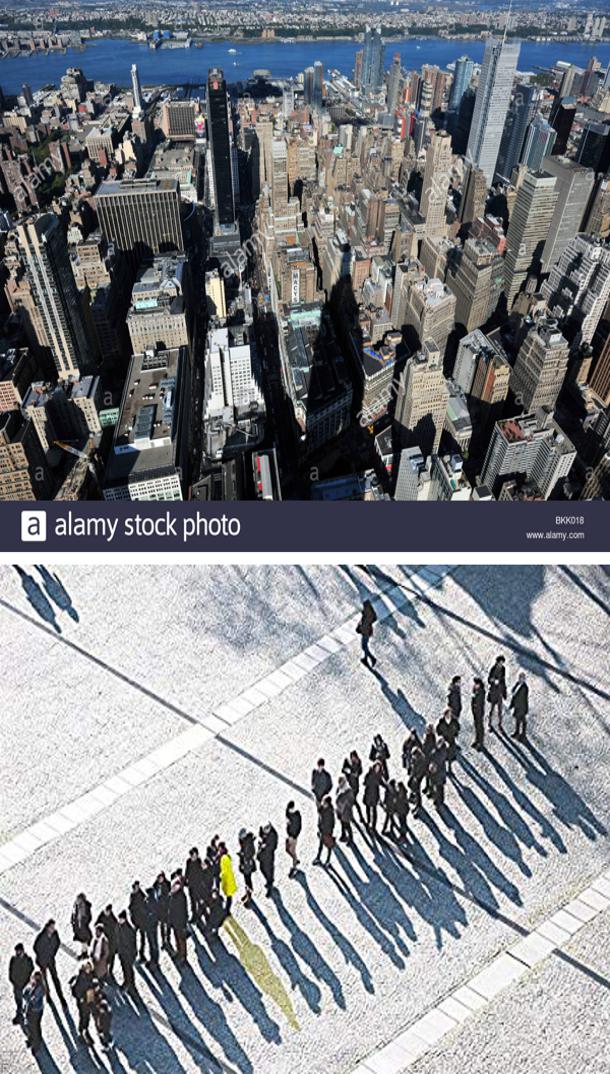} &
\includegraphics[width=0.125\linewidth]{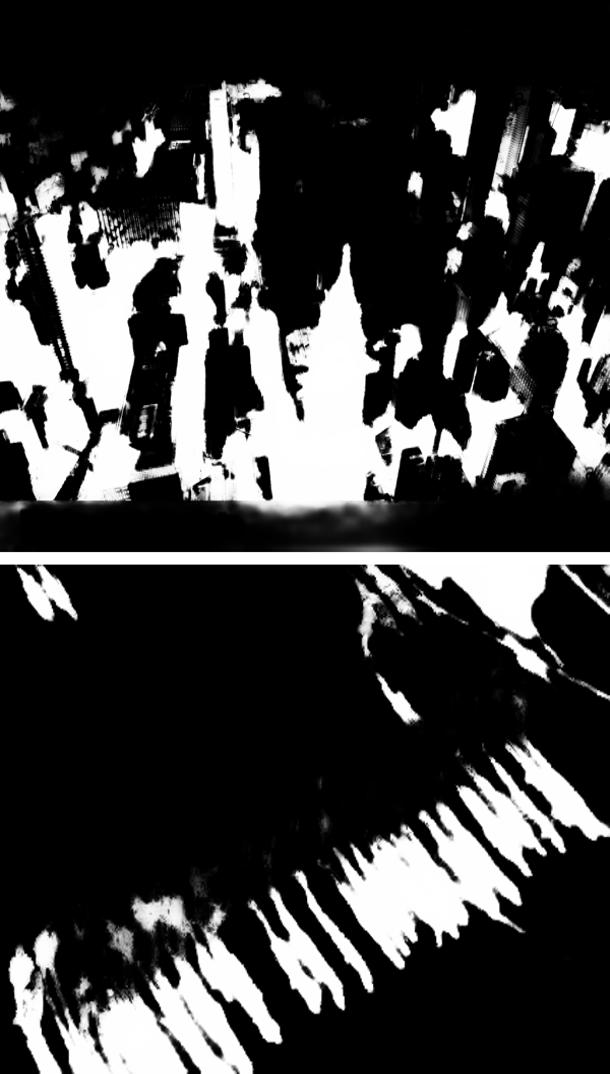} &
\includegraphics[width=0.125\linewidth]{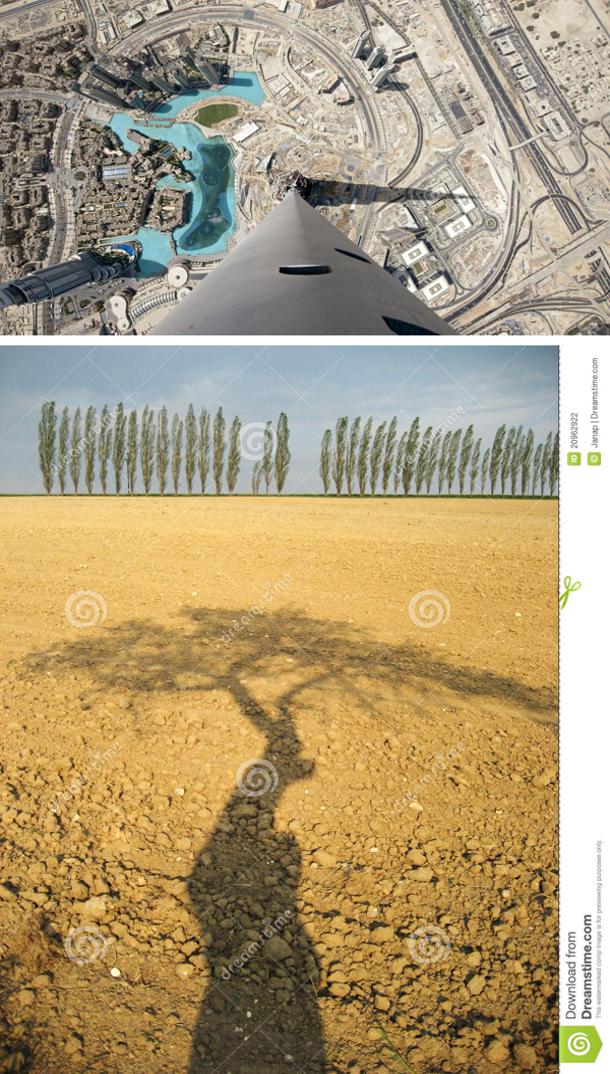} &
\includegraphics[width=0.125\linewidth]{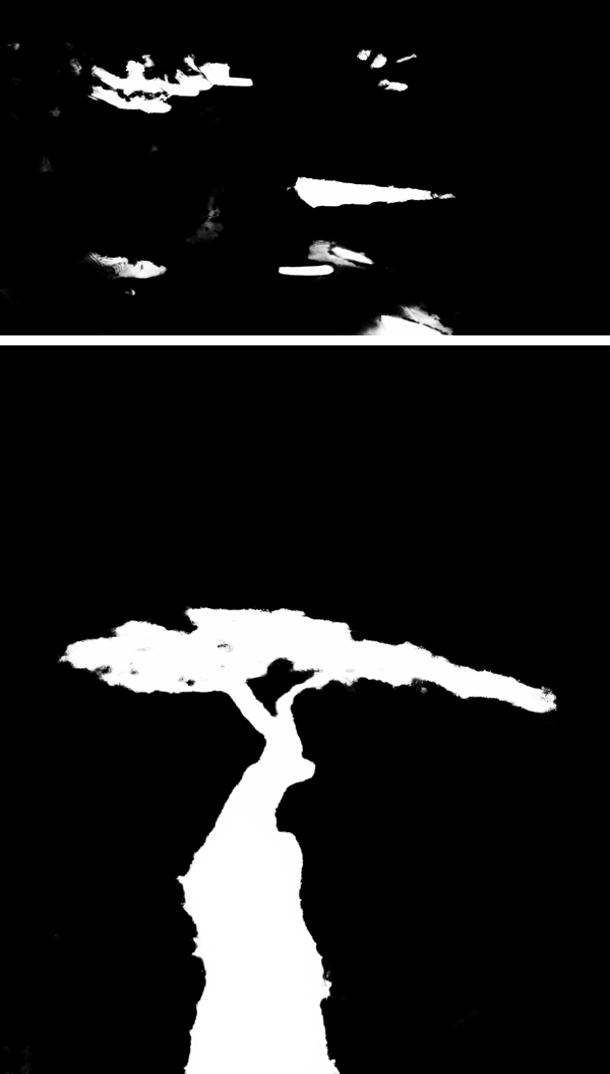} &
\includegraphics[width=0.125\linewidth]{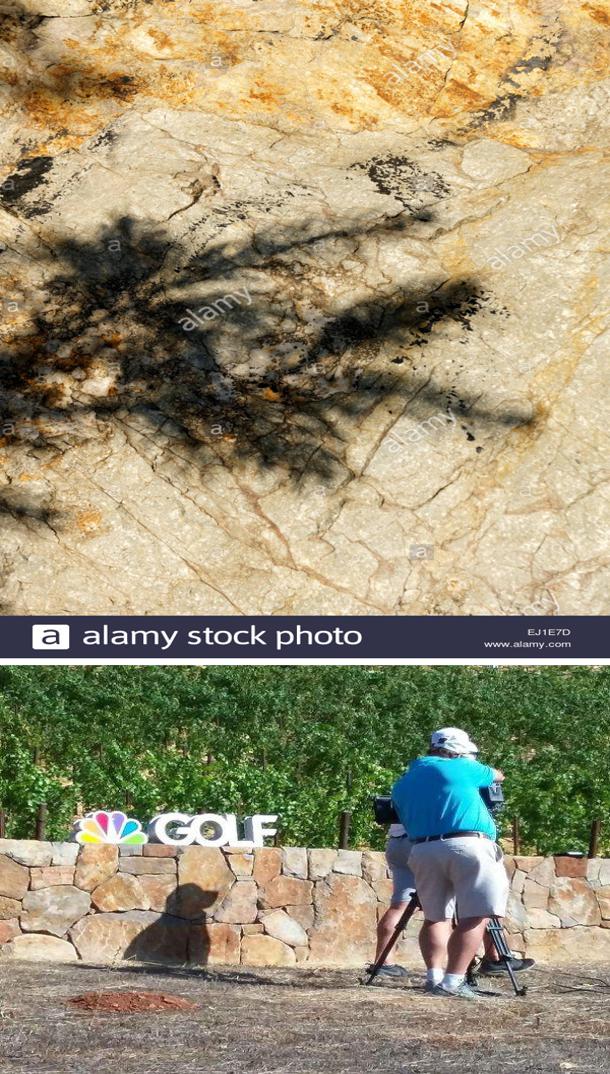} &
\includegraphics[width=0.125\linewidth]{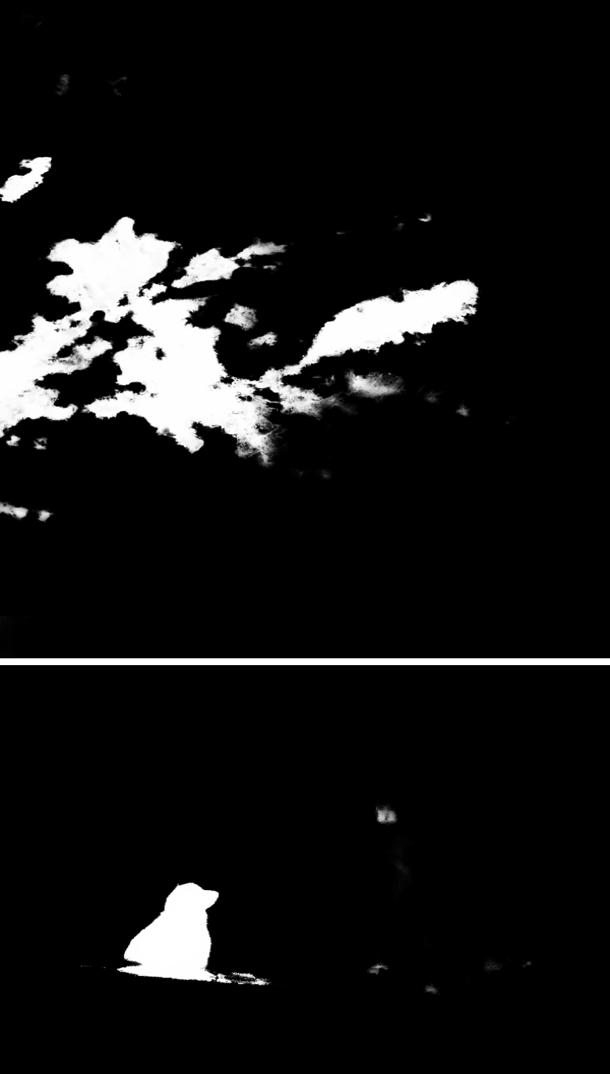} &
\includegraphics[width=0.125\linewidth]{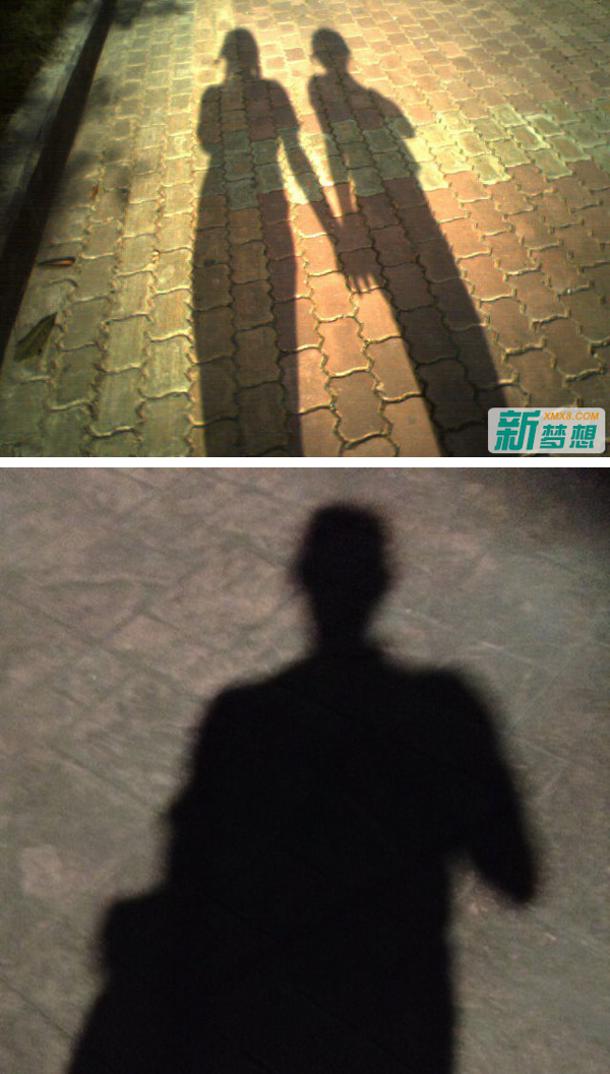} &
\includegraphics[width=0.125\linewidth]{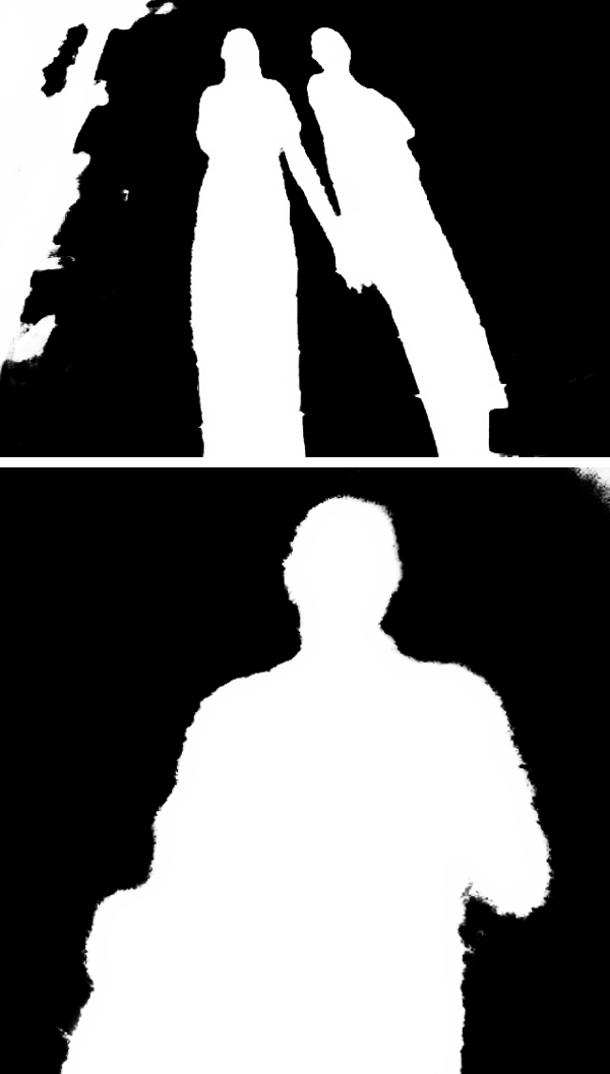} \\

{ \footnotesize{Input} } &
{ \footnotesize{Ours} } &
{ \footnotesize{Input} } &
{ \footnotesize{Ours} } &
{ \footnotesize{Input} } &
{ \footnotesize{Ours} } &
{ \footnotesize{Input} } &
{ \footnotesize{Ours} } \\

\end{tabular}

\end{center}
\caption{Visual results on real data. We test our model on real shadow images collected from the Internet.}
\label{fig:real_part}
\end{figure*}

\noindent
{\bf Visual Results of More Complex Images.} \ghk{In addition to three popular shadow detection datasets, {\it \ie}, SBU~\cite{DBLP:conf/eccv/VicenteHYHS16}, UCF~\cite{CVPR.2010.5540209} and ISTD~\cite{DBLP:conf/cvpr/WangL018}, we further test our model on two kinds of shadow image sets: (1) we synthesize shadow images using SMGAN~\cite{cun2019ghostfree}, and (2) we collect shadow images from the Internet. 
The synthetic shadow images differ from our training data (\ie, SBU or ISTD training images) mainly in the color temperature, while the images collected from the Internet contain shadows of complex shapes, small sizes, low contrasts, {\it etc}. \ryn{Figures~\ref{fig:synthesize_part} and~\ref{fig:real_part}, respectively, show some visual results. We can see} that the proposed model can handle both synthesized scenes and real-world complex scenes \ryn{well}.}


\section{Conclusion}
In this paper, we have investigated a fundamental problem of existing shadow detection methods, and demonstrated that existing methods have a much higher error rate in dark regions than bright regions. 
To address this problem, we propose a dark-region aware approach to exploit the subtle contrast information between shadow and non-shadow pixels inside the dark regions. Our model includes a simple yet effective dark-region recommendation (DRR) module to recommend dark regions from the input image, and a novel dark-aware shadow analysis (DASA) module to learn the dark-aware shadow features within the recommended dark regions. 
%
%
We conduct extensive experiments to verify the effectiveness of the proposed method on detecting shadows in the dark regions, and show that our model achieves superior shadow detection performances on existing shadow detection datasets.

Our model does have some limitations. As shown in Figure~\ref{fig:failure}, our method may fail to detect shadows accurately in some challenging scenes where the intensity difference in the dark regions may be caused by other factors, \eg, highlights (first row) and weak reflections (second row). In these cases, the subtle differences between shadow and non-shadow pixels in the dark regions can be upset by the presence of the highlights or reflections, causing the degradation of detection performances. In the future work, we are interested in resolving such challenging scenarios by incorporating a comprehensive light reflection model.

%

\begin{acknowledgements}
This work desribed in this paper was partly supported by a GRF grant from the Research Grants Council of Hong Kong (RGC Ref.: 11205620).
\end{acknowledgements}

\noindent
{\small  {\bf Data availability statement} The authors confirm that the data supporting the findings of this study are openly available.}

%
%

\bibliographystyle{spmpsci}      
\bibliography{egbib}   

\newpage


\end{sloppypar}
\end{document}